\documentclass[lettersize,journal]{IEEEtran}

\usepackage{amsmath,amsfonts,amssymb,amsthm}
\usepackage{mathtools}
\usepackage{enumerate}
\usepackage[ruled, vlined, linesnumbered, resetcount]{algorithm2e}
\usepackage{array}
\usepackage[caption=false, font=footnotesize, position=top]{subfig}
\usepackage{textcomp}
\usepackage{url}
\usepackage{verbatim}
\usepackage{graphicx}
\usepackage{cite}
\usepackage{multirow}
\usepackage{colortbl}
\usepackage{cases}
\usepackage[usestackEOL]{stackengine}
\usepackage[normalem]{ulem}
\usepackage{lipsum} 
\usepackage{threeparttable}
\usepackage{orcidlink}

\usepackage{nccmath}
\usepackage{dblfloatfix}

\def\BibTeX{{\rm B\kern-.05em{\sc i\kern-.025em b}\kern-.08em
    T\kern-.1667em\lower.7ex\hbox{E}\kern-.125emX}}
\begin{document}

\title{Modelling, Positioning, and Deep Reinforcement Learning Path Tracking Control of Scaled Robotic Vehicles: Design and Experimental Validation}

\author{
Carmine~Caponio,
Pietro~Stano,
Raffaele~Carli,~\IEEEmembership{Senior~Member,~IEEE},\\
Ignazio~ Olivieri,
Daniele~Ragone, 
Aldo~Sorniotti,
~and~Umberto~Montanaro%

\thanks{Manuscript received on \today.\\
(\emph{Corresponding author: U. Montanaro.})}
\thanks{
}
\thanks{C. Caponio, P. Stano, and U. Montanaro are with the School of Mechanical Engineering Sciences, University of Surrey, UK (e-mail: {\tt\small\{c.caponio, p.stano, u.montanaro\}@surrey.ac.uk}).}
\thanks{R. Carli, I. Olivieri, and D. Ragone are with the Department of Electrical and Information Engineering of the Polytechnic of Bari, Italy (e-mail: {\tt\small raffaele.carli@poliba.it}, {\tt\small\{i.olivieri1, d.ragone\}@studenti.poliba.it}).}
\thanks{A. Sorniotti is with the Department of Mechanical and Aerospace Engineering of the Polytechnic of Turin, Italy (e-mail: {\tt\small aldo.sorniotti@polito.it}).}
}

\markboth{}%
{Shell \MakeLowercase{\textit{et al.}}: A Sample Article Using IEEEtran.cls for IEEE Journals}

\maketitle

\begin{abstract}
Mobile robotic systems are becoming increasingly popular. These systems are used in various indoor applications, ranging from warehousing and manufacturing to test benches for the assessment of advanced control strategies, such as artificial intelligence (AI)-based control solutions, just to name a few. Scaled robotic cars are commonly equipped with a hierarchical control architecture that includes tasks dedicated to  vehicle state estimation and control. This paper covers both aspects by  proposing \emph{(i)} a federated extended Kalman filter (FEKF), and \emph{(ii)} a novel deep reinforcement learning (DRL) path tracking controller trained via  an expert demonstrator to expedite the learning phase and increase robustness to the simulation-to-reality gap. The paper also presents the formulation of a vehicle model along with an effective yet simple procedure for identifying its parameters. The experimentally validated model is used for \emph{(i)} supporting the design  of the FEKF and \emph{(ii)} serving as a digital twin for training the proposed DRL-based path tracking algorithm. Experimental results confirm the ability of the FEKF to improve the estimate of the mobile robot’s position. Furthermore, the effectiveness of the DRL path tracking strategy is experimentally tested along manoeuvres not considered during training, showing also the ability of the AI-based solution to outperform model-based control strategies and the demonstrator. The comparison with  benchmarking controllers is quantitatively evaluated through a set of key performance indicators.
\end{abstract}

\def\abstractname{Note to Practitioners}
\begin{abstract}
This paper presents an effective tool-chain to control mobile robots starting from the mobile robotic system modelling to the design of an advanced DRL path tracking solution. Specifically, the paper proposes \emph{(i)} a simple to replicate but yet effective two stage least-square (LS) approach for the parameter identification of the longitudinal and lateral dynamics of the robotic car; \emph{(ii)} the use of a no-reset federated Kalman filter for utilising all the sensors commonly equipping a scaled robotic car to improve its positioning; and \emph{(iii)} the use of a expert demonstrator for speeding up the training phase and  mitigating the simulation-to-reality gap resulting from the mismatch between the simulation and the experimental environment. 
\end{abstract} 

\begin{IEEEkeywords}
automated vehicles, vehicle modelling, position estimation, path tracking, deep reinforcement learning. 
\end{IEEEkeywords}

\section{Introduction}
With the in-depth development of Industry 4.0 worldwide, mobile robotic systems have attracted the attention of large research centres, industry  and governments [1] and have found several applications in laboratories, industry, warehousing and transportation [2]. In this context, scaled robotic cars are commonly used as assessment platforms for autonomous driving because, as shown in [3]-[5],  they provide \emph{(i)} cost-effective testing, signifying substantial cost savings in comparison to conducting experiments with full-sized vehicles; \emph{(ii)} repeatability of the testing environment, enabling for more accurate tuning of the control algorithms; and iii) enhanced safety, as scaled robotic cars entail less severe consequences in the event of system failures or collisions. Consequently, this mobile robots serve as versatile platforms for testing path planning (PP) [6],[7], path tracking (PT) [8], as well as vehicle platooning algorithms [9]. 

Similar to full scale vehicles, also the autonomous driving system for robotic cars requires a precise identification of the car's position and orientation concerning its surroundings, the road, obstacles, as well as other vehicles [10]. Localization typically entails the use of lidar sensors, which determine target distances by measuring signal arrival times at the receiver and infrared intensity reflecting off obstacles. Although lidar sensors offer better position accuracy compared to other localization systems, such as radar and global positioning system (GPS), they often fall short in practice due to their  sensitivity to external factors, such as environmental conditions,  and long data processing times, making them suboptimal for real-time decision-making in driving scenarios [11]. Consequently, lidars  are commonly integrated into sensor fusion systems, combining data from other sensors such as cameras, GPS, and inertial measurement units (IMUs), to compensate individual sensor limitations and enhance perception capabilities [12]. Extended Kalman filters (EKFs) excel in the fusion of data from multiple sources and account for the nonlinear vehicle dynamics, thus enabling a more accurate and robust location and vehicle state estimate even when data from some sensors are  unreliable or missing [13].  For example, Liu \textit{et al.} [14] introduce a kinematic-based EKF to estimate automated vehicle’s (AV) position and orientation by using data from lidar, GPS, and IMU sensors. The EKF is then further coupled with a rule-based strategy, which firstly detects sensor data inconsistencies and subsequently eliminates such data from the measurements. A two-step estimation process is outlined in [15] which consists of: \emph{(i)} a forward kinematic-based EKF to estimate vehicle localization using data from global navigation satellite system (GNSS) and IMU; and \emph{(ii)} a backward filter consisting of a recursive time series smoothing algorithm which refines past state estimates based on all available data.  In [16], the EKF incorporates a dynamic single-track model (STM) that accounts for nonlinear tyre behaviour and a dynamic adjustment of the measurements to improve the vehicle states and pose estimation. Attempts to combine KF and EKF solutions based on different modelling and sensors, aiming to increase estimate accuracy, have also been presented. A possible approach for fusing KFs estimates is through federated Kalman filters (FKFs) consisting of a master filter used for fusing the outputs of a bank of local EK filters. For instance in [17] the traditional KF has been combined with a cubature KF through a federated filtering approach for the indoor positioning of scaled cars. Furthermore, the linearity of the output equation also enabled to reduce the computation burden while guaranteeing the positioning accuracy.  In [18] EKFs have been used as local filters with a reset-based federated KF architecture and it was proven that better positioning estimates for a scaled car can be achieved with respect to a centralised KF approach. However, in  [17],[18] only numerical simulations have been presented to validate the filtering approach.

A precise vehicle state estimation and positioning play a key role for both PP, i.e., the definition of a optimised collision free path from an initial to an end goal state [19],[20] and PT applications, i.e., definition of the actuator commands, typically expressed in terms of steering angle and traction/braking torques, to follow predefined paths while adjusting to dynamic and uncertain real-world conditions [21],[22]. In this framework  scaled robotic cars have been used to experimentally test the effectiveness of advanced PT solutions. For example,  an adaptive backstepping strategy based on a vehicle kinematic model has been presented in [23] to reduce the converge rate of the lateral error dynamics. A hierarchical control strategy is presented in [24] and it consists of \emph{(i)} a feedback+feedforward (FB+FF) control action which determines the desired yaw rate based on the reference curvature and a variable look-ahead distance; and \emph{(ii)} a model reference adaptive controller strategy based on a kinematic reference model to adaptively modulate the steering angle while ensuring accurate tracking of the reference yaw rate. An active disturbance rejection control system is proposed in [25] for lane-keeping applications. The controller design is based on a single-track model and experimental results shown superior tracking performance with respect to those achievable via a PID solution. 
Also model predictive control PT solutions have been tested on scaled cars [26]-[28]. Specifically, an optimal robust linear matrix inequality based MPC is designed in [26] with the aim of minimising the battery consumption. While in [27],[28] a MPC based on a STM is introduced to optimise vehicle speed profile and the reference path.
Despite being effective, model-based approaches require an accurate model of the vehicle, which can be challenging in real-world scenarios where external disturbances can affect tracking performance. Moreover, the model may not account for unexpected changes or handling complex environment [29]. In response to these challenges, AI-based PT algorithms, such as deep reinforcement learning (DRL) strategies, have gained prominence due to their adaptability to new conditions and to potentially enhance overall performance [30]. For example, authors in [31] propose a DRL strategy based on a deep deterministic policy gradient (DDPG) to simultaneously control the steering and vehicle throttle. The proposed DRL is trained and tested in the Carla simulator. Furthermore, this AI strategy has been implemented on a scaled robotic car but experiments exhibit a relatively high lateral error also during a constant-speed test. In the contest of the DRL control, experimental closed-loop performance can degrade due to possible  mismatches between the simulation environment used for the training phase and the experimental one. This drawback of DRL solutions is known as simulation-to-reality gap [32]. Additionally, DRL strategies can suffer of low learning efficiency which consists of a long training phase before the agent's policy converges to an optimal one [33]. A possible approach to address this issue is to equip the DRL scheme with a demonstrator (DRL-D) which combines  deep reinforcement learning strategies with forms of supervised learning [34]. In DRL-D solutions, during the training phase, the agent first learns to mimic the demonstrator's actions, by minimising the difference between its predictions and the demonstrator's actions. Then, the training model finds a new optimal solution thanks to its direct interaction with the environment. This approach speeds up the learning process while also reducing the chances of divergence or instability of the RL during training [35]. 


This paper covers both the aspects of indoor positing  and path tracking control  for robotic scaled vehicles through the following novel contributions:

\textit{1)} a federated EKF (FEKF) filter, composed by two local EKFs and a master filter for fusing together their position estimates, thus improving the accuracy of the positioning system by utilising all the available onboard sensors, such as acceleration and yaw rate measurements from the IMU, lidar, and encoder. The design of the FEKF is based on a no-reset architecture where no feedback from the master filter to the local ones is used with the aim of increasing robustness to sensor faults [36]. The design of the local EKFs is based on different vehicle modelling, i.e., a dynamic bicycle model and a kinematic point  model, which are fed with different sensor outputs.
The proposed filtering system has also the capability  to  \emph{(i)} estimate vehicle side slip angles, a state not directly measured by onboard sensors but important for vehicle stability, and  \emph{(ii)} fuse sensor measurements with different sampling rates (e.g., lidar information sampled at $10$ Hz and  IMU measurements sampled at $100$ Hz) to provide to the path tracking controller accurate system states with a sampling time of $10$ ms. 

\textit{2)} The formulation and the experimental assessment of a novel DRL path tracking control solution based on an expert demonstrator. The expert's control action, generated via a linear quadratic (LQ) control strategy, is integrated into the formulation of the reward function along with reward terms considering the lateral and heading errors, and the control effort. The experimental analysis shows that the inclusion of the expert demonstrator is crucial for mitigating the simulation-to-reality gap and enabling the agent to execute paths not considered during the training phase. Furthermore, it is shown that the trained agent can outperform model-based PT solutions. Specifically, the closed-loop tracking performance of the DRL strategy is compared to those obtained via a feed-forward feedback (FF-FB) controller and two LQ strategies (i.e., the expert demonstrator and the LQ controller presented in [37] equipped with a feedforward action to achieve zero lateral error for constant reference curvatures). The closed-loop analysis is quantitatively carried out through the use of a set of key performance indicators (KPIs).

It is noted that the design of FEKF and the DRL path tracking strategy requires the use of dynamic vehicle models. The former uses vehicle models for vehicle state predictions, while the latter leverages vehicle simulation models for emulating the environment during training. Hence, in this paper also a vehicle model for robotic scaled car is formulated and experimentally validated. Specifically, it is proposed a two stage least-square (LS) identification approach for the fine-tuning of the parameters of the longitudinal and lateral vehicle dynamics. This identification procedure is simple to replicate but yet effective to generate a dynamic model for estimation and control purposes and represents and further point of novelty of the paper.


The rest of the paper is organised as follows. Section~\ref{sec:exp_setup} briefly presents the setup of the robotic car system used in this work (i.e., the QCar produced by Quanser). Sections~\ref{sec:models} and \ref{sec:model_identification_validation} present the vehicle model and the identification of its parameters, respectively.  The design of the local EKFs and the federated filter is described in Section \ref{sec:EKF_desing}, while experimental results are shown in Section \ref{sec:EKF_validation}. The DRL-based path tracking control framework is described in Section~\ref{sec:DRL_desing}. Experimental results and comparisons of the proposed DRL controller with respect to classical and LQ-based control algorithms are discussed in Section~\ref{sec:DRL_validation}. Finally, conclusions are drawn in Section~\ref{Sec:conclusion}.

\section{Experimental Setup} \label{sec:exp_setup}
\begin{figure}[t]
    \centering
    \includegraphics[width=0.4\textwidth]{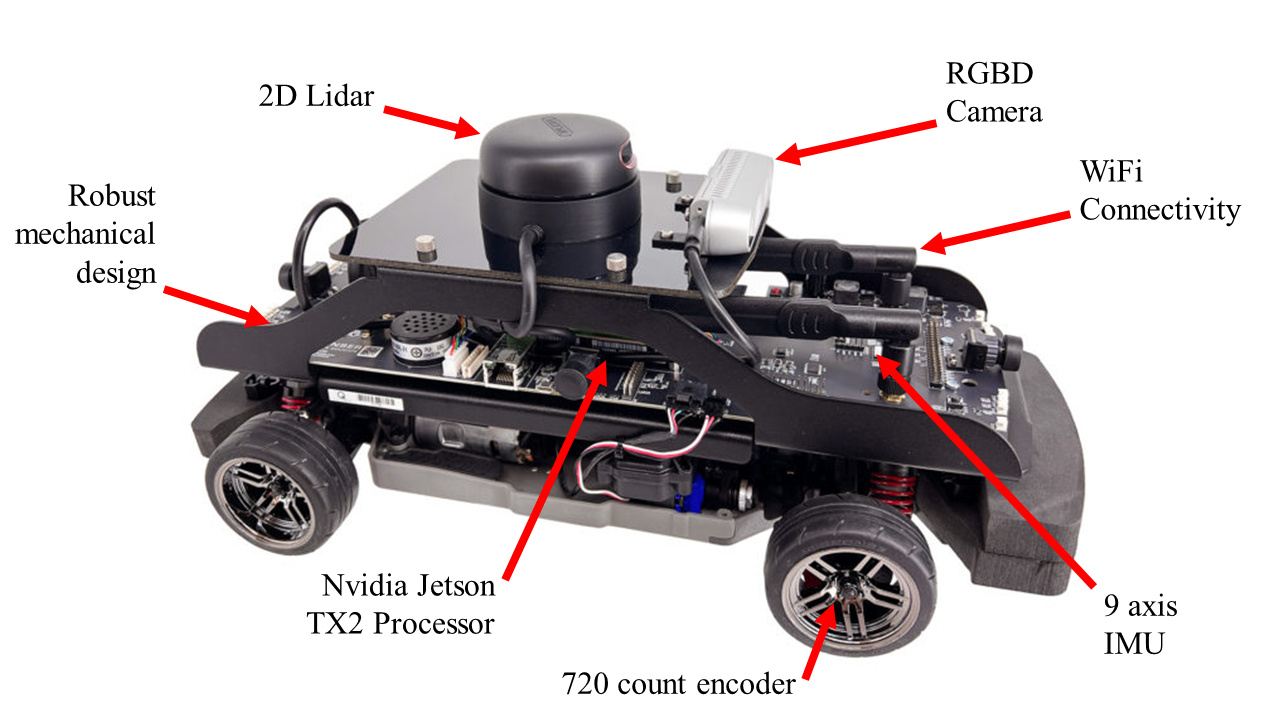}
    \caption{Quanser QCar picture with sensors [38]}
    \label{fig:Qcar}
    \vspace{-3mm}
\end{figure}

\begin{figure}[t]
    \centering
    \includegraphics[width=0.4\textwidth]{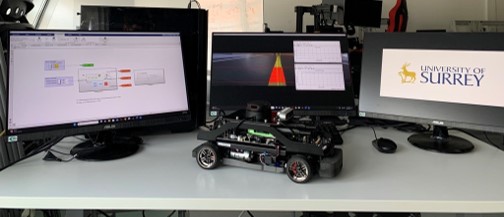}
    \caption{Base station and QCar used as experimental platform.}
    \label{fig:Qlab}
    \vspace{-3mm}
\end{figure}

The robotic scaled car used to validate the system identification procedure, the state estimator, and the DRL-based path tracking strategy is the Quanser self-driving car (QCar) platform shown in Fig.~\ref{fig:Qcar} [38]. The QCar is an open-architecture, scaled model car designed for academic research. The robotic car is an all-wheel-drive (AWD) configuration, actuated by a direct current (DC) electric motor, which has its rotational speed measured by an encoder, and operates with a single transmission ratio gearbox. The steering actuation is applied by a servo motor, which guarantees a maximum steering angle of about ${\pm 30}$ deg.  Moreover, it is equipped with an onboard 9-axis IMU and a lidar. The  onboard computing system is an NVIDIA Jetson TX2 processor which uses a Quad-Core ARM Cortex-A57 microcontroller. The QCar communicates through WiFi with the ground station in Fig.~\ref{fig:Qlab}, where the proposed algorithms are developed using Matlab 2022b  [39].

\begin{figure}[t]
    \centering
    \includegraphics[width=0.4\textwidth]{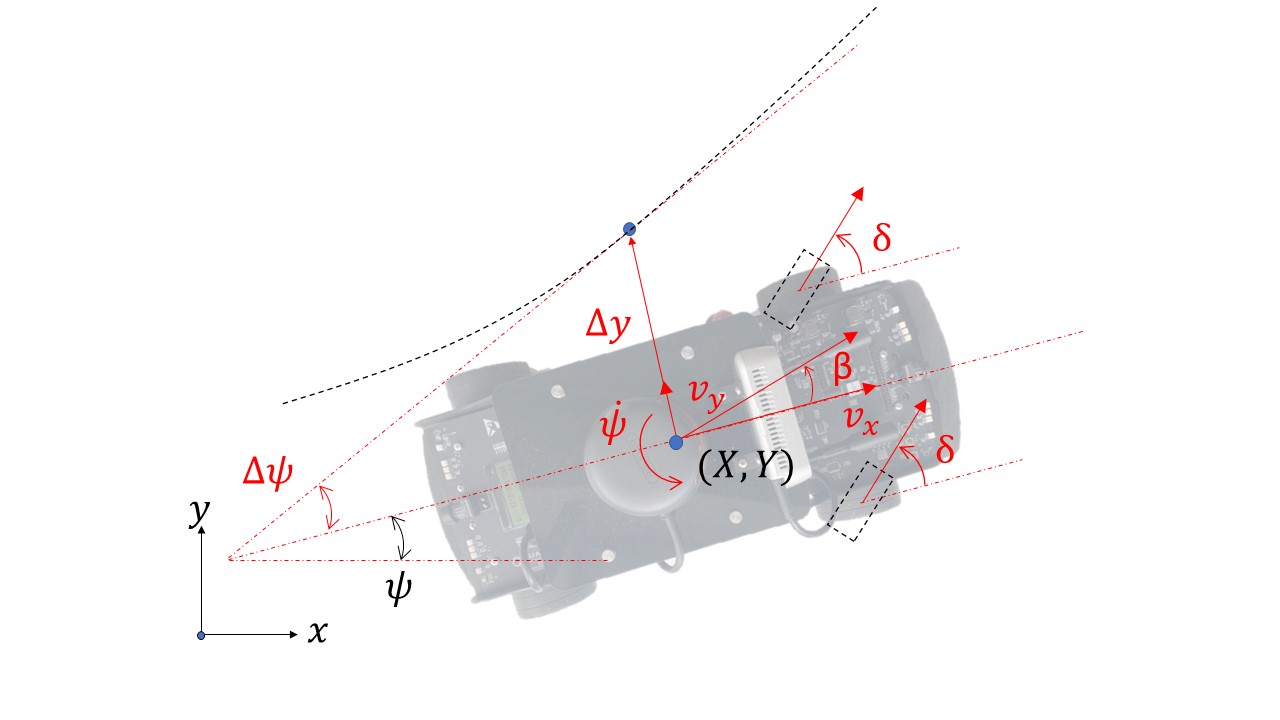}
    \vspace{-5mm}
    \caption{Reference path  and relative position and heading errors.}
    \label{fig:path_errors_schematic}
\end{figure}

\section{Vehicle Model} \label{sec:models}
This section presents vehicle models used or both the FEKF positioning system and the path tracking controller. It details \emph{(i)} the longitudinal vehicle dynamics; \emph{(ii)} the lateral vehicle dynamics based on a dynamic bicycle model (BM); and \emph{(iii)} a point kinematic model. 
Fig.~\ref{fig:path_errors_schematic} shows a schematic of the top view of the robotic car along with key vehicle variables and the adopted sign conventions.

\subsection{Longitudinal Dynamics}\label{sec:long_dyn}
Assuming negligible tyre-road longitudinal tyre slip, small values of sideslip angle and negligible electric dynamics of the DC motor in comparison to its mechanical aspect, the longitudinal vehicle speed model is:
\begin{subequations}\label{eq:long_dynamics}
    \begin{align}
\dot \omega  &= {P_1}{V_a} - {P_2} \; \omega  - {P_3} \; {\mathop{\rm sgn}\!} \left( \omega  \right), \label{eq:motor_dyn}\\
      v &= \frac{\mathcal{G}}{{{R_w}}}\omega , \label{eq:omega_2_vx}
    \end{align}
\end{subequations}
where $v$ is the vehicle speed, $P_1=\left(K_tK_v/R_a +B \right)/J_w$, $P_2=K_t/(J_wR_a)$ and $P_3=T_c/J_w$ are lumped parameters for the DC motor in which ${J_w}$ is the equivalent inertia, ${R_a}$ is the armature resistance, $B$ is the viscous friction coefficient, ${K_t}$ is the the torque constant, ${K_v}$ is the motor back-emf constant,  ${T_c}$ is the the Coulomb friction,  ${V_a}$ is the armature voltage applied to the DC motor, $\mathcal{G}$ is the transmission ratio, ${R_w}$ is the wheel radius and $\omega$ is the motor angular speed. 

\subsection{Lateral Dynamics} \label{sec:lat_dyn}
By considering a dynamic BM and assuming small sideslip angle values, the lateral dynamics equations are defined as:

\begin{subequations}\label{eq:lat_dynamics}
\footnotesize{
    \begin{align}
mv\dot \beta  &=  - ({C_f} + {C_r} + m\dot v)\beta  - \left( {\frac{{{C_f}{L_f}}}{v} - \frac{{{C_r}{L_r}}}{v} + mv} \right)r + {C_f}\delta, \label{eq:beta_dyn}\\
         {I_G}\dot r &=  - ({C_f}{L_f} - {C_r}{L_r})\beta  - \left( {\frac{{{C_f}L_f^2}}{v} + \frac{{{C_r}L_r^2}}{v}} \right)r + {C_f}{L_f}\delta,\label{eq:r_dyn}
    \end{align}
    }
\end{subequations}

\noindent where $\beta$ is the vehicle sideslip angle, $\delta$ is the steering angle, and $r=\dot{\psi}$ is the yaw rate with $\psi$ being the heading angle. 
The system parameters are the yaw mass moment of inertia $I_G$, the vehicle mass $m$, the front and rear semi-wheelbases, $L_f$ and $L_r$, and the front and rear axle cornering stiffness $C_f$ and $C_r$, respectively.

The Qcar centre of gravity (CoG) coordinates in the inertial frame, i.e., $X$ and $Y$, are then obtained as: 
\begin{subequations}\label{eq:bm_x_y_dyn}
    \begin{align}
        \dot{X} &= v \cos(\beta + \psi), \label{eq:bm_x_dyn}\\
        \dot{Y} &= v \sin(\beta + \psi),\label{eq:bm_y_dyn}
    \end{align}
\end{subequations}


\subsection{Kinematic Point Model}\label{sec:KP_model}
When the longitudinal and lateral components of the CoG acceleration in the inertial frame, $a_{x,CG}$ and $a_{y,CG}$, are known, the vehicle motion can be described as:
\begin{subequations}\label{eq:imu_x_y_dyn}
    \begin{align}
        \ddot{X} &= a_{x,CG}, \label{eq:imu_x_dyn}\\
        \ddot{Y} &= a_{y,CG}.\label{eq:imu_y_dyn}
    \end{align}
\end{subequations}

Both acceleration components are linked to the ones measured by the IMU in the vehicle body frame, $a_x$ and $a_y$, as follows:

\begin{equation}\label{eq:rotation_acc}
\footnotesize
\left[ {\begin{array}{*{20}{c}}
{{a_{x,CG}}}\\
{{a_{y,CG}}}
\end{array}} \right] = R\left( \psi  \right)\left[ {\begin{array}{*{20}{c}}
{{a_x}}\\
{{a_y}}
\end{array}} \right],\;\;R\left( \psi  \right) = \left[ {\begin{array}{*{20}{c}}
{\cos (\psi )}&{ - \sin (\psi )}\\
{\sin (\psi )}&{\cos (\psi )}
\end{array}} \right],
\end{equation} 
where $R(\psi)$ is the rotation matrix about the z-axis.


\section{Parameters Identification and Vehicle Model Experimental validation}\label{sec:model_identification_validation}
This section presents the methodologies for the tuning of the parameters $P_1$, $P_2$, $P_3$, $C_r$, and $C_f$ of the models \eqref{eq:motor_dyn} and \eqref{eq:lat_dynamics} along their experimental validation. The remaining parameters, i.e., $\mathcal{G}$, $m$, $R_w$, $I_G$, $L_r$ and $L_f$ are assumed known and equal to the nominal values provided by the robotic car manufacturer [38].  

\subsection{Longitudinal Dynamics}\label{sec:long_dyn_validation}
Assuming the car moves forward, model \eqref{eq:motor_dyn} is a first order linear system. Therefore, its dynamics are characterised by the DC-gain and the system time constant. For a step variation of the armature voltage, the steady-state motor speed $\omega_s$ is: 
\begin{equation}\label{eq:omega_ss}
\omega_s= m_lV_a - b_l,
\end{equation}
where $m=P_1/P_2$ and $b=P_3/P_2$.

The optimal parameters $m_l^\star$ and $b_l^\star$ of the linear regression model \eqref{eq:omega_ss} have been identified through a LS method by using as identification data set $\mathcal{I}_\omega=\left\{(V_{a,i}, \omega_{s,i})\right\}_{i=1,\ldots 5}$, where the set of armature voltages is $\left\{V_{a,i}\right\}_{i=1,\ldots 5}=\left\{0.75; 1.25; 1.5; 2; 2.5\right\}$ V, and  $\left\{\omega_{s,i}\right\}_{i=1,\ldots 5}$ is the set of the corresponding steady-state speed. 

\begin{figure}[t]
    \centering
    \includegraphics[width=0.4\textwidth]{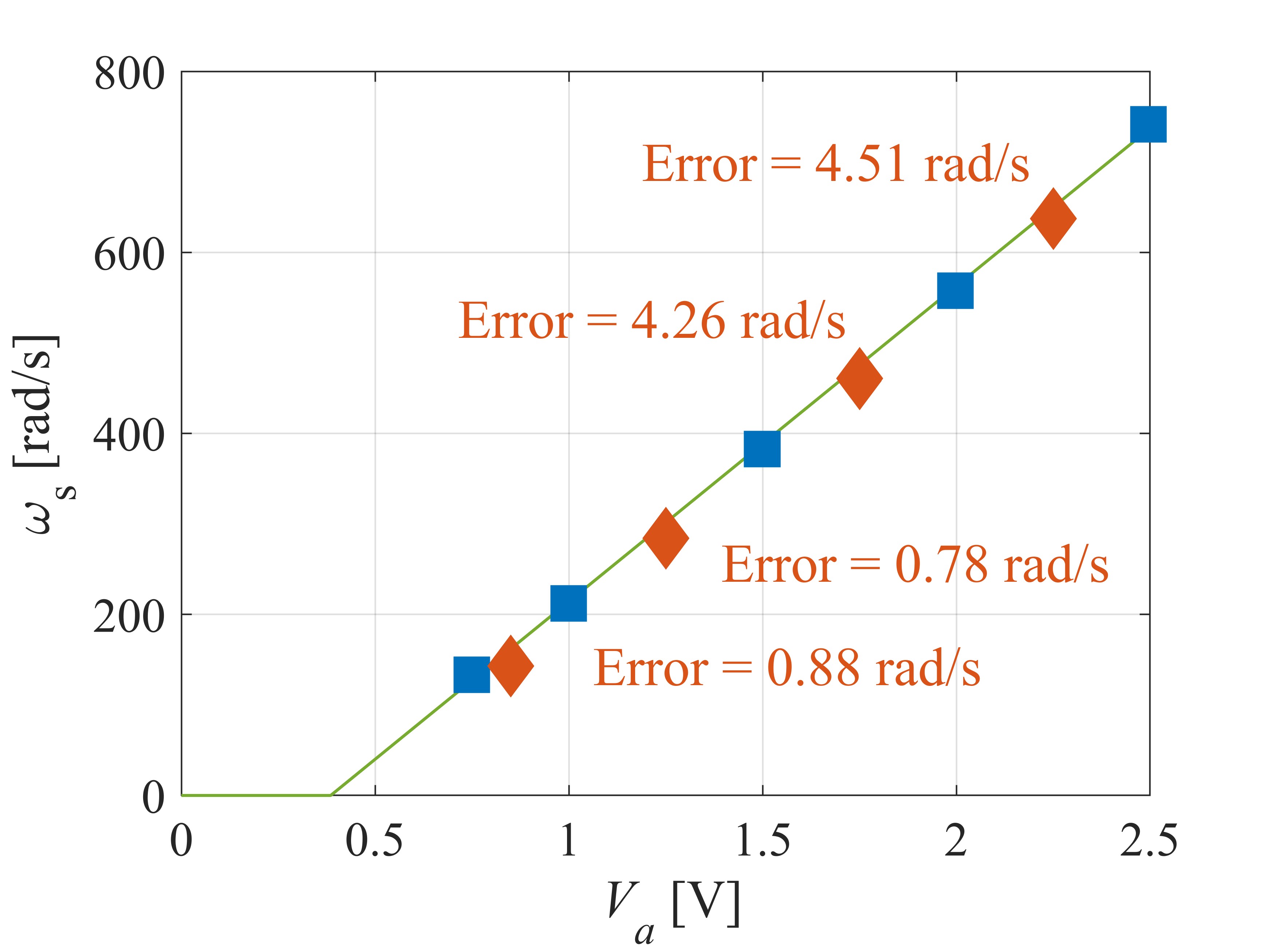}
    \caption{Experimental validation of the steady-state longitudinal dynamics: identification data set (blue square), validation data set (red diamond), and optimised steady-state speed regression model (green solid line).}
    \label{fig:steady_state_omega}
\end{figure}

Fig.~\ref{fig:steady_state_omega} shows the identification data set $\mathcal{I}_\omega$, the regression model \eqref{eq:omega_ss} with the optimal parameters along with a validation data set $\mathcal{V}_\omega=\left\{(V_{a,j}, \omega_{s,j})\right\}_{j=1,\ldots 4}$ obtained for armature voltages not included in the identification data set, i.e., $\left\{V_{a,j}\right\}_{j=1,\ldots 4}=\left\{1.1; 1.35;  1.85; 2.25\right\}$ V. Fig.~\ref{fig:steady_state_omega} confirms that the optimised regression model \eqref{eq:omega_ss} well predicts the steady-state speed with a maximum error of 4.51 rad/s.

The parameter $P_2$ is related to the system time constant and it has been selected to minimise the mismatch between the transient response of system \eqref{eq:motor_dyn} and the experimental step response for different steps of the armature voltage. Specifically, the optimal $P_2$-parameter, denoted as $P_2^\star$, is selected as:
\begin{subequations}\label{eq:P2_opt}
    \begin{align}
    P_2^\star &= \mathop {\arg \min}\limits_{{P_2} \in \left[ {P_2^{\mathop{\rm m}\nolimits} ,P_2^M} \right]} 
    {\sum\limits_{j = 1}^{{M_s}} {\frac{1}{{{T_j}}}\int_0^{{T_j}} {{{\left( {{\omega _j}(t) - \widehat \omega (t;{P_2})} \right)}^2}dt} } } 
    , \label{eq:P2_cost}\\
      & \text{s.t.} \;\;{P_1} = {m_l^\star}{P_2}\;\;{\rm{and}}\;\;{P_3} = {b_l^\star}{P_2}\label{eq:P2_constr},
    \end{align}
\end{subequations}
where $M_s$ is the number of experimental armature steps used for the identification, $T_j$ and $\omega_j$ are the duration of the transient response and experimental step response for the $j$-th experiment, $P_2^m$ and $P_2^M$ are the lower and upper limits for $P_2$, and $\widehat{\omega}(t;P_2)$ is the step response of system \eqref{eq:motor_dyn} when $P_1$ and $P_3$ are set equal to ${P_1} = {m_l^*}{P_2}$ and ${P_3} = {b_l^*}{P_2}$, respectively. The experimental speed $\omega$ is measured through the on-board encoder and filtered via an anticausal moving average filter. Furthermore, the equality constraints for the optimisation problem \eqref{eq:P2_opt} guarantee a good estimate of the system steady-state for any selection of the parameter $P_2$. The optimisation problem
has been solved numerically through the Matlab Optimization Toolbox [39] when the cost function \eqref{eq:P2_cost} has been discretised with a sampling time of $t_s=10$ ms.

\begin{figure}[t]
     \centering
     \subfloat[]
      {\includegraphics[width=0.25\textwidth]{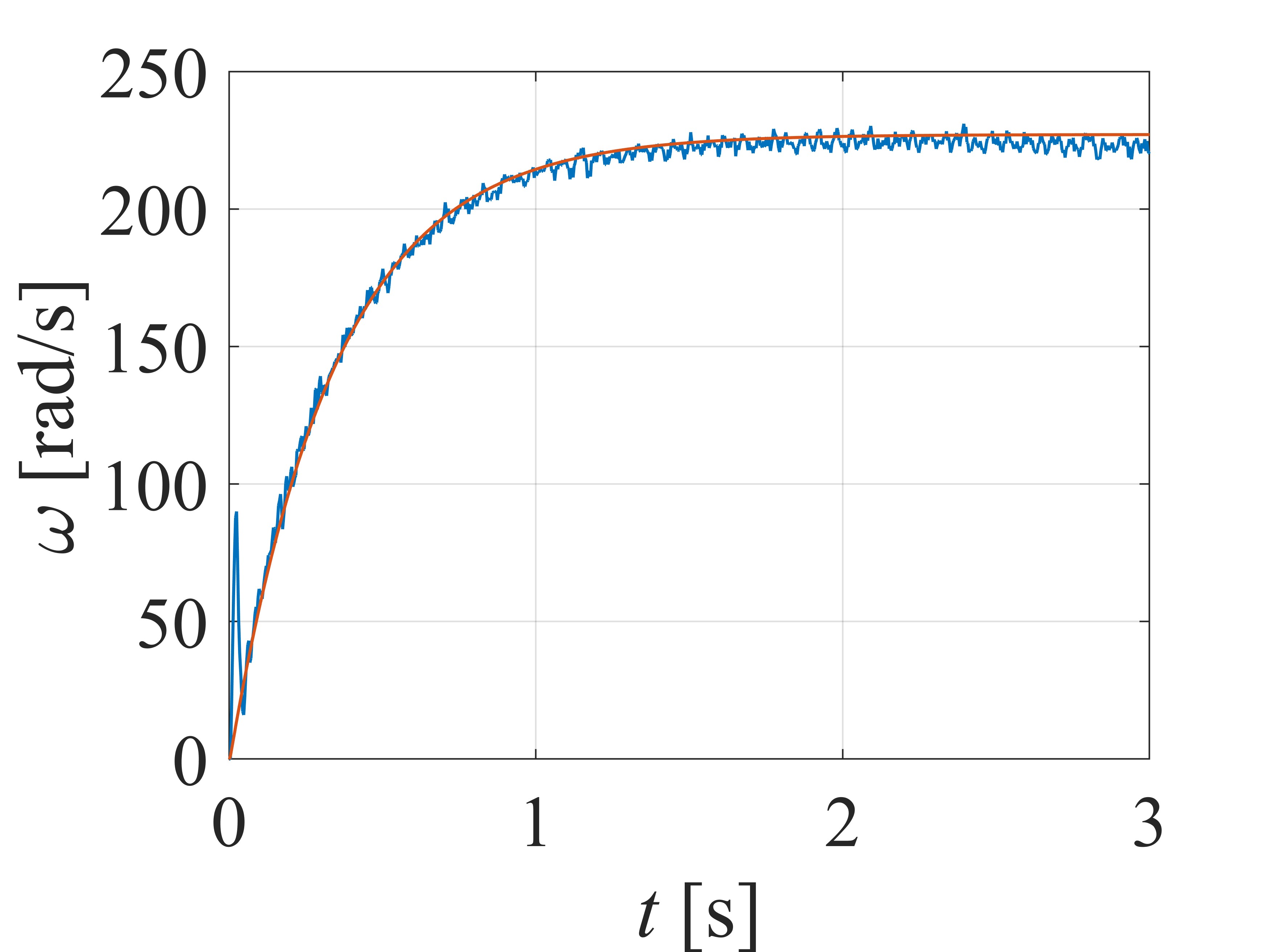}\label{fig:long_dyn_transient_ident1}} 
     \subfloat[]
     {\includegraphics[width=0.25\textwidth]{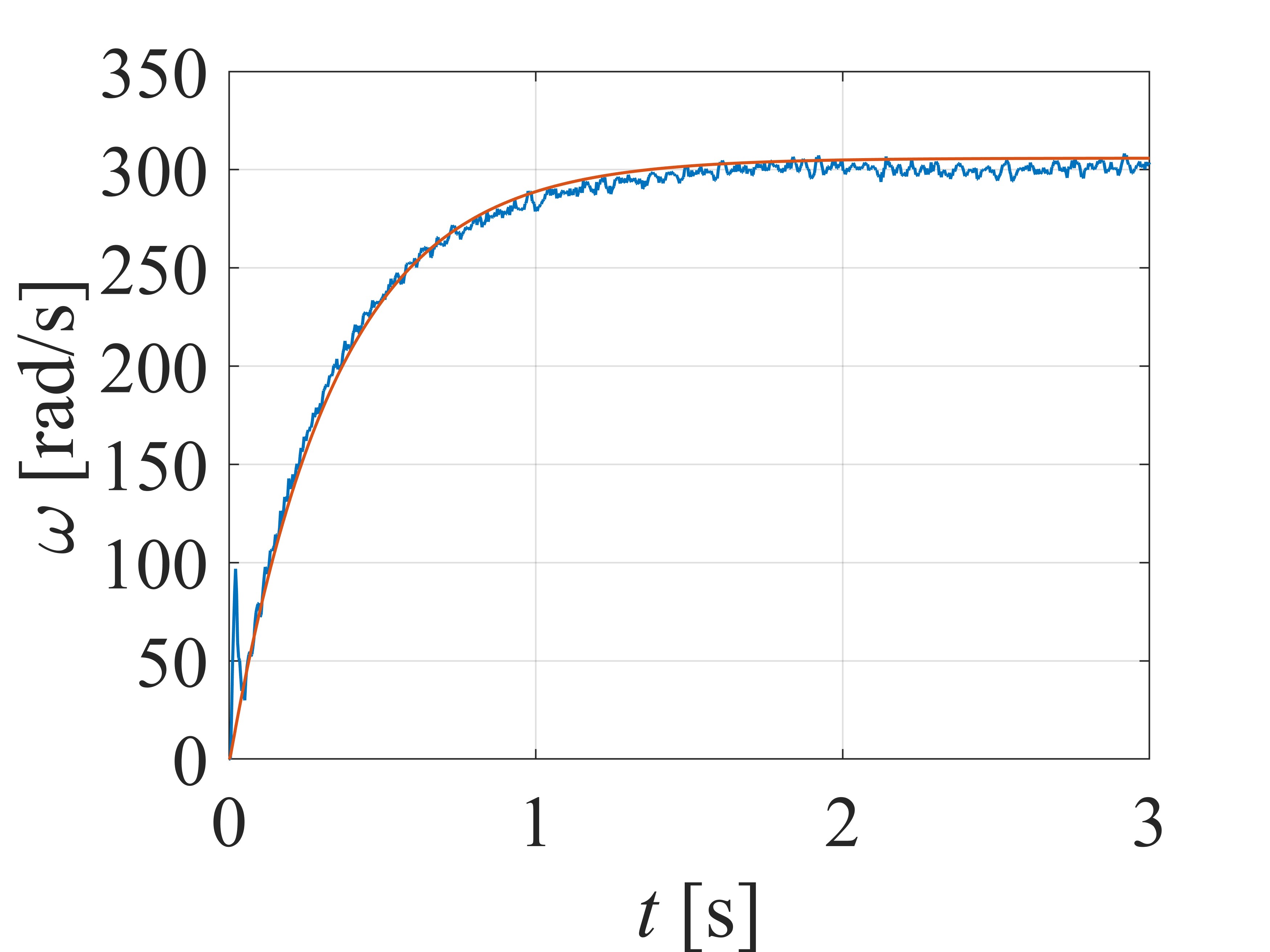}\label{fig:long_dyn_transient_ident2}}\\
          \subfloat[]
     {\includegraphics[width=0.25\textwidth]{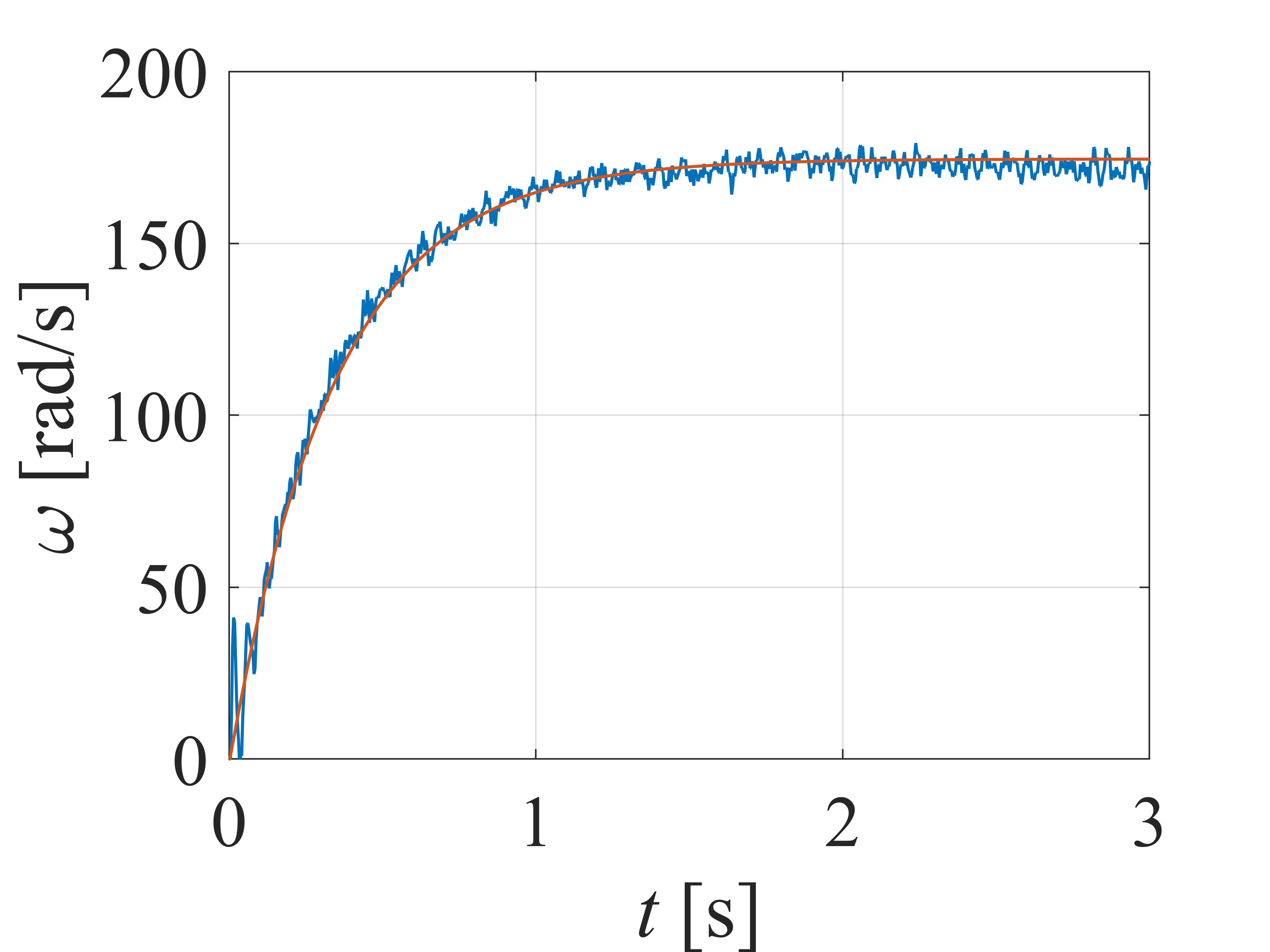}\label{fig:long_dyn_transient_valid1}}
         \subfloat[]{\includegraphics[width=0.25\textwidth]{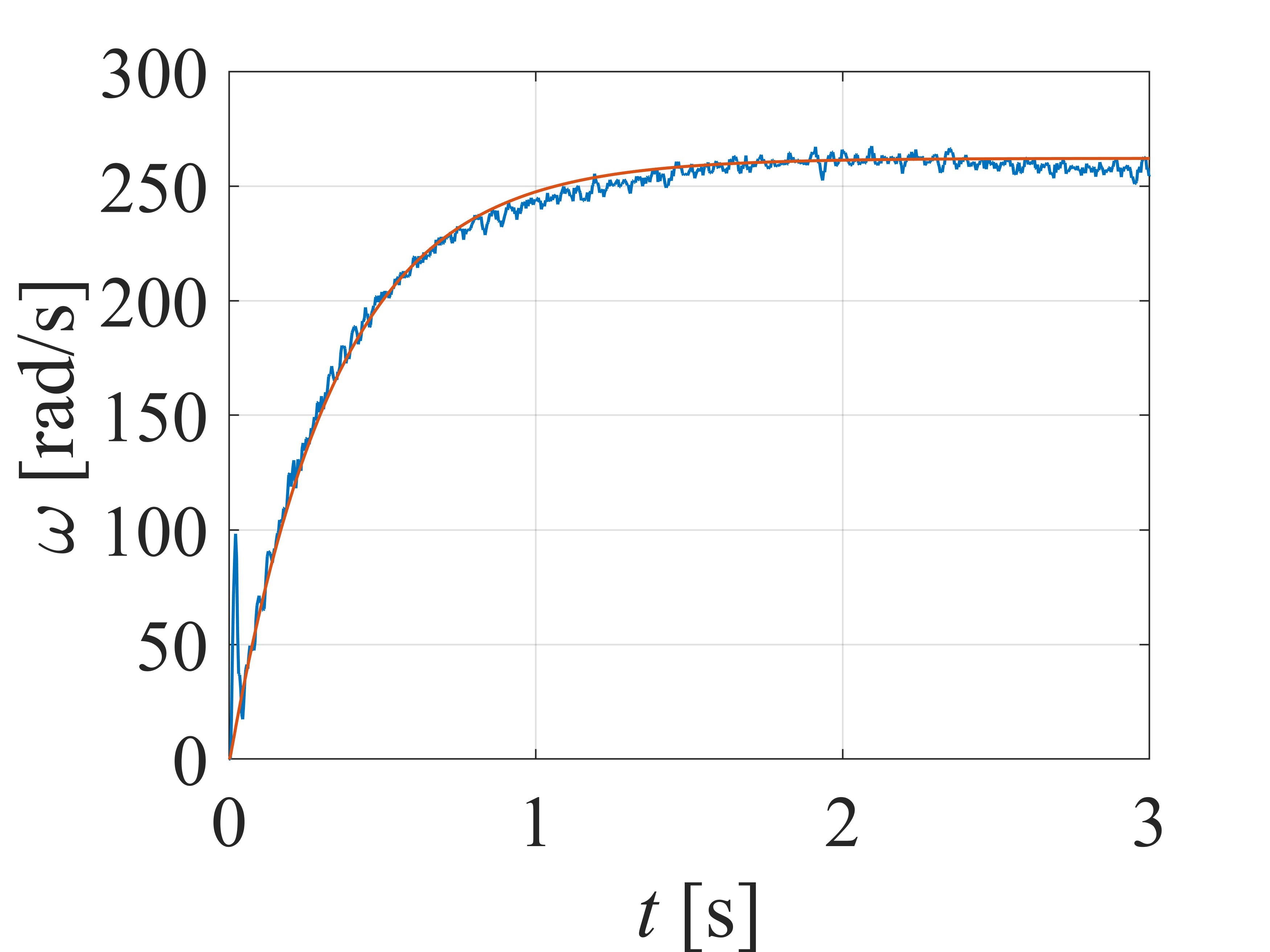}\label{fig:long_dyn_transient_valid2}}
     \caption{Experimental validation of the transient longitudinal dynamics: experimental data (blue solid line) and response of the optimised model \eqref{eq:motor_dyn} (red solid line) when the step inputs are (a) $V_a=$ 1.25 V , (b) $V_a=$ 1.5 V , (c) $V_a=$ 1.1 V, and (d) $V_a=$ 1.35 V.}
\label{fig:long_dyn_transient}
\end{figure}

Fig.~\ref{fig:long_dyn_transient} shows the ability of the optimised model \eqref{eq:motor_dyn} to capture the  transient experimental speed dynamics for different step inputs belonging either to the identification data set used for problem \eqref{eq:P2_opt} (Figs.~\ref{fig:long_dyn_transient_ident1} and \ref{fig:long_dyn_transient_ident2}) or  the validation data set (Figs.~\ref{fig:long_dyn_transient_valid1} and \ref{fig:long_dyn_transient_valid2}). The average percentage root mean square errors (RMSEs) computed for both identification and validation trajectories are 1.82\% and 1.88\%, respectively, thus confirming a good match between experimental data and model predictions.

\subsection{Lateral Dynamics}\label{sec:lat_dyn_validation}
\begin{figure}[t]
    \centering
    \includegraphics[width=0.4\textwidth]{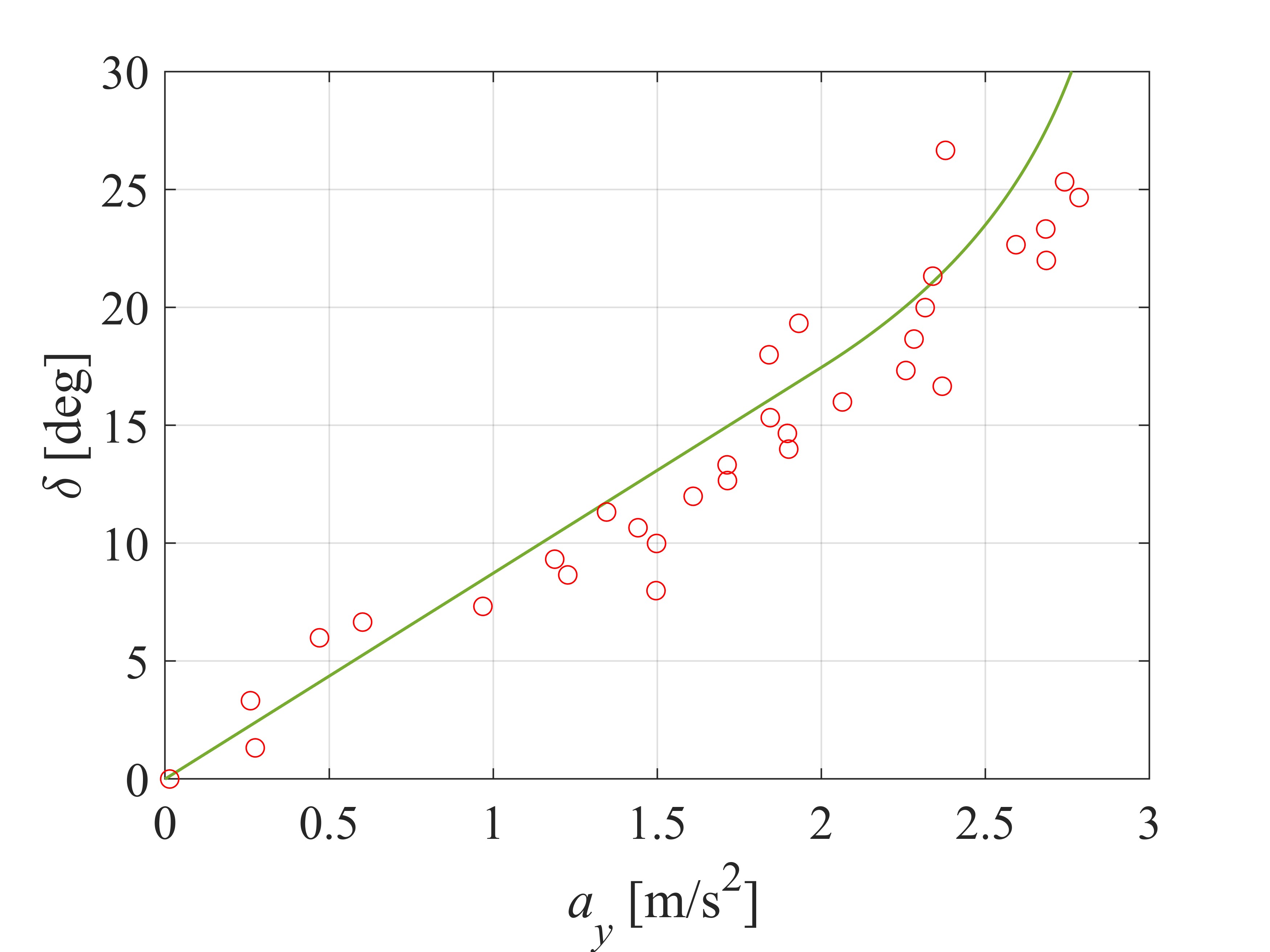}
    \caption{Experimental understeer characteristic.}
    \label{fig:under_steering}
\end{figure}

Fig.~\ref{fig:under_steering} shows the experimentally nonlinear understeer characteristic of the robotic car obtained through a ramp steer manoeuvre executed at a constant speed $v=$ 1.5~m/s. The lateral acceleration provided by the IMU has been filtered through an anticasual moving average filter. Fig.~\ref{fig:under_steering} provides an estimate  of 2 m/s$^2$ as the maximum lateral acceleration within which the linear model in \eqref{eq:lat_dynamics} holds. 

Under the assumption of linear regime,  the steady-state lateral acceleration,  $a_{ys}$, is calculated from \eqref{eq:lat_dynamics} for step steering manoeuvres at constant speed: 
\begin{equation}\label{eq:ay_steady_state}
{a_{ys}} = \frac{1}{k_{su}}\delta, \;\;\text{with} \;\; k_{su} =\frac{{{C_f}L_f^2 + {C_r}L_r^2}}{{{C_f}{L_f}{v^2}}},
\end{equation}
where $k_{su}$ is the understeer gradient.

The optimal understeer gradient, denoted as $k_{su}^\star$, has been estimated experimentally through step-manoeuvres executed with $v=1.5$ m/s, and an LS method based on the identification data set $\mathcal{I}_a=\left\{(\delta_i, a_{ys,i})\right\}_{i=1, \ldots, 5}$, where $\left\{\delta_{i}\right\}_{i=1, \ldots, 5}=\left\{4; 7; 9; 11; 14\right\}$ deg, and $\left\{a_{ys,i}\right\}_{i=1, \ldots, 5}$~m/s$^2$ are the corresponding steady-state lateral accelerations. A proportional–integral (PI) controller has been used  to impose the desired speed over the manoeuvres.

\begin{figure}[t]
    \centering
    \includegraphics[width=0.4\textwidth]{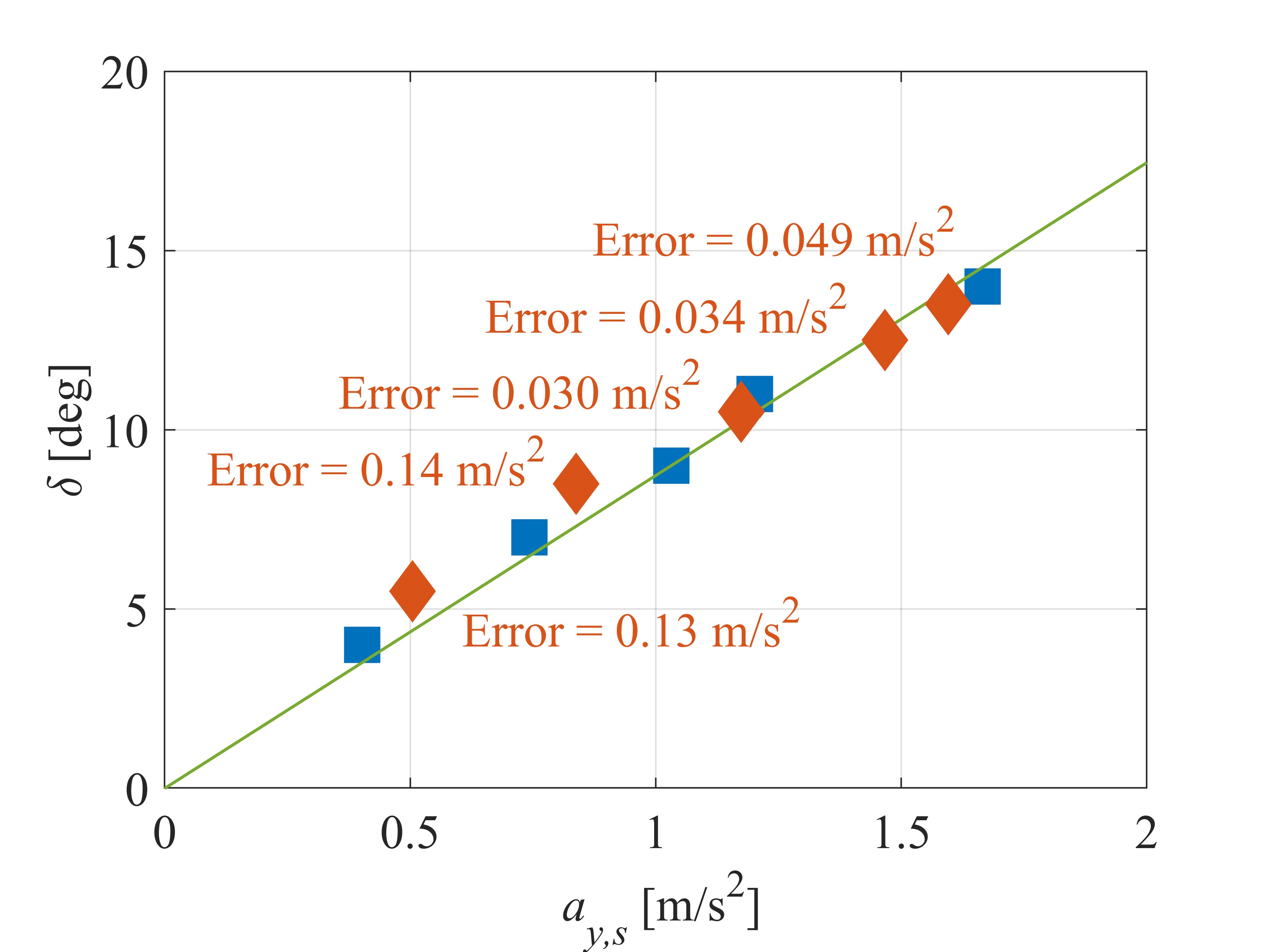}
    \caption{Experimental validation of the steady-state lateral dynamics: identification data set (blue square), validation data set (red diamond), and optimised steady-state lateral acceleration regression model (green solid line).}
    \label{fig:steady_state_ay}
\end{figure}

The ability of the optimised regression model \eqref{eq:ay_steady_state} to capture the understeer characteristic in the linear region is shown in Fig.~\ref{fig:steady_state_ay}, where the model predictions are reported along with the identification data set $\mathcal{I}_a$ and the validation data set $\mathcal{V}_a=\left\{(\delta_j, a_{ys,j})\right\}_{j=1, \ldots, 5}$, obtained for step steering inputs not included in the identification data set, i.e., $\left\{\delta_{j}\right\}_{j=1, \ldots, 5}=\left\{5.5; 8.5; 10.5; 12.5; 13.5 \right\}$ deg.

To identify $C_r$ and $C_f$ from the lumped coefficient $k_{us}^\star$,  $C_f$ has been selected to minimise the mismatch between the transient response of the lateral acceleration predicted through system \eqref{eq:lat_dynamics} and the experimental lateral acceleration obtained from the previous step steering tests. Then, the optimal $C_r$ is selected to have the understeering gradient equivalent to  $k_{us}^\star$. Specifically, the optimal $C_f$ is computed as: 

\begin{subequations}\label{eq:Cf_opt}
    \begin{align}
C_f^\star &=  \mathop {\arg \min}\limits_{{C_f} \in \left[ {C_f^{\mathop{\rm m}\nolimits} ,C_f^M} \right]} 
{\sum\limits_{j = 1}^{{N_s}} {\frac{1}{{{t_j}}}\int_0^{{t_j}} {{{\left( {{a_j}(t) - \widehat a(t;{C_f})} \right)}^2}dt} } } 
, \label{eq:Cf_cost} \vspace{2mm} \\ 
      & \text{s.t.} \;\;
\frac{{{C_f}L_f^2 + {C_r}L_r^2}}{{{C_f}{L_f}{v^2}}} = {k_{su}^\star},      
      \label{eq:Cf_constr}
    \end{align}
\end{subequations}
where $N_s$ is the experimental step-steering manoeuvres considered for the identification, $t_j$ and $a_j$ are the duration of the transient response and the experimental step lateral acceleration response for the $j$-th experiment, $C_f^\text{m}$ and $C_f^\text{M}$ are the lower and upper limits for $C_f$, and $\widehat{a}(t;C_f)$ is the lateral acceleration step response of system \eqref{eq:lat_dynamics} when $C_r$ is set as $C_r = \left( {k_{su}^*{C_f}{L_f}{v^2} - {C_f}L_f^2} \right)/L_r^2$. The equality constraint \eqref{eq:Cf_constr} guarantees a good estimate of the steady-state lateral acceleration for any selection of the parameter $C_f$. The optimisation problem \eqref{eq:Cf_opt} has been solved numerically through the Matlab Optimization Toolbox when the cost function \eqref{eq:Cf_cost} has been discretised with a sampling time of $t_s=10$ ms.

\begin{figure}[t]
     \centering
     \subfloat[]
      {\includegraphics[width=0.25\textwidth]{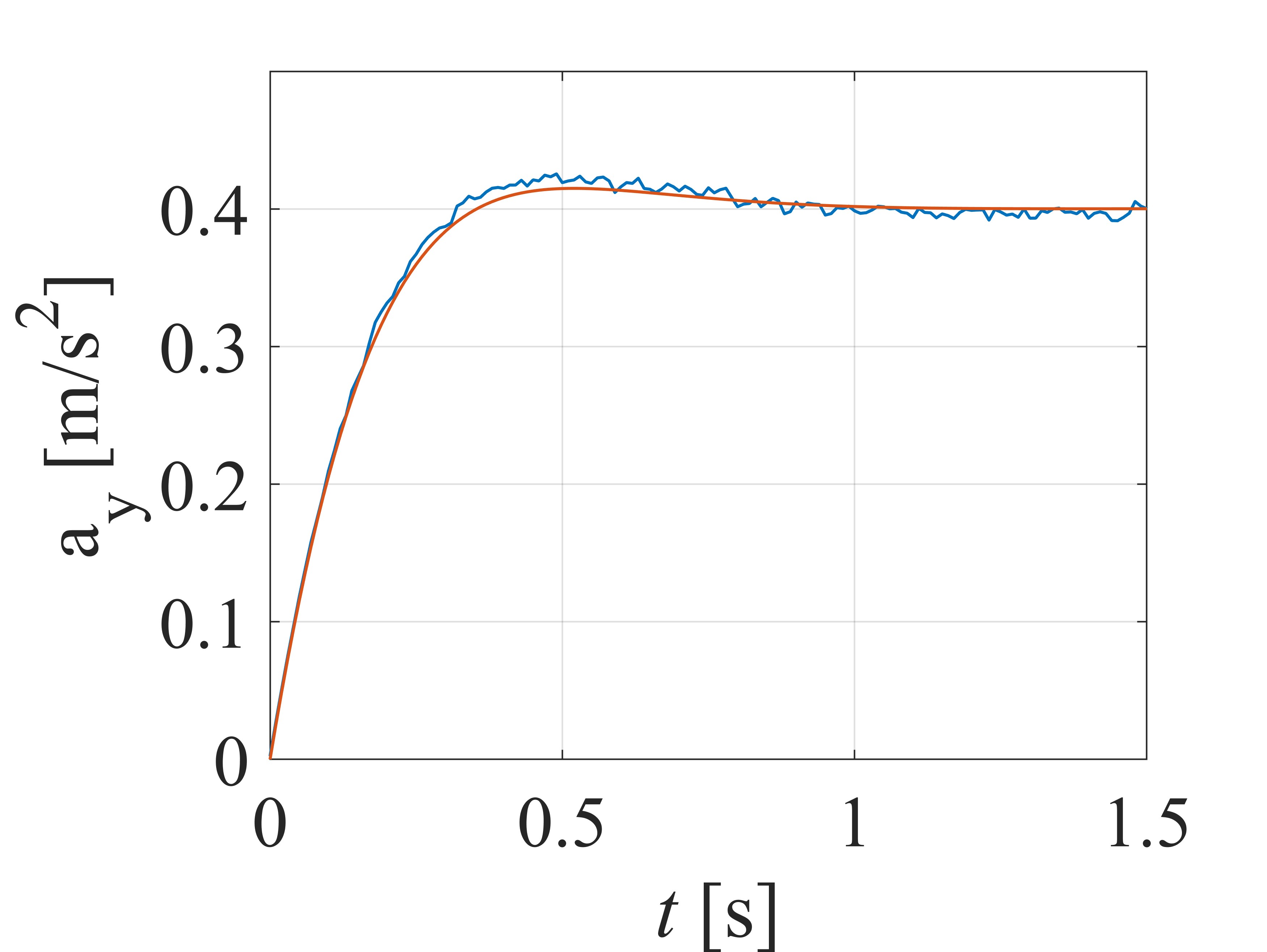}\label{fig:lat_dyn_transient_ident1}} 
     \subfloat[]
     {\includegraphics[width=0.25\textwidth]{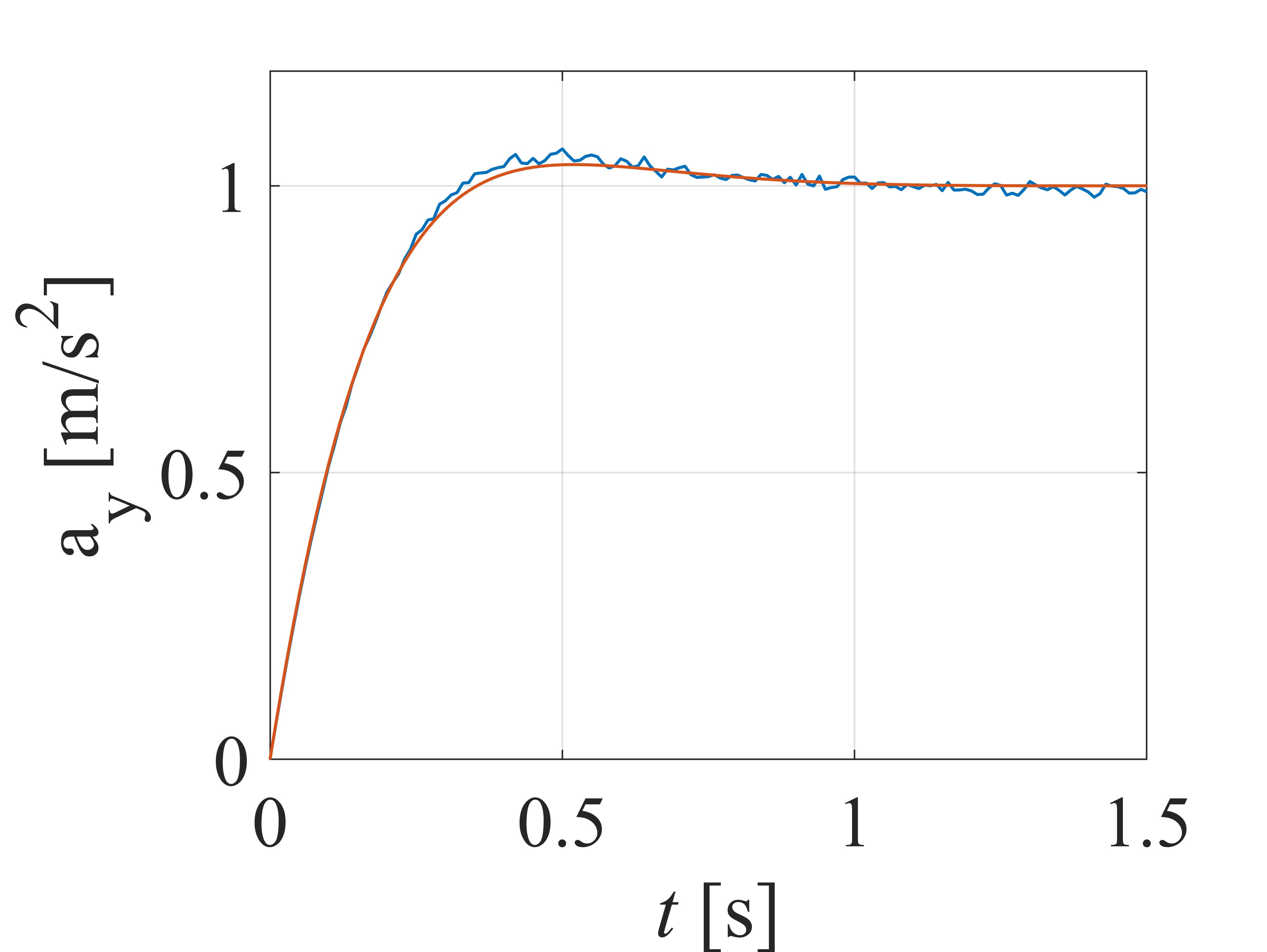}\label{fig:lat_dyn_transient_ident2}}\\
          \subfloat[]
     {\includegraphics[width=0.25\textwidth]{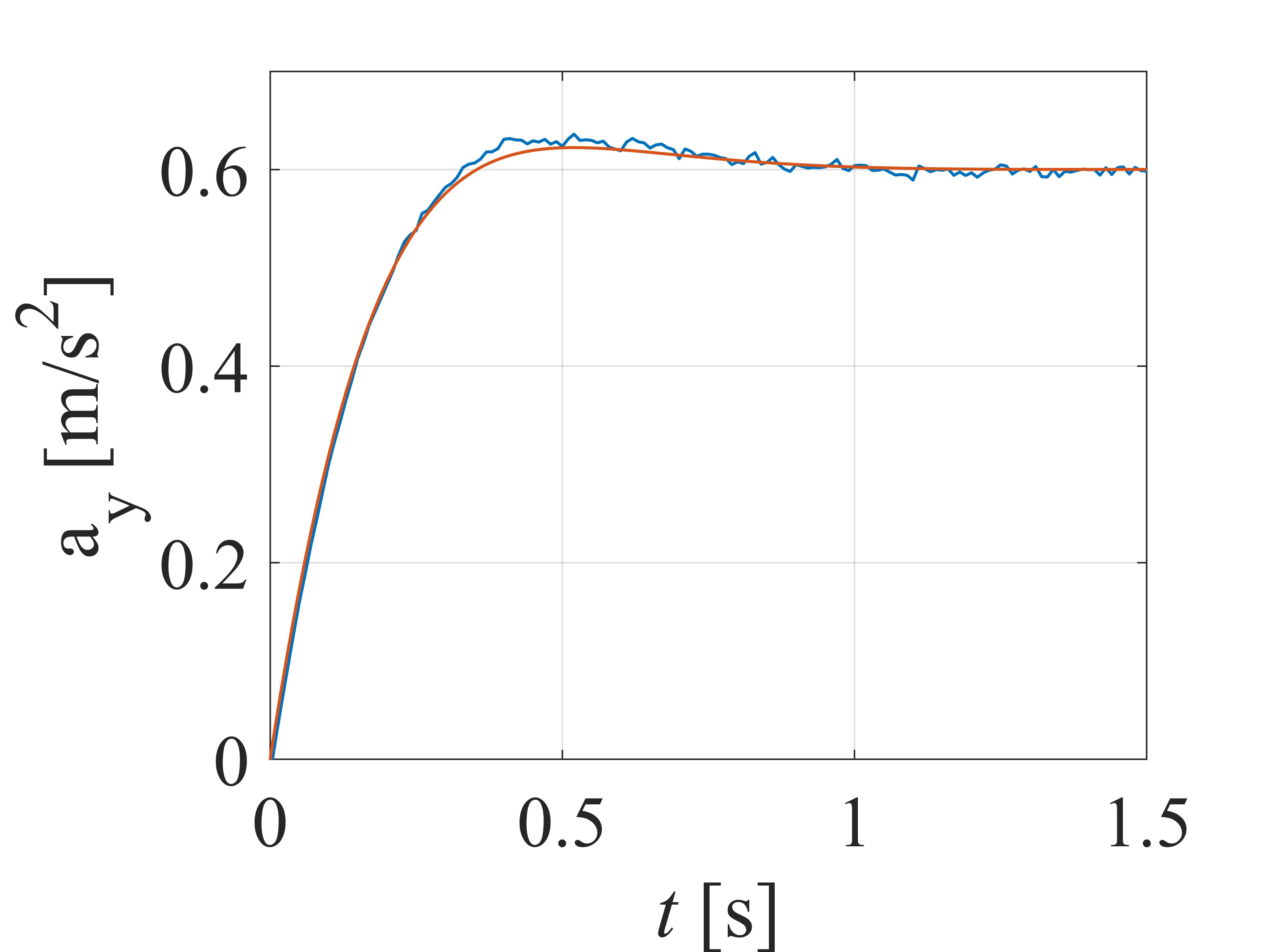}\label{fig:lat_dyn_transient_valid1}}
         \subfloat[]{\includegraphics[width=0.25\textwidth]{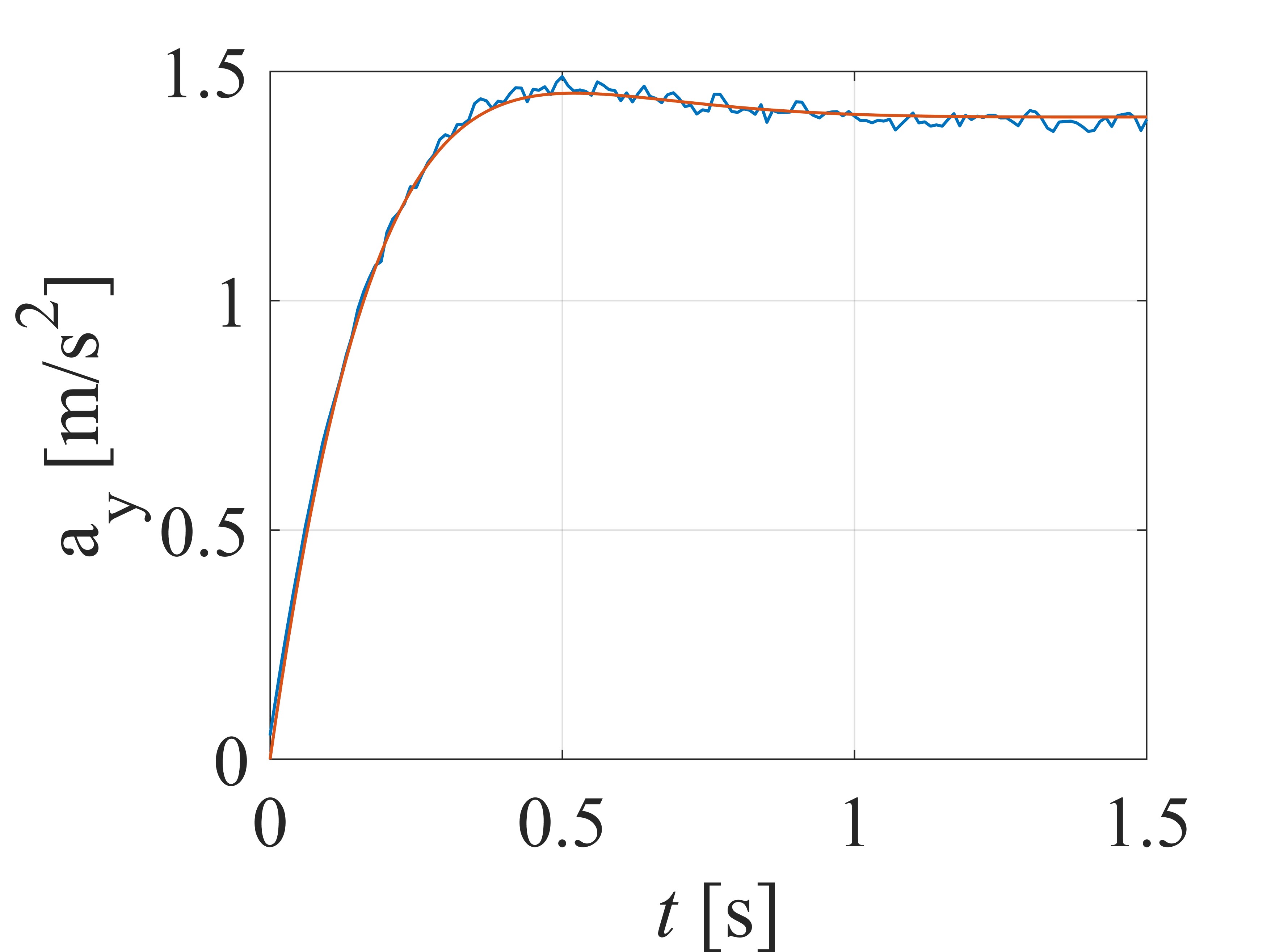}\label{fig:lat_dyn_transient_valid2}}
     \caption{Experimental validation of the lateral model during transient: experimental data (blue solid line) and response of the optimised yaw rate model \eqref{eq:lat_dynamics} (red solid line) when the step input is (a) $\delta=$ 4 deg, (b) $\delta=$ 10 deg, (c) $\delta=$ 6.5 deg and (d) $\delta=$ 13.5 deg.}
\label{fig:lat_dyn_transient}
\end{figure}

The transient response of the optimised model \eqref{eq:lat_dynamics} and the experimental lateral accelerations for step steering manoeuvres executed at $v=1.5$ m/s are shown in Fig.~\ref{fig:lat_dyn_transient} for steering inputs belonging either to the identification set (i.e., Figs.~\ref{fig:lat_dyn_transient_ident1} and \ref{fig:lat_dyn_transient_ident2}) or the validation set  (i.e., Figs.~\ref{fig:lat_dyn_transient_valid1} and \ref{fig:lat_dyn_transient_valid2}). The average RMSE percentage computed over the trajectories used for identification and validation are $6.70 \%$ and $7.33 \%$, respectively. Therefore, the optimised lateral model well captures the experimental lateral acceleration. 

\subsection{Validation of the Model}\label{sec:model_validation}
\begin{figure}[t]
     \centering
     \subfloat[]
      {\includegraphics[width=0.25\textwidth]{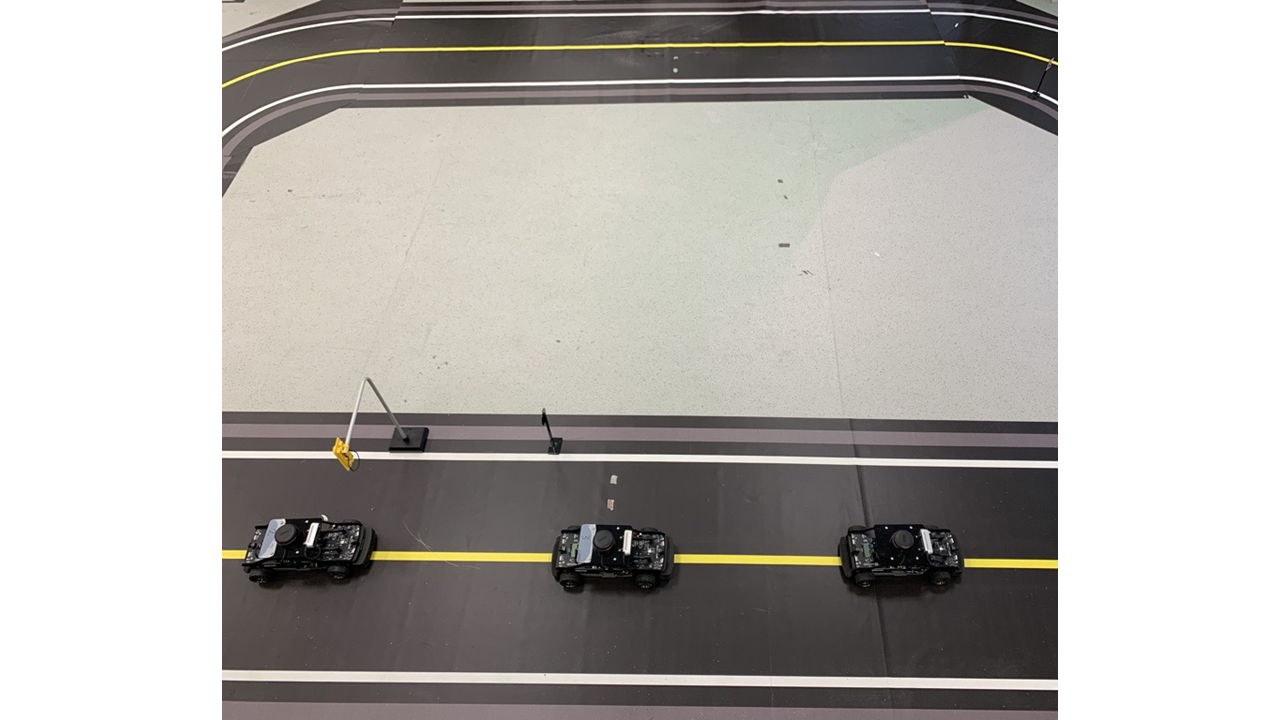}\label{fig:oval_track}} 
     \subfloat[]
     {\includegraphics[width=0.25\textwidth]{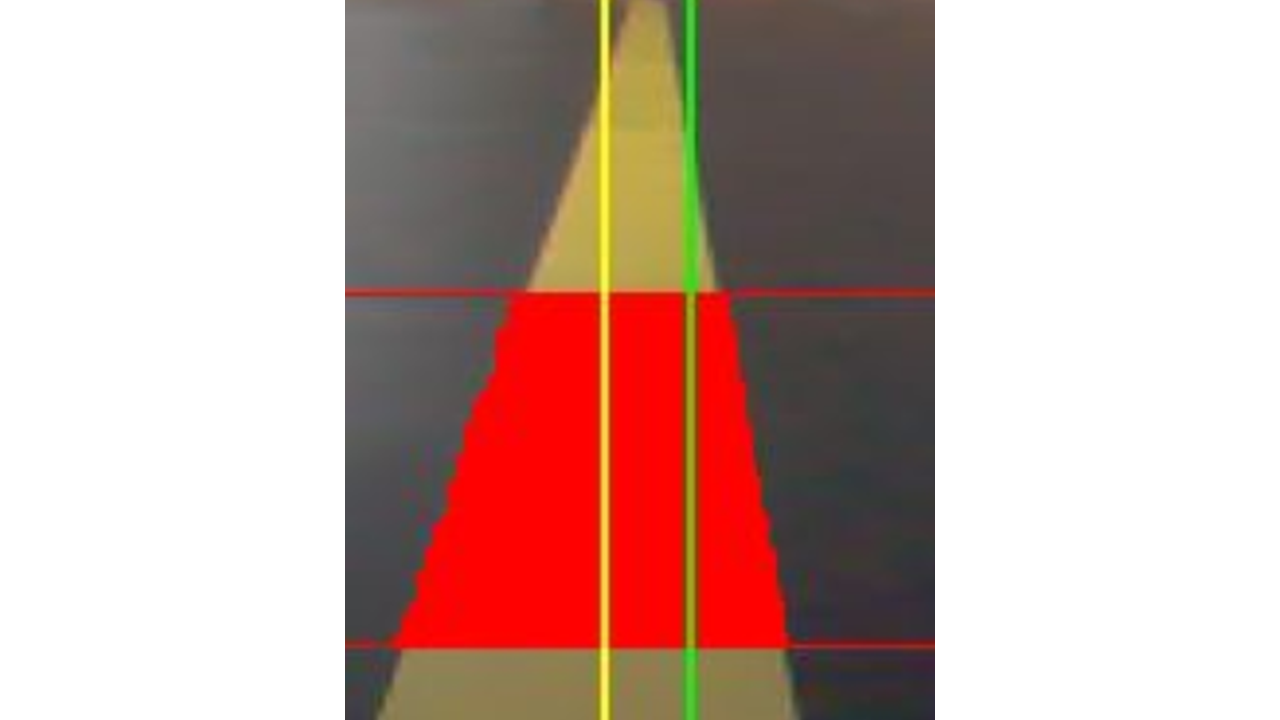}\label{fig:camera}}
     \caption{(a) Oval track and (b) camera output: detected yellow line (red patch), centre of the detected line (yellow vertical line), and vehicle heading (green vertical line).}
\label{fig:track_camera}
\end{figure}

In this section, the proposed vehicle model is further validated when the robotic car follows the oval track shown in Fig.~\ref{fig:oval_track} with a constant longitudinal speed. To this aim, the line follower control algorithm provided by the robotic car manufacturer has been used. This line keeping strategy leverages the camera for detecting the yellow line for adjusting the steering angle to reduce the mismatch between the vehicle heading and the line to follow (see also Fig.~ \ref{fig:camera} showing the detection of the line to follow). 

\begin{figure}[t]
     \centering
     \subfloat[]
      {\includegraphics[width=0.25\textwidth]{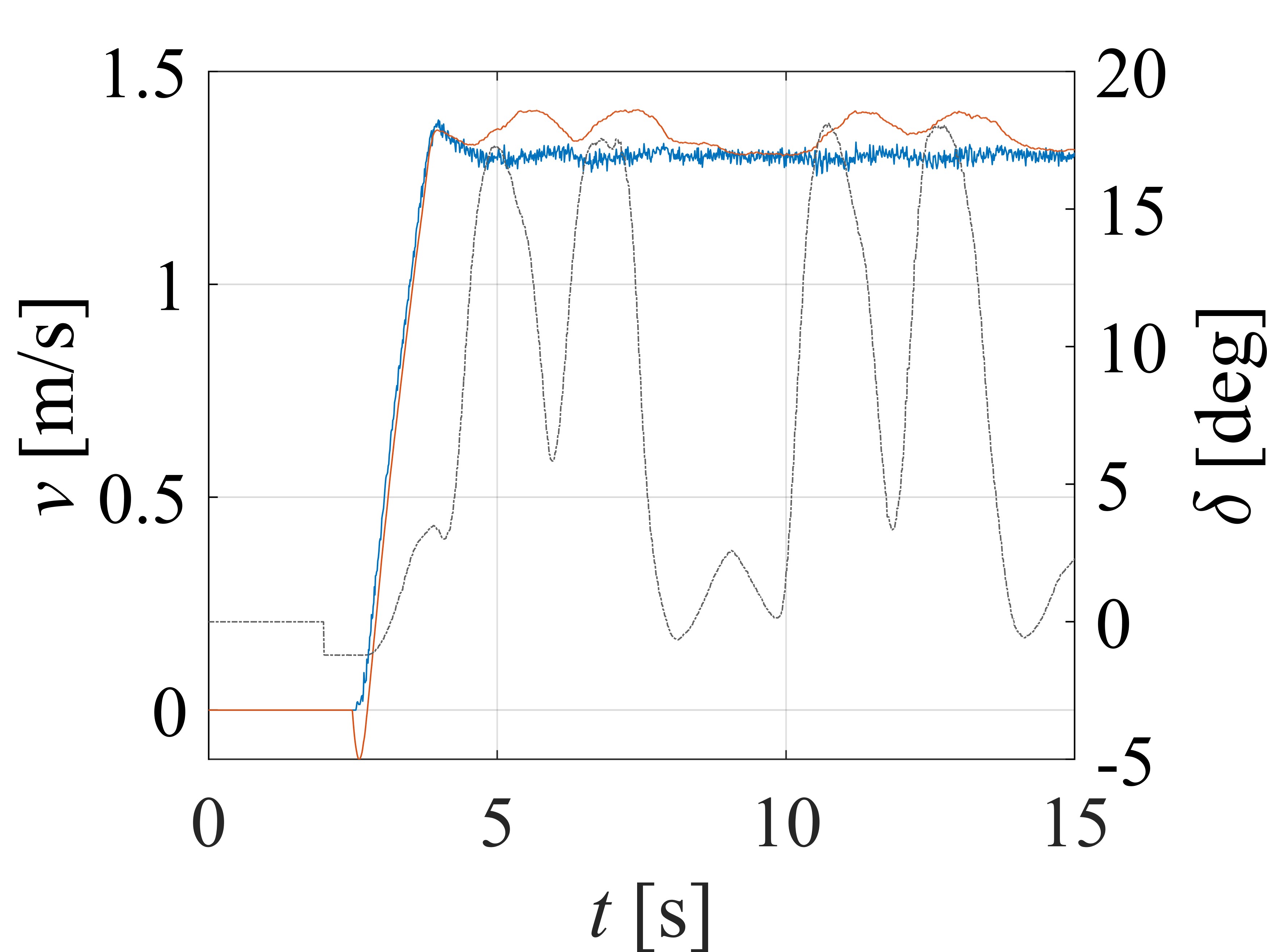} \label{fig:oval_validation_ident1}} 
     \subfloat[]
     {\includegraphics[width=0.25\textwidth]{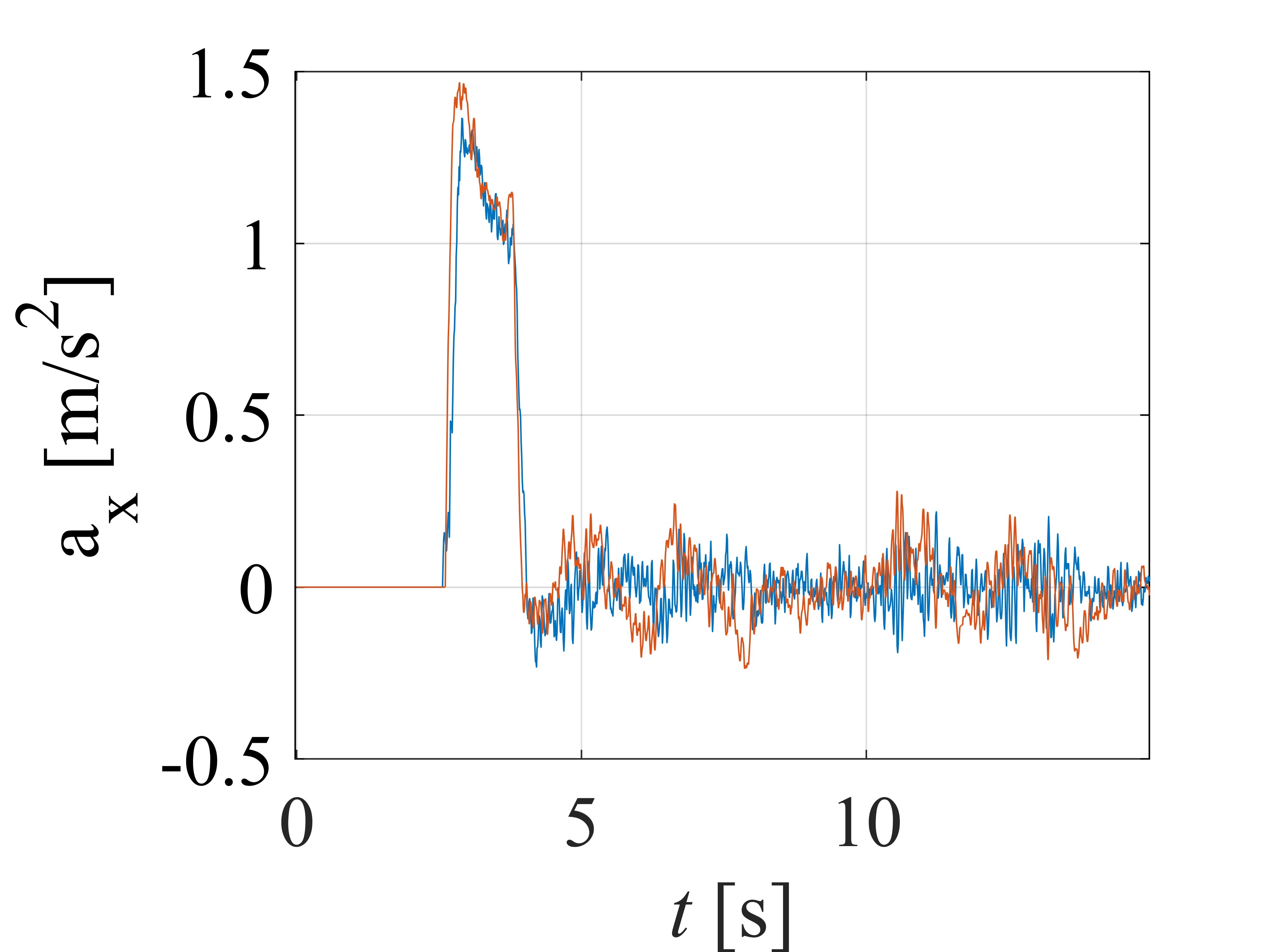}\label{fig:loval_validation_ident2}}\\
          \subfloat[]
     {\includegraphics[width=0.25\textwidth]{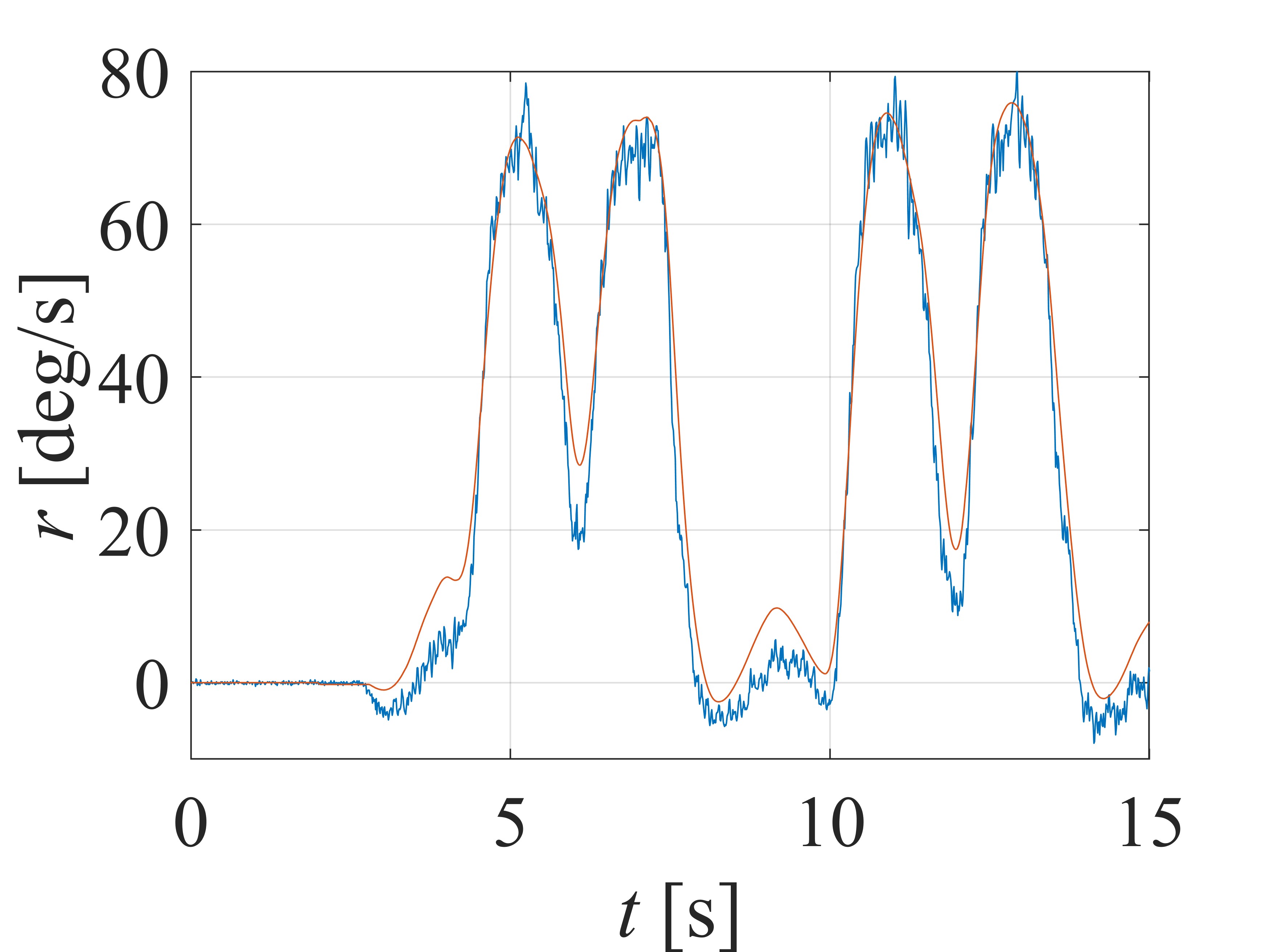}\label{fig:oval_validation_valid1}}
         \subfloat[]{\includegraphics[width=0.25\textwidth]{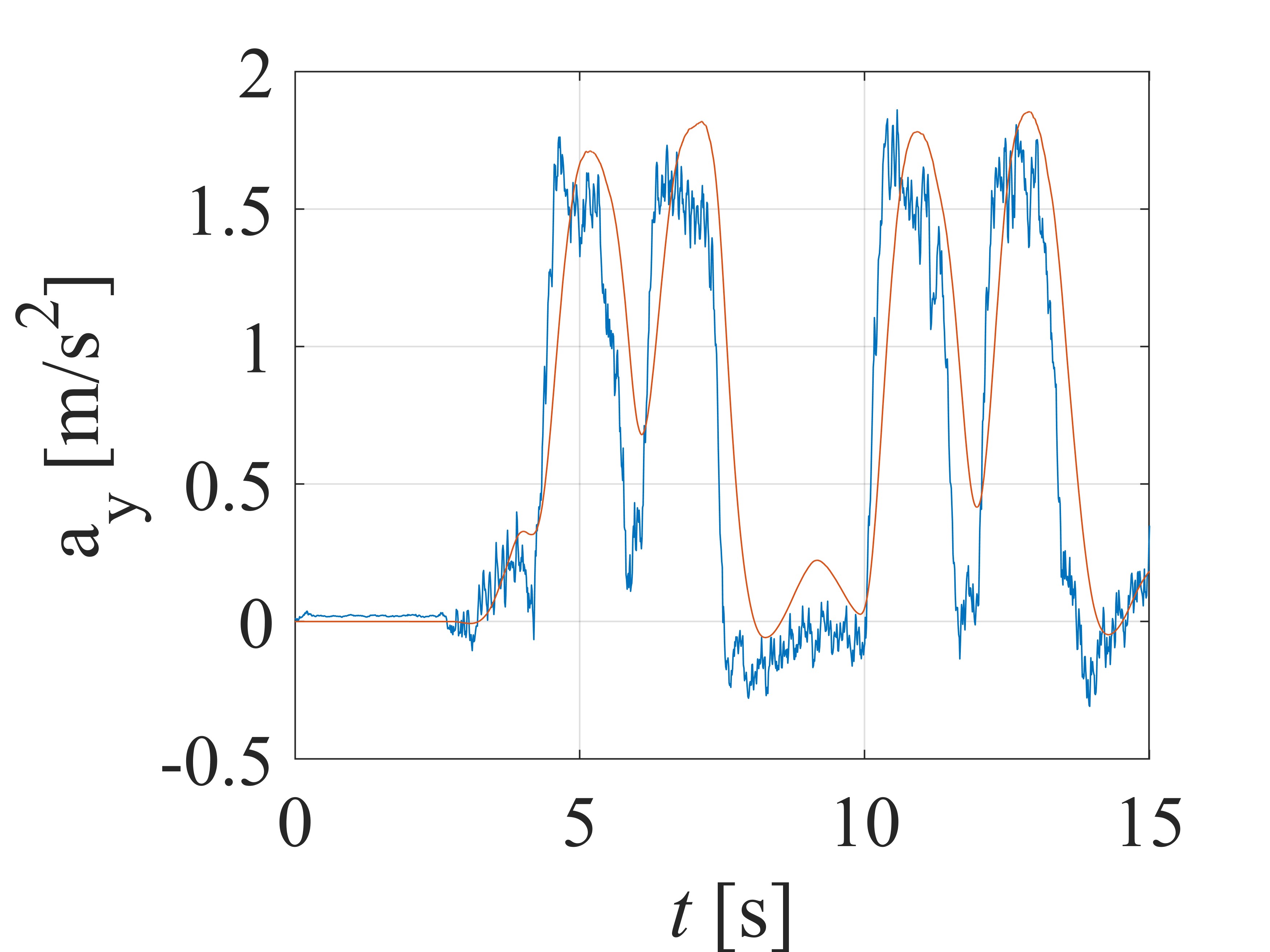}\label{fig:oval_validation_valid2}}
     \caption{Experimental model validation over the oval track: experimental data (blue line) and model predictions (red line) for (a) longitudinal speed and steering command (dashed gray line), (b) longitudinal acceleration, (c) yaw rate, and (d) lateral acceleration.}
\label{fig:oval_validation}
\end{figure}

Fig.~\ref{fig:oval_validation} depicts the response of the longitudinal and lateral dynamics when the inputs (i.e., the armature voltage $V_a$ and the steering angle $\delta$) are those generated by the line following algorithm along with the corresponding experimental responses. Fig.~\ref{fig:oval_validation} shows a good matching of the model predictions and the experimental results considering this challenging scenario. Specifically, the RMSE in the prediction of the longitudinal speed, longitudinal acceleration, yaw rate and  lateral acceleration, are 0.066 m/s, 0.13 m/s$^2$, 5.04 deg/s, 0.37 m/s$^2$, respectively, which are acceptable residual errors considering the order of magnitude of the respective signals. 

\section{Design of EKFs and the Federated EKF}\label{sec:EKF_desing}
\begin{figure}[t]
    \centering
    \includegraphics[width=0.4\textwidth]{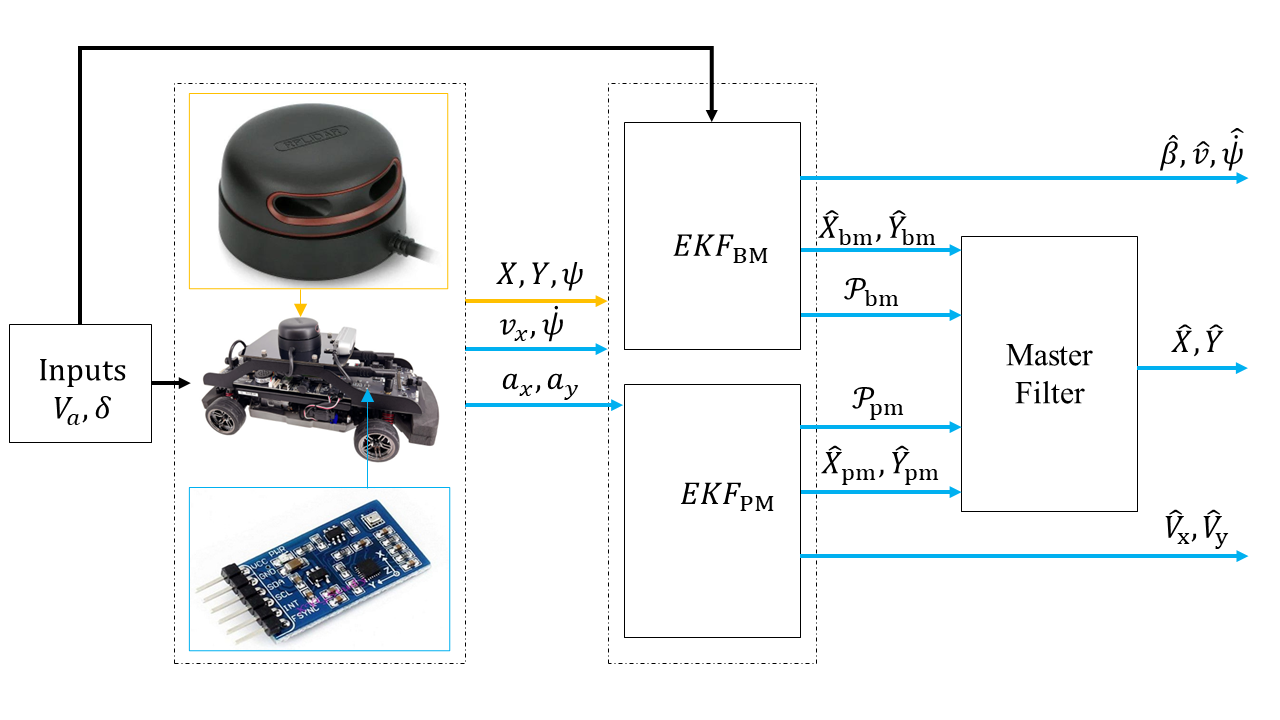}
    \caption{Scheme of the proposed federated extended Kalman filter for the robotic vehicle positioning.}
    \label{fig:FKF_architecture}
\end{figure}

Federated filtering is a data fusion technique in which  system states are first estimated through local filters. Subsequently, the local estimates are fused together through a master  filter to obtain a more reliable state estimate. Federated KFs uses KFs as local estimators and have  been  adopted in vehicle positioning and navigation (e.g., see [17],[18]). This work proposes a FEKF for increasing the accuracy of the estimate of the position of the robotic vehicle in the inertial frame. The proposed architecture 
is depicted in Fig.~\ref{fig:FKF_architecture} which also reports the inputs to the local filters along with the corresponding sampling frequencies (e.g., 100~Hz in light blue and 10 Hz in yellow).  Moreover,  given a vehicle state $\xi$, then $\widehat{\xi}$ in Fig.~\ref{fig:FKF_architecture} denotes the corresponding estimate. The proposed FEKF has a first stage composed by two EKFs --denoted as EKF$_{\text{PM}}$ and EKF$_{\text{BM}}$-- and operates in a \emph{no-reset} mode as the global fused estimation does not effect the local estimates, and the master filter does not retain fused data. According to [36] the \emph{no-reset} configuration increases tolerance to sensor faults (e.g., in the case of faults in the accelerometers). Fig.~\ref{fig:FKF_architecture} underlines that the master filter only operates on the positioning information while the remaining vehicle states are estimated  by the local filters. In the following subsections the design of the local filters and the master filter are presented in detail.


\subsection{Local EKFs}\label{sec:EKF_local}
Both local filters are based on the EKF algorithm, and can systematically tackle system nonlinearities and measurements with different sampling times. Specifically, the sampling time of the measures from the IMU and lidar are $10$ ms and $100$ ms, respectively. These measurements are fused by the local EKFs along with model predictions to obtain vehicle state estimates every $10$ ms to be used for vehicle control algorithms. 

The filter denoted as EKF$_{\text{BM}}$ uses the BM in Sections~\ref{sec:long_dyn} and \ref{sec:lat_dyn} for model predictions, i.e., 

\begin{subequations}\label{eq:EKF_bm}
\footnotesize
    \begin{align}
        \dot{X}_{\text{bm}} &= v \cos(\beta + \psi), \label{eq:EKF_bm_x}\\
        \dot{Y}_{\text{bm}} &= v \sin(\beta + \psi),\label{eq:EKF_bm_y}\\
        \dot v &= \frac{\mathcal{G}P_1^\star}{{{R_w}}}{V_a} - {P_2^\star}v  - \frac{\mathcal{G}P_3^\star}{{{R_w}}}{\mathop{\rm sgn}} \left( v  \right), \label{eq:EKF_bm_v}\\
        \dot{\psi} &=r,\label{eq:EKF_bm_psi}\\
             mv\dot \beta  &=  - ({C_f^\star} + {C_r^\star} + m\dot v)\beta  - \left( {\frac{{{C_f^\star}{L_f}}}{v} - \frac{{{C_r^\star}{L_r}}}{v} + mv} \right)r + {C_f^\star}\delta, \label{eq:eq:EKF_bm_beta}\\
             {I_G}\dot r &=  - ({C_f^\star}{L_f} - {C_r^\star}{L_r})\beta  - \left( {\frac{{{C_f^\star}L_f^2}}{v} + \frac{{{C_r^\star}L_r^2}}{v}} \right)r + {C_f^\star}{L_f}\delta, \label{eq:EKF_bm_r}
    \end{align}
\end{subequations}
where ${X}_{\text{bm}}$ and ${Y}_{\text{bm}}$ are the vehicle coordinates predicted via the BM. The system state is $\eta_{\text{bm}}=\left[{X}_{\text{bm}}\; {Y}_{\text{bm}}\; v \; \psi\; \beta\; r\right]^T$, and the measurable system output vector is $\zeta_{\text{bm}}=\left[{X}_{\text{bm}}\; {Y}_{\text{bm}}\; v \; \psi\; r\right]^T$ provided by the lidar and the encoder.

The filter denoted as EKF$_\text{PM}$ uses the kinematic point model (PM) presented in Section~\ref{sec:KP_model} for model predictions, i.e.,
\begin{subequations}\label{eq:EKF_pm}
    \begin{align}
        \dot{X}_{\text{pm}} &= \upsilon_x, \label{eq:EKF_pm_x}\\
        \dot{Y}_{\text{pm}} &= \upsilon_y,\label{eq:EKF_pm_y}\\
        \dot{V}_x &= {a_x}\cos (\psi ) - {a_y}\sin (\psi ), \label{eq:EKF_pm_vx}\\
        \dot{V}_y &= {a_x}\sin (\psi ) + {a_y}\cos (\psi ), \label{eq:EKF_pm_vy}
    \end{align}
\end{subequations}
where ${X}_{\text{pm}}$ and ${Y}_{\text{pm}}$ are the vehicle coordinates predicted via the kinematic point model and $V_x$ and $V_y$ are the components of the vehicle speed in the inertial frame. The system state is $\eta_{\text{bm}}=\left[{X}_{\text{pm}}\; {Y}_{\text{pm}} \; V_x  \; V_y \right]^T$, and the measurable system output is $\zeta_{\text{pm}}=\left[{X}_{\text{pm}}\; {Y}_{\text{pm}}\right]^T$.

For the design of the local filters EKF$_{\text{BM}}$ and EKF$_{\text{PM}}$, models \eqref{eq:EKF_bm} and \eqref{eq:EKF_pm}  have been discretised with a sampling time of 10 ms.

For the local filters, the entries of the covariance matrix of the process noise and observation noise  have been tuned heuristically via experiments and they have been assumed constants with the only exception of the entries of the observation noise matrices related to the lidar. The entries on the diagonal of the covariance matrices for the measurements provided by the lidar are time-varying and based on an additional output of the lidar software module provided by the robotic car manufacturer, which is known as \emph{lidar\_score}. The higher the value of the \emph{lidar\_score}, the higher the accuracy of the lidar measurement (i.e., the smaller its variance at a given sampling time). A nonlinear mapping has been used to transform the  \emph{lidar\_score} signal into the variance of each lidar output, i.e.,
\begin{equation}\label{eq:R_lidar}
{\mathcal{R}_\xi } = {\kappa _{\xi,1}}\tanh \left( {\frac{{{\kappa _{\xi,2}}}}{{{l_s}}} - {\kappa _{\xi,3}}} \right) + {\kappa _{\xi,4}},
\end{equation}
where  $l_s$ is the \emph{lidar\_score},  $\mathcal{R}_\xi$ is the variance of the lidar measurement $\xi$,  and $\kappa_{\xi,i}$,  with $\xi\in\left\{x,y,\psi\right\}$ and $i=1,2,3,4$, are positive constants experimentally tuned. 

\subsection{Master Filter}\label{sec:EKF_master}
By denoting  $\widehat{p}_{\text{bm}}=\left[\widehat{X}_{\text{bm}}\; \widehat{Y}_{\text{bm}}\right]^T$ and $\widehat{p}_{\text{pm}}=\left[\widehat{X}_{\text{pm}}\; \widehat{Y}_{\text{pm}}\right]^T$ the estimates of the robotic car position in the inertial frame predicted via the EKF$_{\text{BM}}$ and EKF$_{\text{PM}}$, respectively, then the  estimate at the discrete time instant $k$ of the vehicle position $\widehat{p}(k)=\left[\widehat{X}(k)\;\; \widehat{Y}(k)\right]^T$ is computed by the master filter as follows:

\begin{subequations}\label{eq:master}
\small
    \begin{align}
         \mathcal{P}_f^{ - 1}(k) &= \mathcal{P}_\text{bm}^{ - 1}(k) + \mathcal{P}_\text{pm}^{ - 1}(k),\label{eq:master_P}\\
         \widehat p(k) &= {\mathcal{P}_f}(k)\left[ {\mathcal{P}_\text{bm}^{ - 1}(k){{\widehat p}_\text{bm}}(k) + \mathcal{P}_\text{pm}^{ - 1}(k){{\widehat p}_\text{pm}}(k)} \right] \label{eq:master_p_xy},
    \end{align}
\end{subequations}
where  the inverse of the global covariance matrix $\mathcal{P}_f$ is known as information matrix, while $\mathcal{P}_\text{bm}(k)$ , $\mathcal{P}_\text{pm}(k)\in \mathbb{R}^2$ 
are the sub-blocks of the covariance matrices of the EKF$_{\text{BM}}$ and EKF$_{\text{PM}}$, respectively, corresponding to the vehicle position. 

\section{Validation of the Filtering System}\label{sec:EKF_validation}
This section aims to show the effectiveness of the proposed filtering system in both simulations and experiments. Specifically, simulations are used to demonstrate the ability of the filtering system to predict the vehicle sideslip angle which usually is not directly measurable through  on-board sensors of  robotic ground cars but it is used for vehicle control. Experiments aim at showing the effectiveness of the approach to generate smooth predictions of the vehicle state which can be used for  control tasks. Moreover, it is experimentally shown the ability of the FEKF to  remove spikes in the lidar measurements which could jeopardise vehicle control systems.

\subsection{Simulation Results} \label{sec:EKF_simulation}
\begin{figure}[t]
     \centering
     \subfloat[]
      {\includegraphics[width=0.25\textwidth]{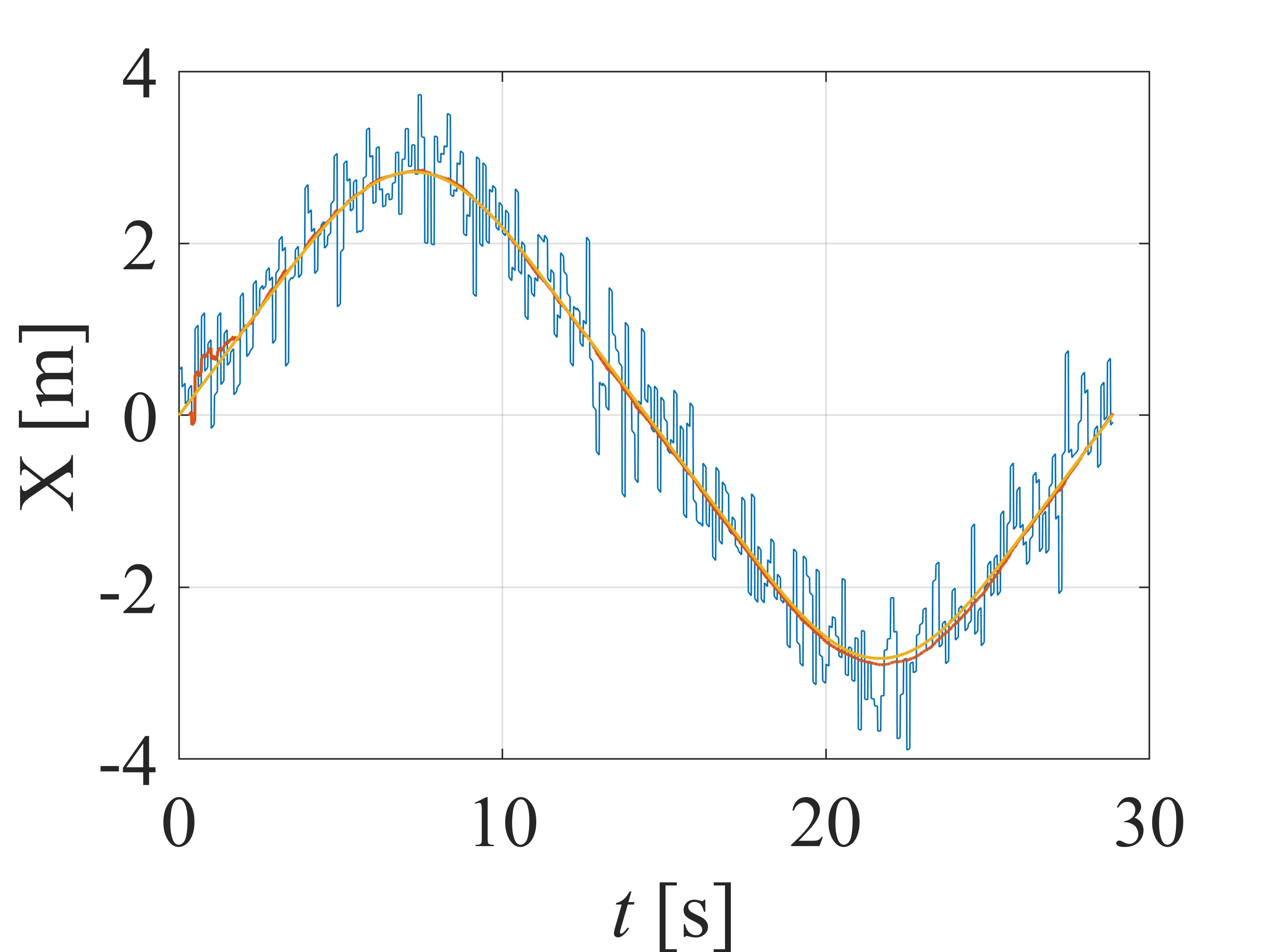}\label{fig:ekf_sim_x}} 
     \subfloat[]
     {\includegraphics[width=0.25\textwidth]{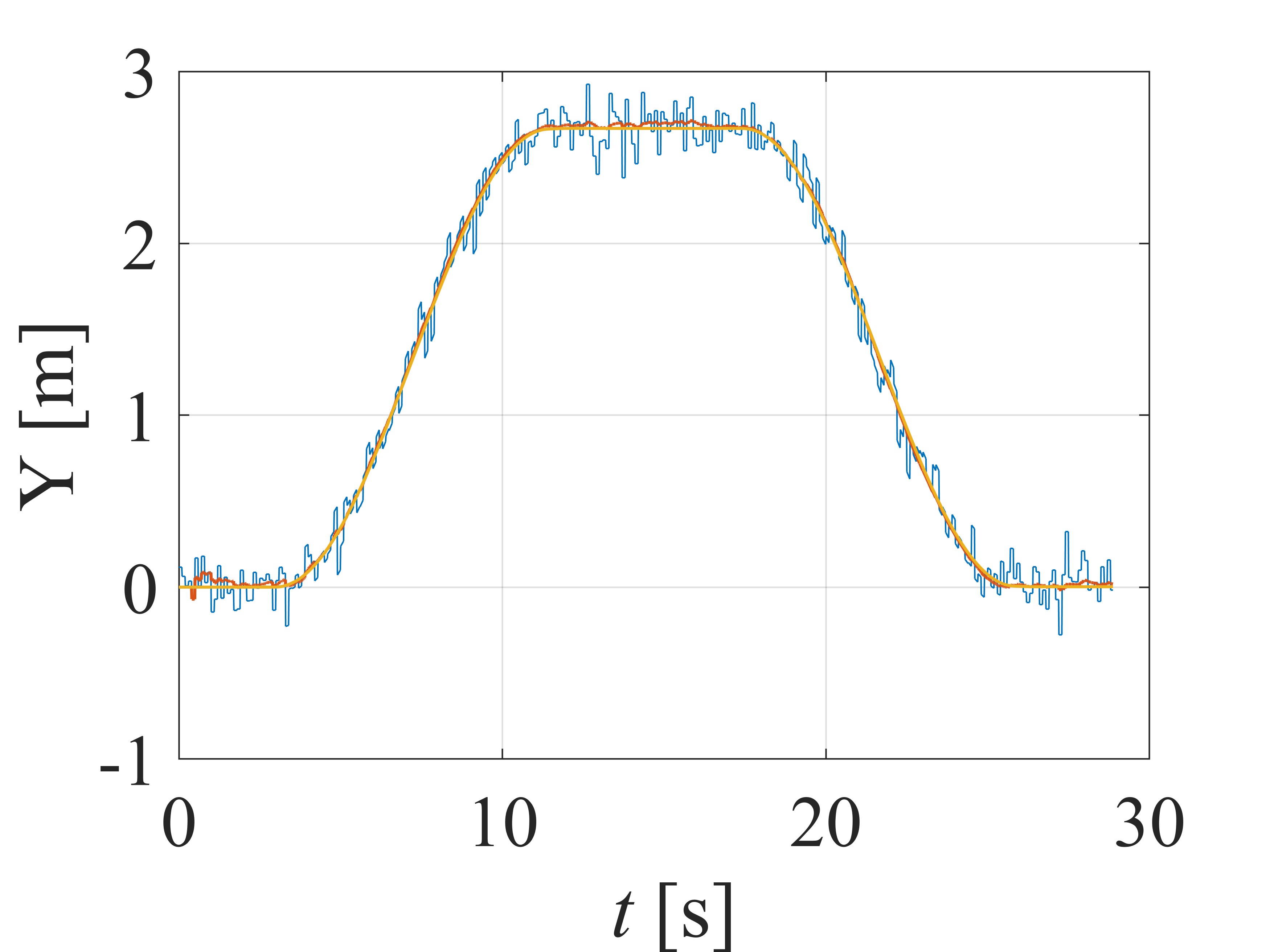}\label{fig:ekf_sim_y}}\\
          \subfloat[]
     {\includegraphics[width=0.25\textwidth]{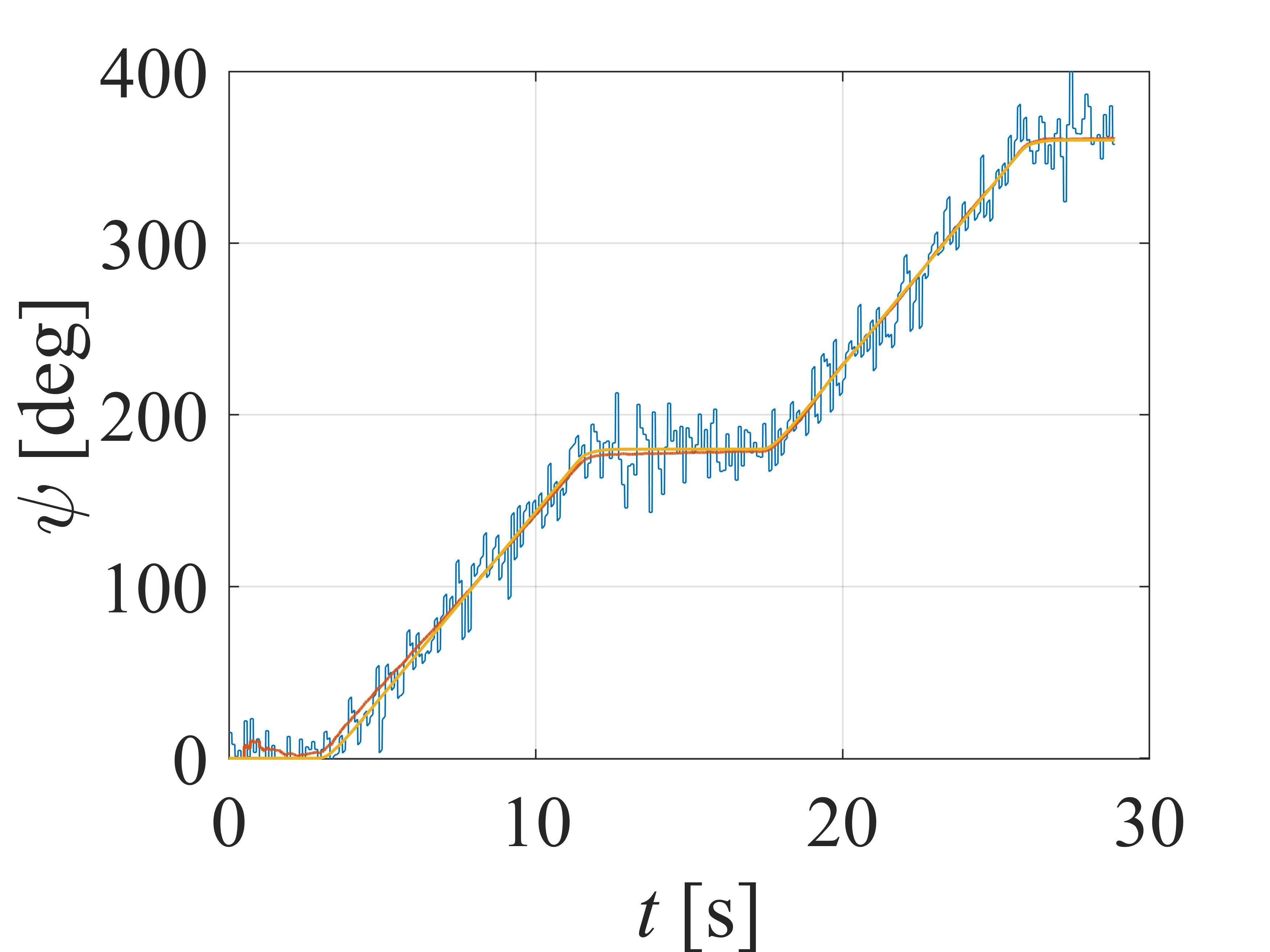}\label{fig:ekf_sim_yaw}}
         \subfloat[]{\includegraphics[width=0.25\textwidth]{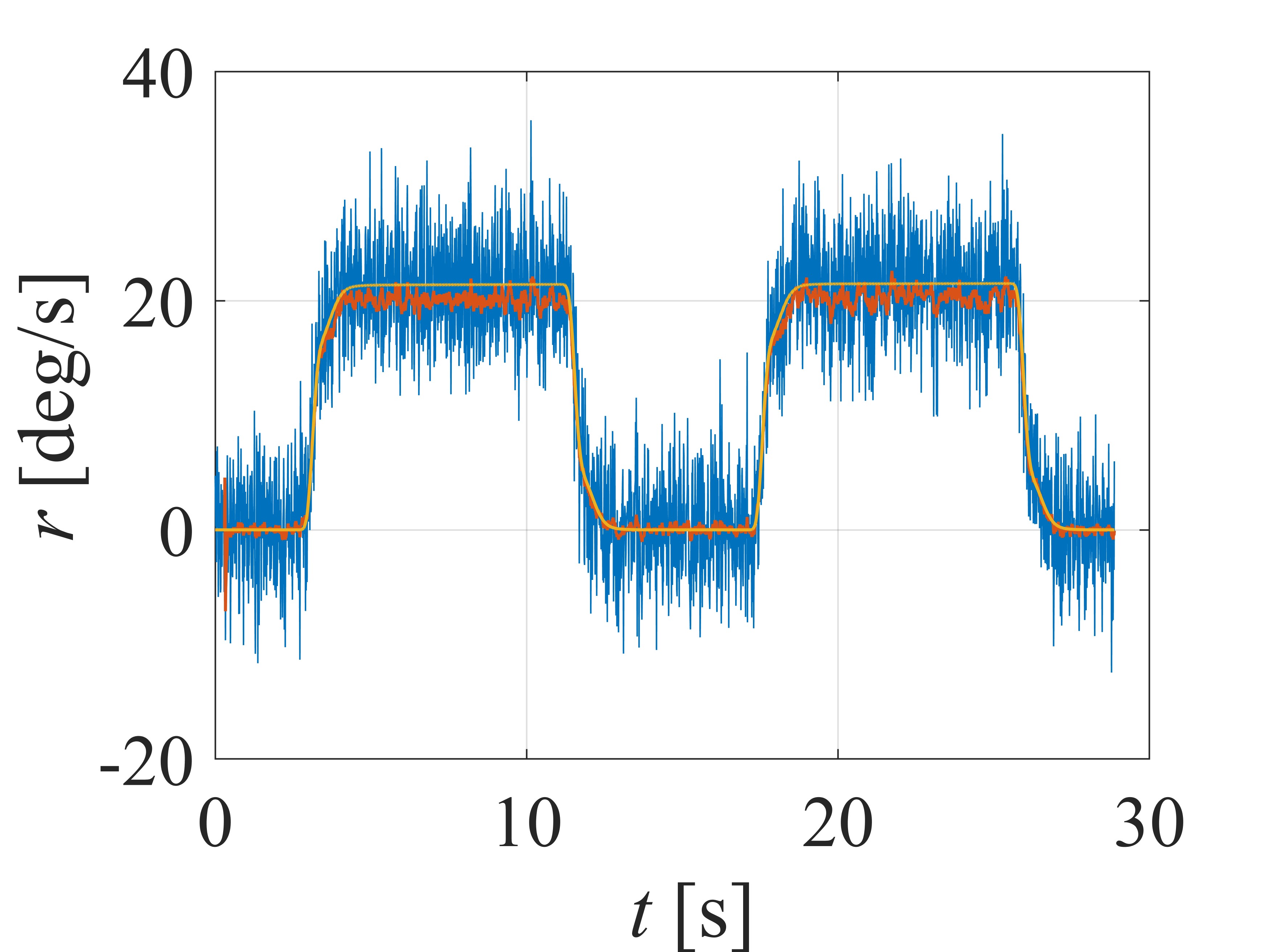}\label{fig:ekf_sim_yaw_rate}}\\
                  \subfloat[]{\includegraphics[width=0.25\textwidth]{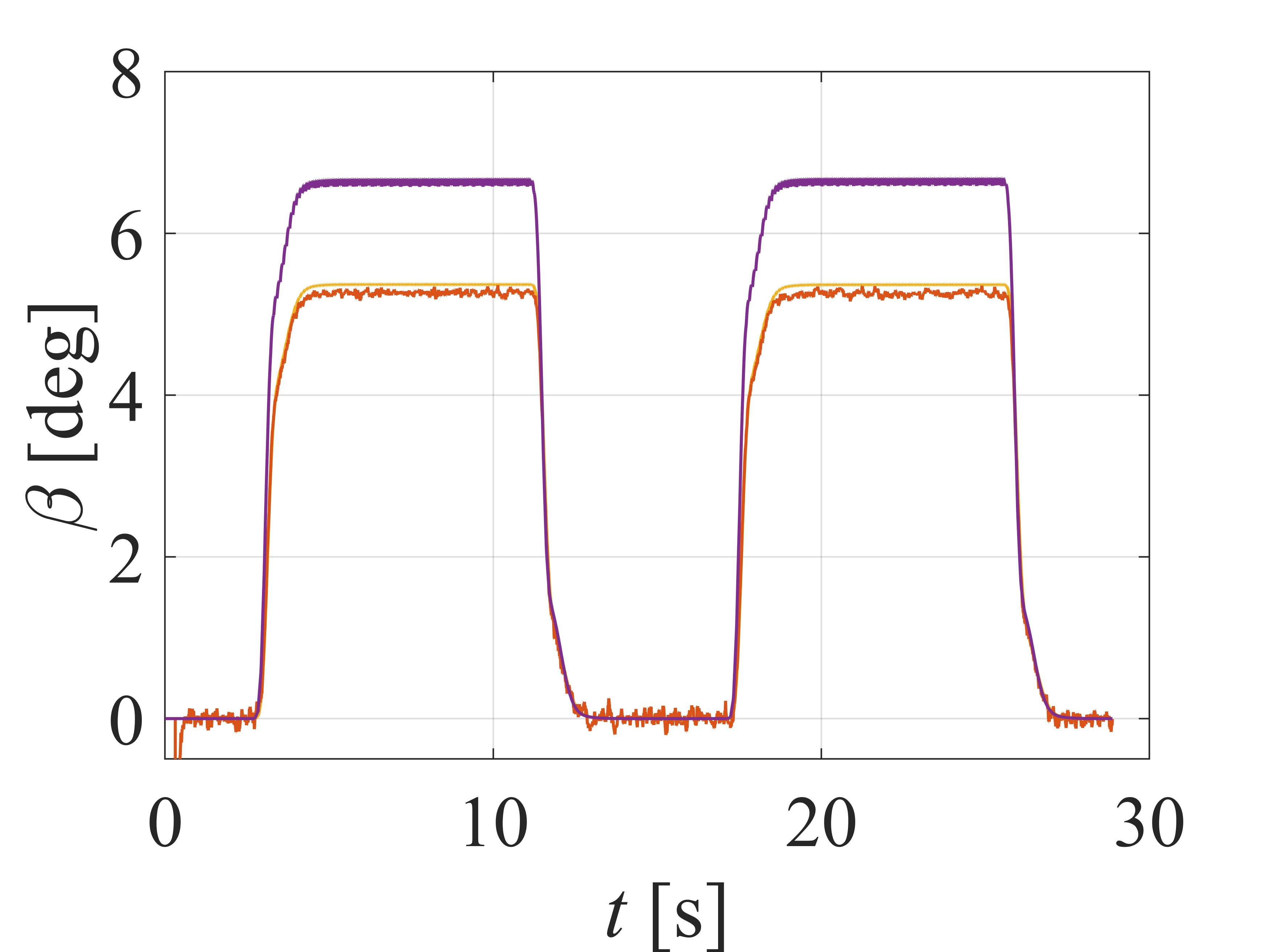}\label{fig:ekf_sim_beta}}
     \caption{Simulation results of the filtering system:  measurement  (blue line), estimate (red line) and actual state (yellow line),; (a) $X$-position; (b) $Y$-position;  (c) heading angle $\psi$; (d) yaw rate $r$ and (e) sideslip angle $\beta$ and its kinematic approximation (purple line). }
\label{fig:EKF_simulations}
\end{figure}

To test the FEKF approach in Section \ref{sec:EKF_desing} in simulation,  Gaussian noises with variance tuned based on the experimental data of the available sensors  have been added to the simulation model. An oval track similar to the one used for the line following algorithm provided by the robotic vehicle manufacturer has been simulated. The oval path has been imposed to the vehicle through the use of 
\emph{(i)}~a~PI  controller to keep the speed constant to $0.5$ m/s  and \emph{(ii)} the  
LQ strategy presented in [37] to keep the vehicle on the path during the simulation.
The estimate of some vehicle states are shown in Fig.~\ref{fig:EKF_simulations} to demonstrate the ability of the proposed estimation system to predict vehicle dynamics by filtering out noises.

\noindent\textbf{Remarks}
\begin{itemize}
\item Figs. \ref{fig:ekf_sim_x} and \ref{fig:ekf_sim_y} show that the estimates of the $X$ and $Y$ coordinates provided by the master filter well capture the actual vehicle position.

\item Fig. \ref{fig:ekf_sim_yaw} and Fig. \ref{fig:ekf_sim_yaw_rate} demonstrate that EKF$_\text{BM}$ provides good estimates of the actual  heading angle and yaw rate by filtering out noise given by the lidar and IMU measurements. 

\item Figs. \ref{fig:ekf_sim_beta} shows the effectiveness of the EKF$_\text{BM}$ to estimate the dynamics of the sideslip angle. Furthermore,  Fig.~ \ref{fig:ekf_sim_beta} demonstrates that the approximation of the side slip-angle via the kinematic relation: 
\begin{equation}\label{eq:beta_kinematic}
\widehat{\beta}_{KM}= \arctan \left( {\frac{{{C_f}}}{C_f + C_r} } \right) \tan(\delta)
\end{equation}
is not effective for scaled robotic cars and it provides estimation errors larger than those given by the EKF$_\text{BM}$.
 
\end{itemize}

\subsection{Experimental Results} \label{sec:EKF_experiment}
The experimental test of the FEKF approach in Section~\ref{sec:EKF_desing} has been carried out deploying the same controllers used in simulation on the oval track (see also Fig.~\ref{fig:oval_track}). In this case the manoeuvre includes speed changes during cornering, i.e., the speed is reduced approximately by 15\% to facilitate the manoeuvre. 

\begin{figure}[t]
     \centering
               \subfloat[]
     {\includegraphics[width=0.25\textwidth]{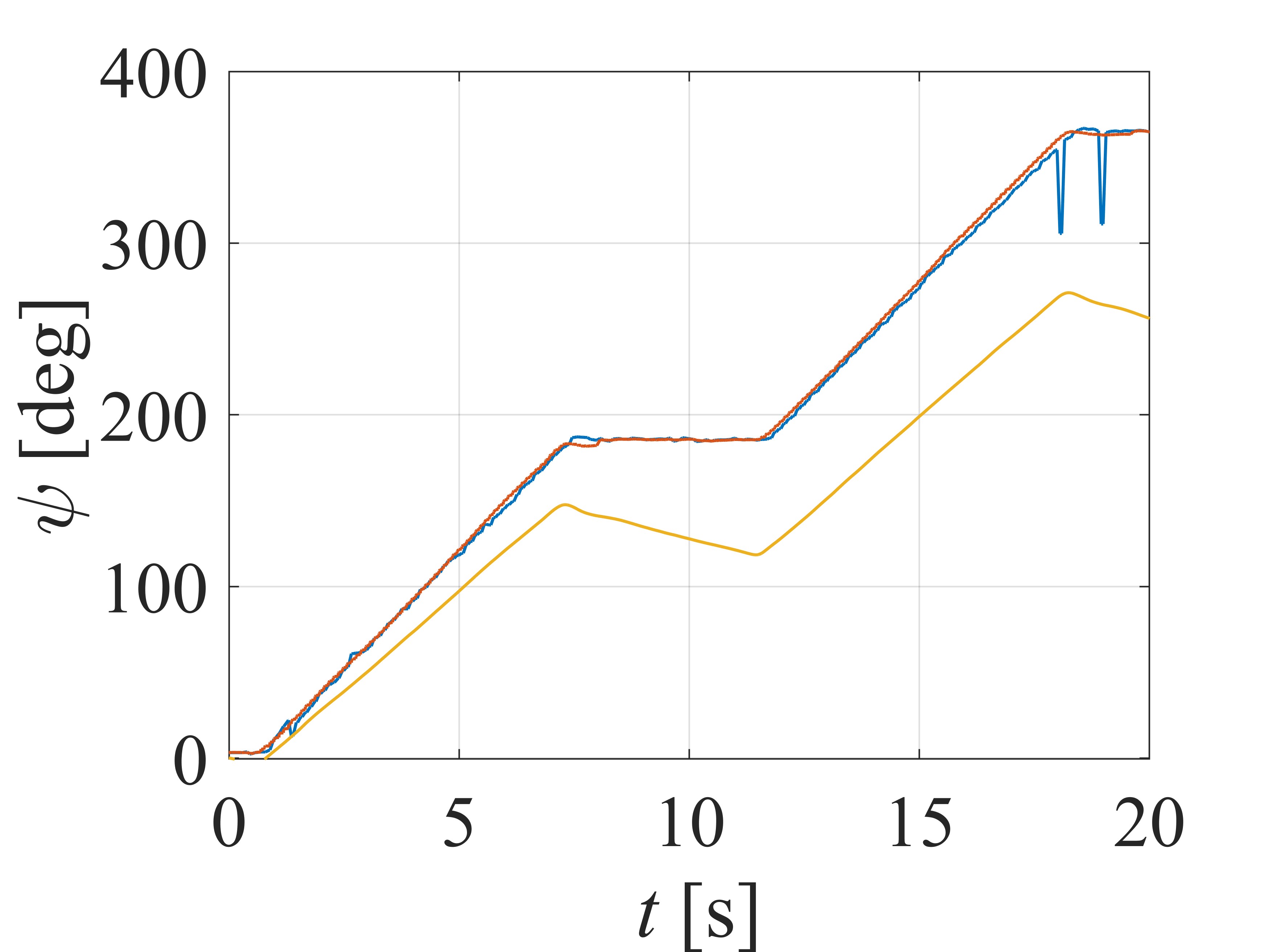}\label{fig:ekf_exp_yaw}}
         \subfloat[]{\includegraphics[width=0.25\textwidth]{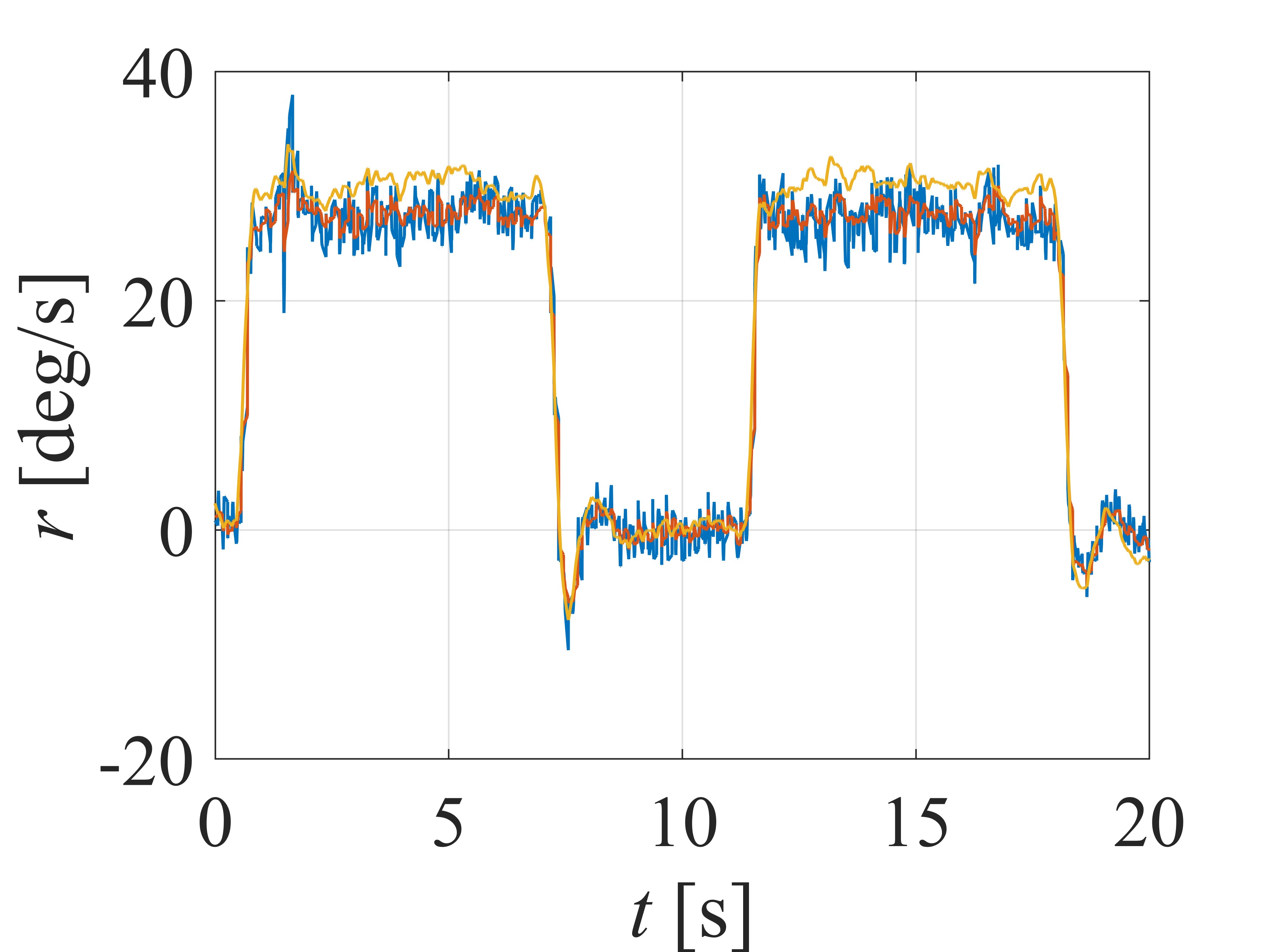}\label{fig:ekf_exp_yaw_rate}}\\
                  \subfloat[]{\includegraphics[width=0.25\textwidth]{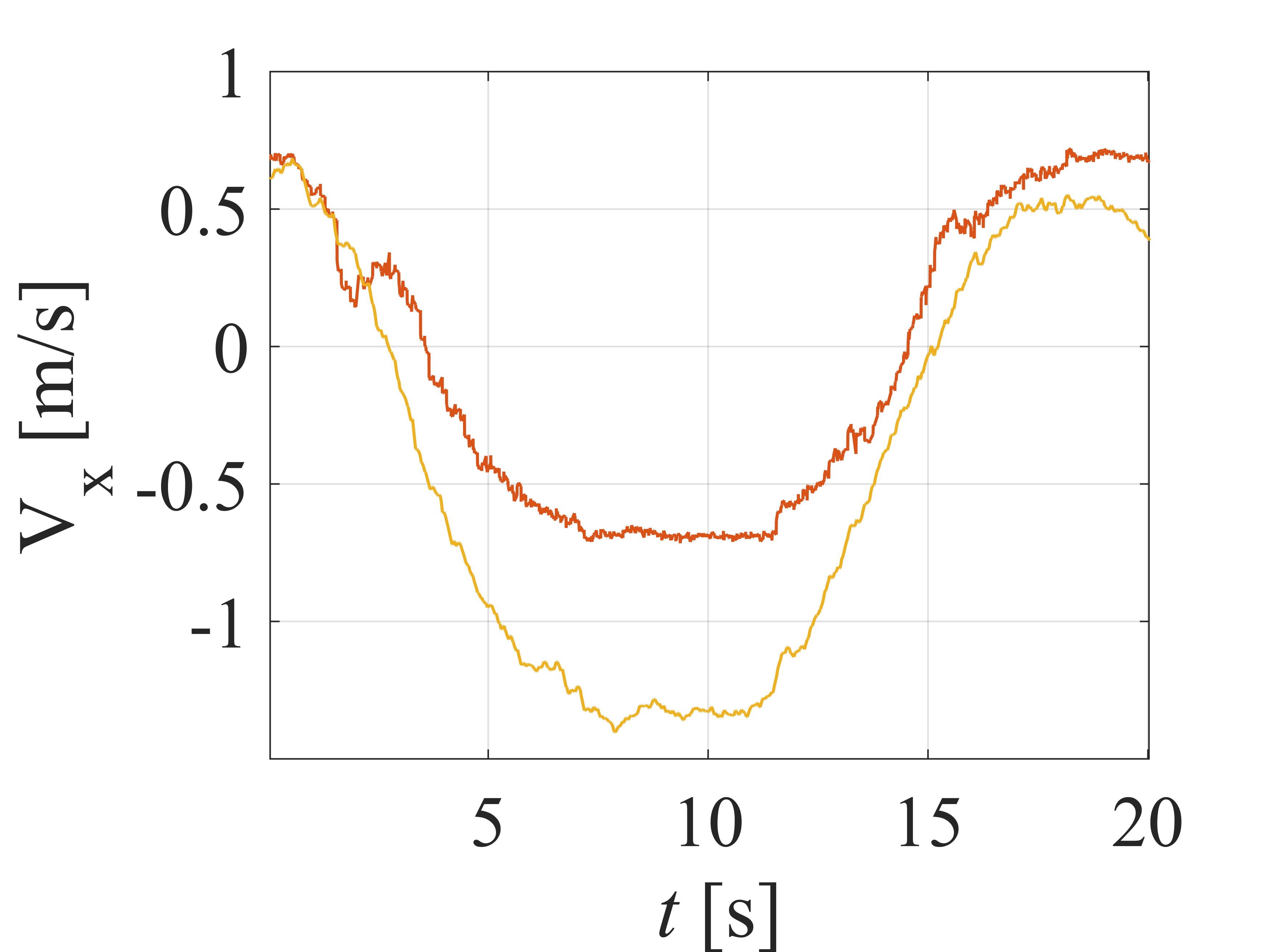}\label{fig:ekf_exp_vx}}
               \subfloat[]{\includegraphics[width=0.25\textwidth]{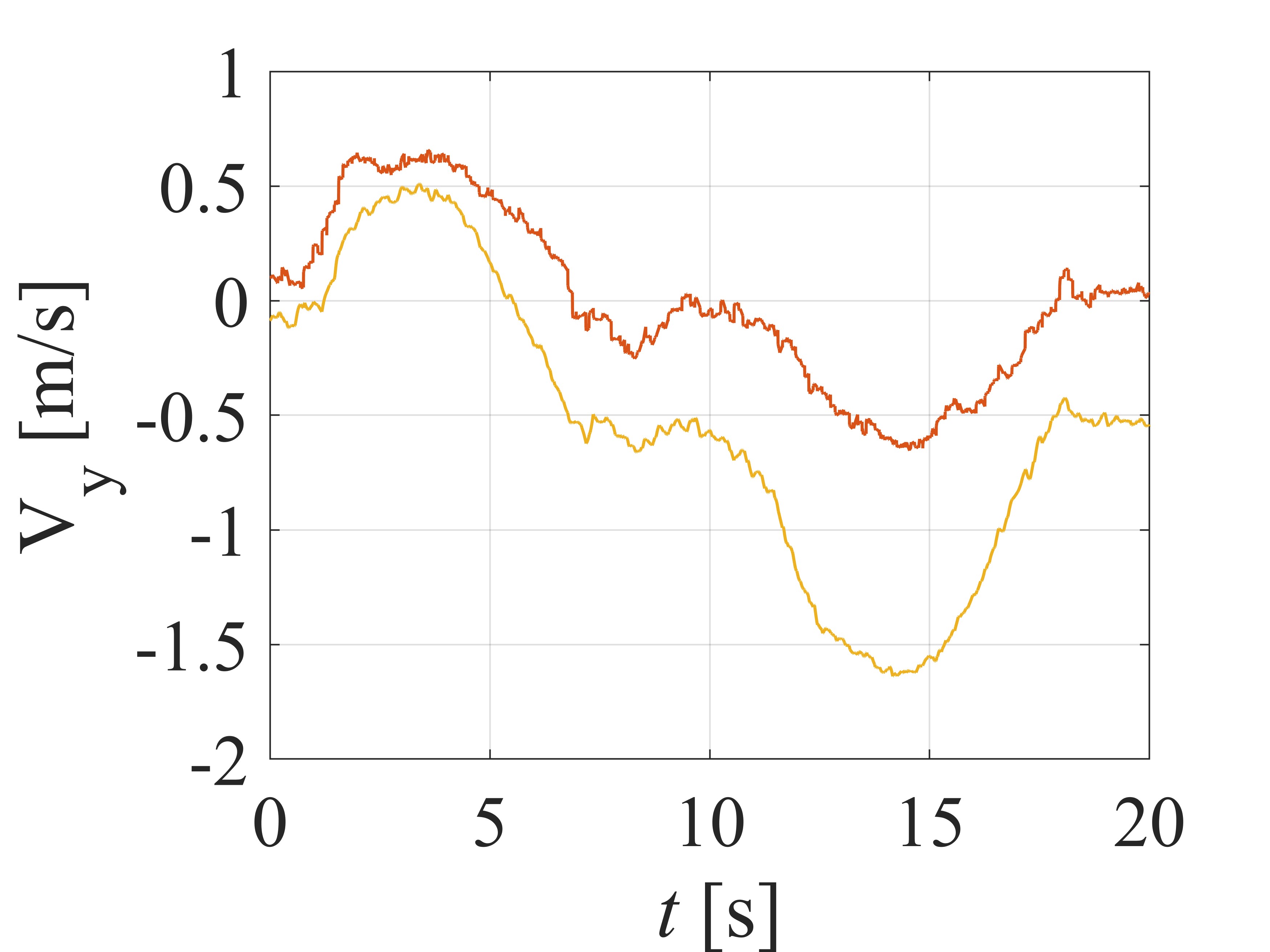}\label{fig:ekf_exp_vy}}                 \\
  \subfloat[]
      {\includegraphics[width=0.25\textwidth]{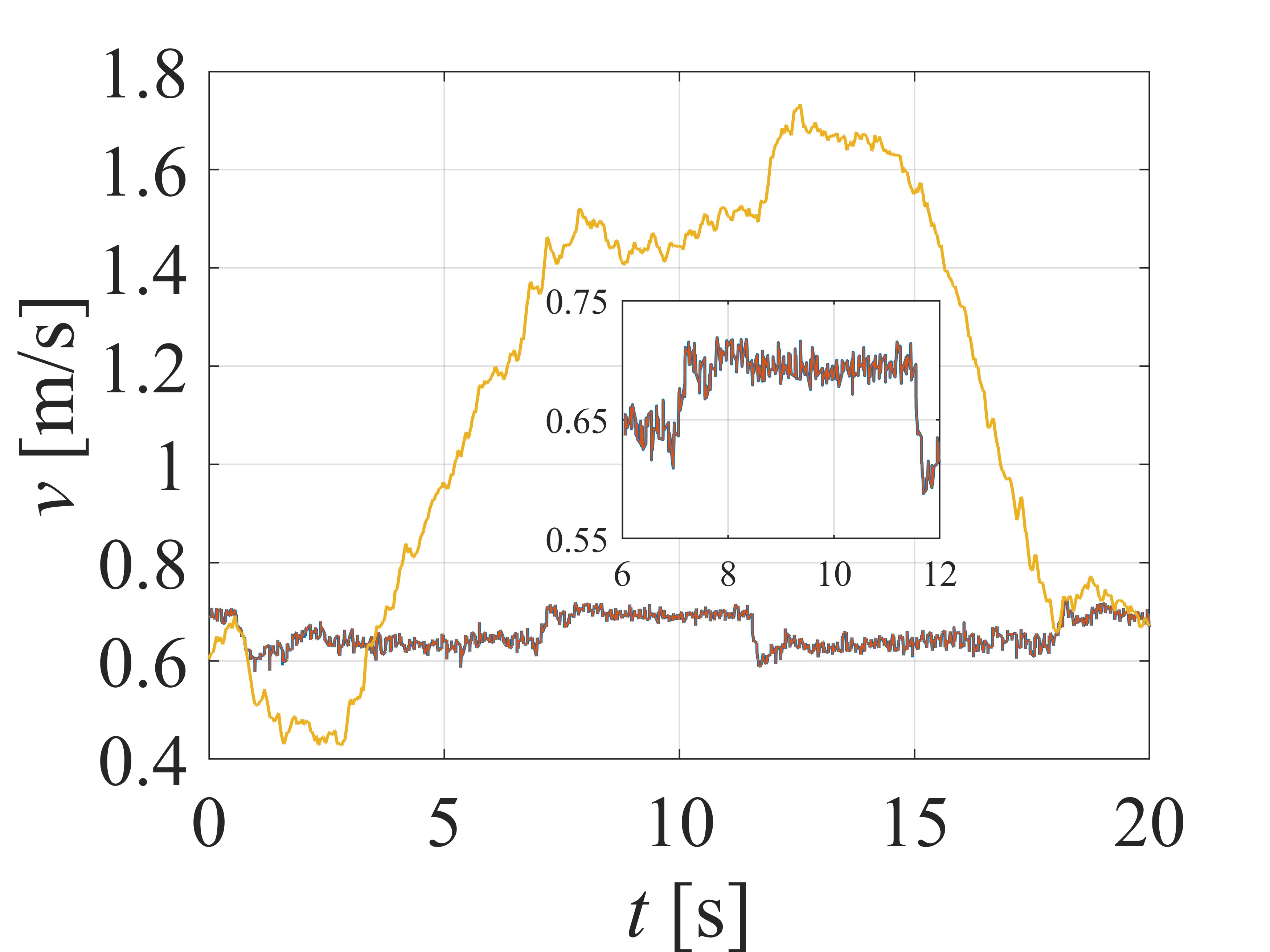}\label{fig:ekf_exp_v}} 
     \subfloat[]
     {\includegraphics[width=0.25\textwidth]{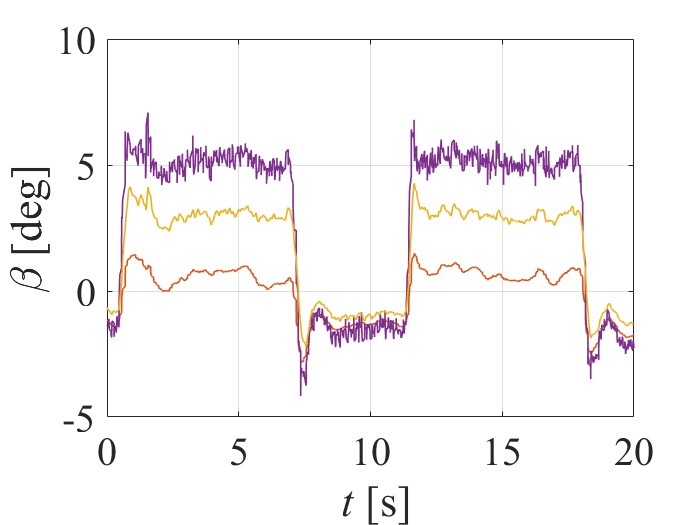}\label{fig:ekf_exp_beta}}\\
     \caption{Experimental validation of the filtering system: measurement (blue line), model predictions (yellow line), and estimates (red line) of (a) heading angle $\psi$, (b) yaw rate $r$, (c) longitudinal speed in the inertial frame $V_x$, (d) lateral speed in the inertial frame $V_y$, (e) vehicle speed $v$ and (f) sideslip angle $\beta$ and its kinematic approximation (purple line).}
\label{fig:EKF_exp}
\end{figure}

The ability of the proposed filtering approach to predict vehicle state through multiple sensors with different sampling time is shown in Fig.~\ref{fig:EKF_exp}, which reports the EKF estimates for some vehicle states, the sensor outputs and the vehicle states predicted via the integration of systems \eqref{eq:EKF_bm} and \eqref{eq:EKF_pm}.

\noindent\textbf{Remarks}
\begin{itemize}
\item Vehicle states predicted via the EKF$_\text{BM}$ are shown in Figs.~\ref{fig:ekf_exp_yaw},\ref{fig:ekf_exp_yaw_rate},\ref{fig:ekf_exp_v} and \ref{fig:ekf_exp_beta} and confirm the ability of the EKF$_\text{BM}$ of \emph{(i)} filtering out IMU and lidar measurement noise from the corresponding predictions; and \emph{(ii)} avoiding the drift of the state predictions obtained through the integration of model \eqref{eq:EKF_bm_v}-\eqref{eq:EKF_bm_r} without measurements corrections. 
\item Figs. \ref{fig:ekf_exp_vx} and \ref{fig:ekf_exp_vy} show that EKF$_\text{PM}$ estimates of the longitudinal speed and lateral speed in the inertial frame are not affected by the drift of the predictions obtained by integrating model \eqref{eq:EKF_pm}. 
\item  Fig. \ref{fig:ekf_exp_v} depicts  \emph{(i)} the magnitude of the vector speed estimated by the filter EKF$_\text{PM}$ (red line); \emph{(ii)} the magnitude of the vector speed predicted by integrating model \eqref{eq:EKF_pm} (yellow line); and \emph{(iii)} the output of the encoder (blue line). Fig. \ref{fig:ekf_exp_v} confirms that the vector speed provided by the EKF$_\text{PM}$ is more accurate than that obtained by integrating \eqref{eq:EKF_pm} as its magnitude better matched the encoder data.
\item Fig. \ref{fig:ekf_exp_beta} shows that the predicted sideslip angle obtained via the EKF$_\text{BM}$ differs from the one obtained through the kinematic approximation \eqref{eq:beta_kinematic}. However, based on the simulation analysis in Section~\ref{sec:EKF_simulation}, it is expected that the estimates provided by the EKF are more accurate compared to those  based on the kinematic equation.
\end{itemize}

\begin{figure}[t]
     \centering
  \subfloat[]
      {\includegraphics[width=0.25\textwidth]{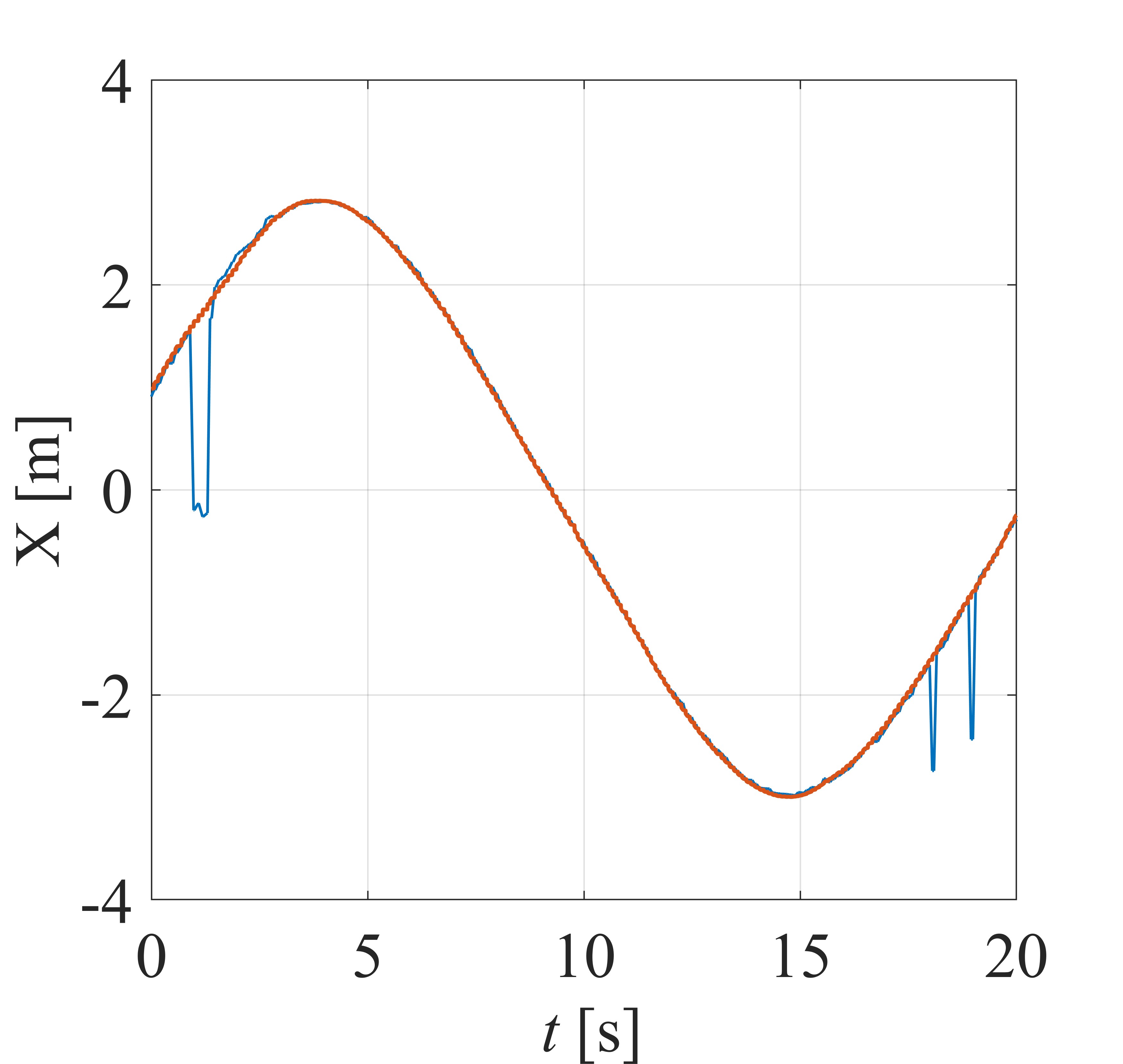}\label{fig:ekf_exp_x}} 
     \subfloat[]
     {\includegraphics[width=0.25\textwidth]{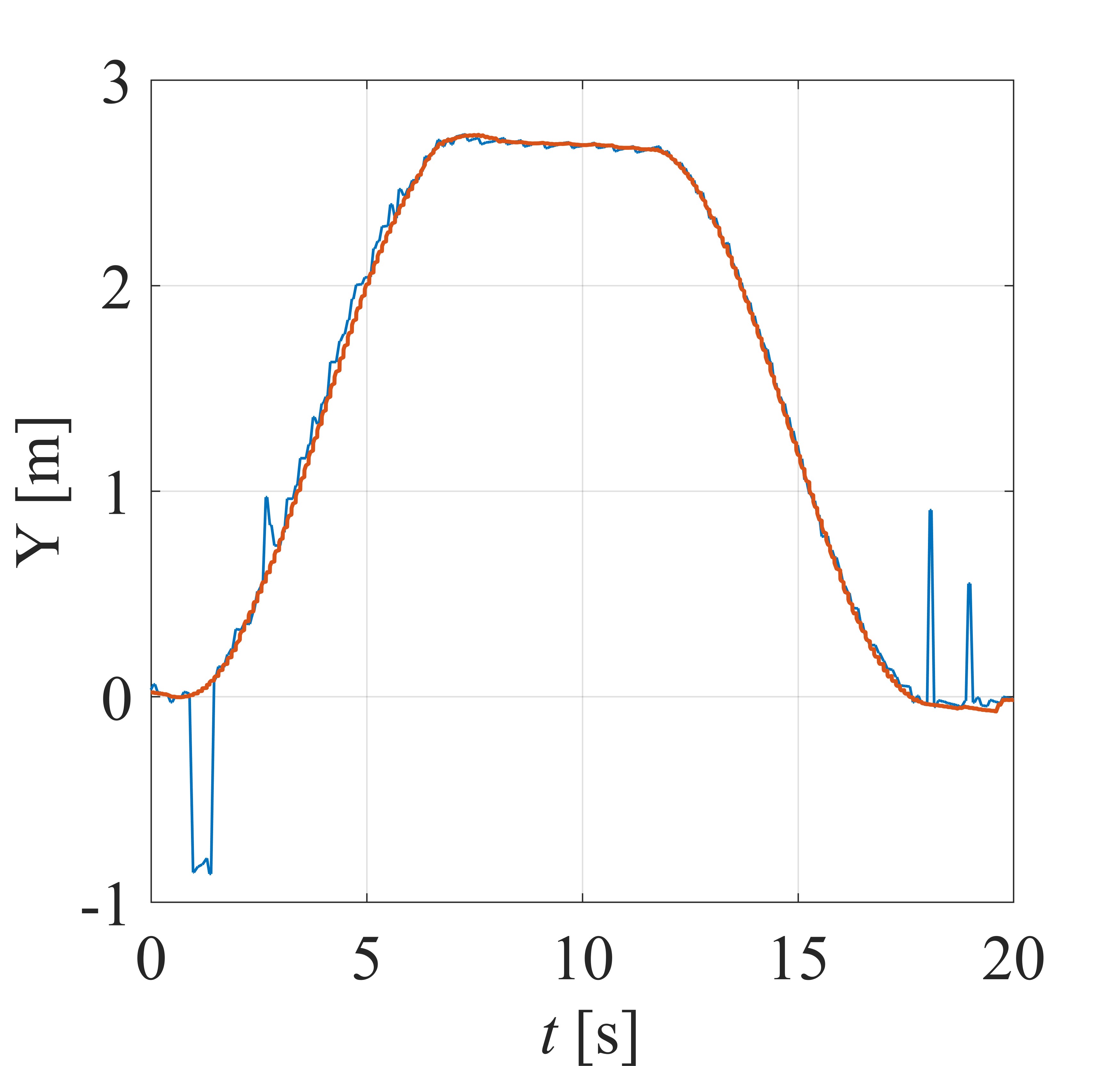}\label{fig:ekf_exp_y}}\\
     \subfloat[]
          {\includegraphics[width=0.3\textwidth]{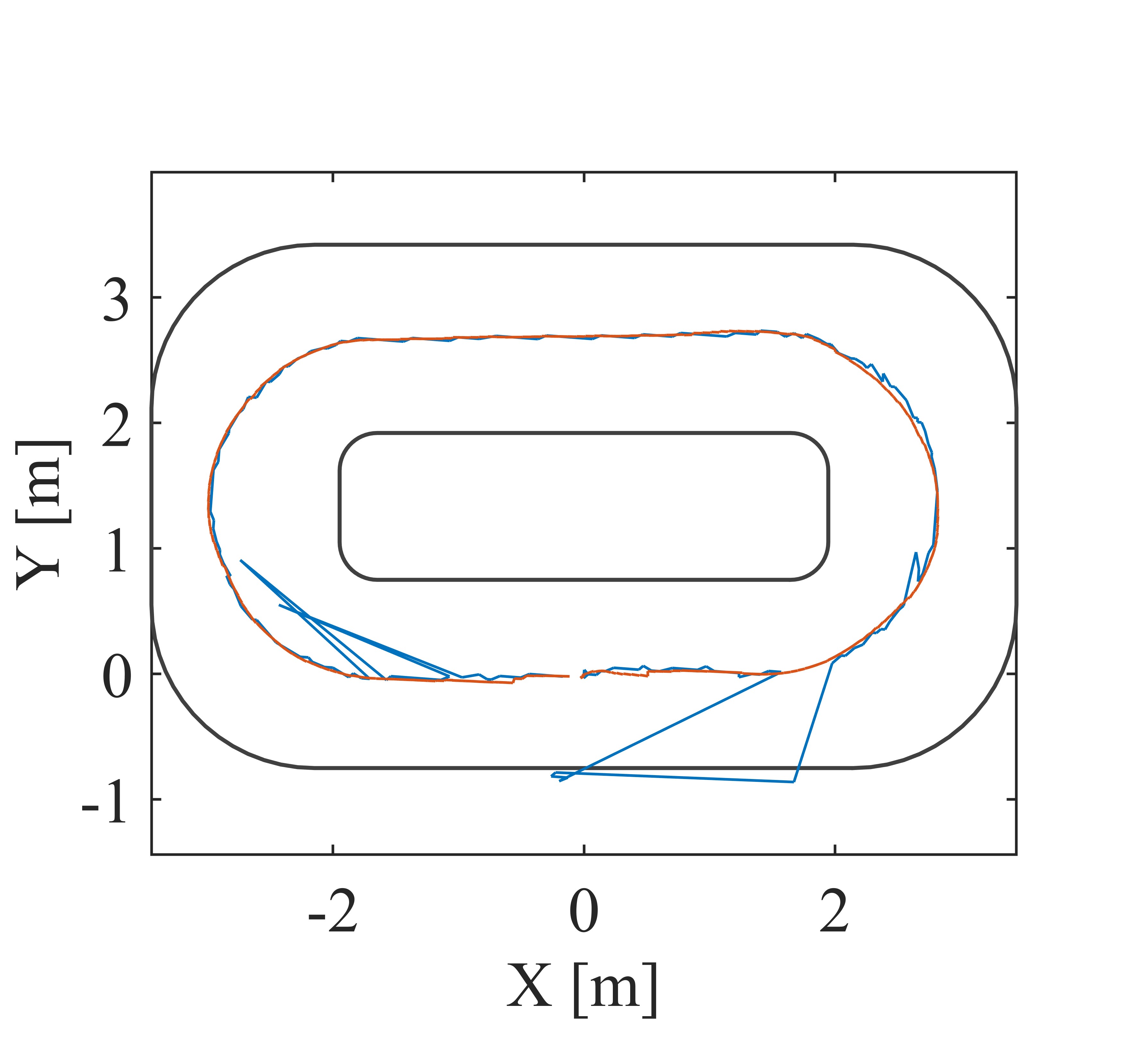}\label{fig:ekf_exp_xy}}\\
     \caption{Experimental validation of the filtering system: lidar measurements (blue line) and master filter predictions (red line) of (a) $X$ position, (b) $Y$ position, and (c) path.}
    \label{fig:EKF_exp_positions}
\end{figure}

Fig.~\ref{fig:EKF_exp_positions} shows the estimate of the vehicle position and highlights the ability of the proposed positioning system to remove the spikes in the lidar measurements.  The lidar spikes  cause false jumps in the vehicle path as  shown in Fig.~\ref{fig:ekf_exp_xy} which can negatively affect vehicle control systems. Fig.~\ref{fig:ekf_exp_xy} also confirms that the proposed filtering method is able to provide a smooth trajectory which can be exploited for control purposes.

\begin{figure}[t]
    \centering
    \includegraphics[width=0.4\textwidth]{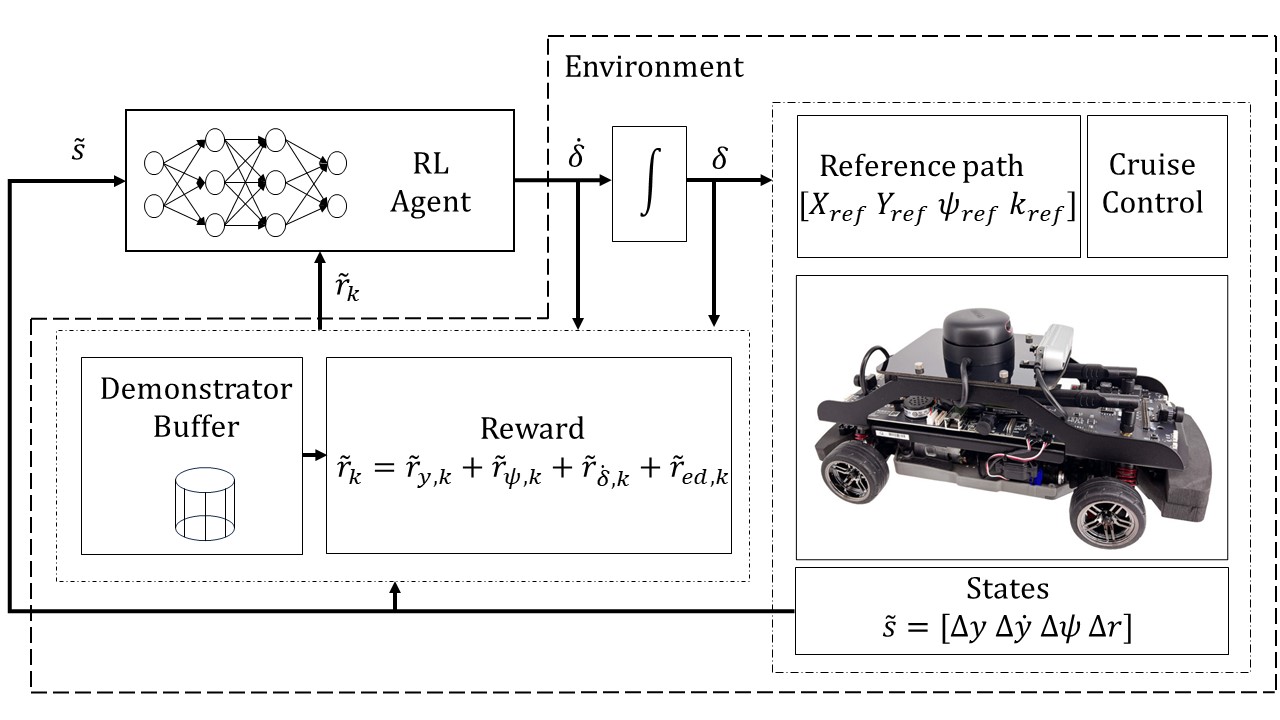}
    \caption{Working principle of the learning process for the DRL path tracking solution.}
    \label{fig:DRL_working_principle}
\end{figure}

\section{DRL-based Path Tracking Design}\label{sec:DRL_desing}
In the DRL control framework the agent (i.e., the controller)  decides the action $\widetilde{a}$ (i.e., the control input) based on a set of observations $\widetilde{s}$ (i.e., input data to the agent, also known as states) from the environment, consisting of the nominal plant and its relevant disturbances. The policy $\pi$ (i.e., the control strategy) mapping the observations onto the action is learnt through the interaction of the agent with the environment via the maximisation of a cumulative reward 
over a sequence of simulations, denoted also as episodes. The cumulative reward is expressed as the sum of rewards $\widetilde{r}_k$ computed via a reward function at each simulation step $k$. Fig.~\ref{fig:DRL_working_principle} shows the working principle of the learning process for the proposed DRL-based PT strategy. 

This work proposes a DRL-based PT control design where the formulation of the reward function includes the dynamics of an expert demonstrator. Including in the DRL learning process some forms of expert demonstrations has the following advantages [34] - [40]: 
\begin{itemize}
\item reduces the need for extensive explorations of the environment, resulting in faster learning of the optimal policy;
\item helps avoid unacceptable suboptimal policies during the random exploration phase;
\item informs the agent's learning process of system constraints (e.g., safety constraints) by considering them in the expert's policy;
\item enhances sample efficiency by reducing the need for excessive explorations. Training DRL agents from scratch usually demands extensive environment interactions to uncover a viable policy;
\item provides a reference for good actions, making it easier for the agent to explore the environment effectively. This is particularly relevant for DRL problems involving a high-dimensional action space.
\end{itemize}
Furthermore, the learning process can finely adjust the expert's policy through an ad-hoc design of reward functions, thus the trained agent can provide better closed-loop performance (e.g., tracking performance) compared to the demonstrator. 

The agent-environment interaction is modelled as a Markov decision process (MDP) described by the tuple $\left(\mathfrak{S}, \mathcal{A},  \gamma, \mathcal{R}, \mathcal{T}\right)$, where $\mathfrak{S}$ is the set of the system state with dimension $n_\mathfrak{S}$;  $\mathcal{A}$ is the set of possible actions with dimension $n_\mathcal{A}$; $\gamma \in (0,\;\; 1)$ is the discount factor; $\mathcal{R}$  is the set or rewards and $\mathcal{T}(\widetilde{s}_k,\widetilde{a}_k,\widetilde{s}_{k+1})$ is the transition function at the time step $k$, respectively. While training, at each time step $k$, the agent receives the states from the environment $(\widetilde{s}_k \in \mathfrak{S})$ as inputs, and outputs a control action $(\widetilde{a}_k \in \mathcal{A})$. Along with the states, the agent also receives the reward $(\widetilde{r} \in \mathcal{R})$ which is a measure of the goodness of the transition $\mathcal{T}$ with respect to the task the agent has to perform. The rewards are used to adjust the agent's policy and the optimal agent's policy $\pi^\star$ is found by solving the following stochastic optimisation problem [41],[42]:

\begin{subequations}\label{eq:optimisation}
    \begin{align}
 {\pi ^\star} &=  \mathop {\arg \min}\limits_\pi  {E_{(\widetilde s,\widetilde a) \sim {\rho ^\pi }}}\left[ {{E_{\scriptstyle\tau  \sim {\tau ^\pi }\hfill\atop
{\scriptstyle{{\widetilde s}_0} = \widetilde s\hfill\atop
\scriptstyle{{\widetilde a}_0} = \widetilde{a}\hfill}}}\left( R_\tau \right)} \right],   \label{eq:pi_opt}\\
\text{s.t.}\nonumber\\
R_\tau &= {\sum\limits_{k = 1}^{{n_T}} {{\gamma ^{k - 1}}{{\tilde r}_k}} },\label{eq:tot_R_dp}\\
\tau  &= \left( {{{\widetilde s}_0},{{\widetilde a}_0},{{\widetilde r}_1}, \ldots ,{{\widetilde s}_{{n_T} - 1}},{{\widetilde a}_{{n_T} - 1}},{{\widetilde r}_{{n_T}}},{{\widetilde s}_{{n_T}}}} \right),\label{eq:tau_dp}\\
{{\widetilde a}_k} &\sim \pi \left( \cdot{\backslash {{\widetilde s}_k}} \right) \subset {\mathbb{R}^{{n_{\mathcal{A}}}}},\label{eq:ak_dp}\\
\widetilde{s}_k & = g(\mathcal{I})\in \mathcal{C}_o\subset {\mathbb{R}^{{n_{\mathfrak{S}}}}},\label{eq:sk_dp}\\
\widetilde{s}_{k+1} &\sim p(\widetilde{s}_{k},\widetilde{a}_{k}), \label{eq:sk1_dp}\\
\widetilde{r}_k & = \widetilde{r}(\widetilde{s}_{k-1},\widetilde{a}_{k-1},k)\in \mathbb{R}, \label{eq:r_dp} 
    \end{align}
\end{subequations}
where $R_\tau $ is the cumulative reward; $\tau  = \left( {{{\widetilde s}_0},{{\widetilde a}_0},{{\widetilde r}_1}, \ldots ,{{\widetilde r}_{{n_T}}},{{\widetilde s}_{{n_T}}}} \right)$ is MDP trajectory; $\tau^\pi$ and $\rho^\pi$ are the distribution of trajectories and 
state-action marginals of the trajectory distribution induced by the policy $\pi$, respectively; $p({\widetilde s}_k, {\widetilde a}_k)$ is the
state transition probability depending on the dynamics of
the environment; the $g$-map models the relationship between the state
and the set of information $\mathcal{I}$ determining the evolution of $s_k$; $\mathcal{C}_o$ is the set of achievable states; and $n_T$ is the terminal time instant. 

\subsection{Action and Observations} \label{sec:Action_observation}
For the proposed DRL path tracking solution, the action is the  derivative of the steering angle, i.e., $\widetilde{a}=\dot{\delta}\in \left[\dot{\delta}_{\min};\;\dot{\delta}_{\max}\right]$, with $\dot{\delta}_{\min}$ and $\dot{\delta}_{\max}$ being the minimum and maximum control action, respectively. The derivative of the steering angle has been chosen as the control input to generate a smooth variation in the steering input by constraining its rate of change. The vector of the observations/states is:
\begin{equation}\label{eq:observation}
\widetilde{s} = x_e = {\left[ {\begin{array}{*{20}{c}}
{\Delta y}&{\Delta \dot y}&{\Delta \psi}&{\Delta r}
\end{array}} \right]^T},
\end{equation}
where $\Delta y$, $\Delta \psi$ and $\Delta r$ are the lateral error (i.e., the distance of the center of the mass of the robotic vehicle to the reference path), heading angle error and the yaw rate error of the vehicle, respectively. These errors in relation to the reference path are defined in Fig.~\ref{fig:path_errors_schematic} and analytically computed as: 
\begin{subequations}\label{eq:state_observation}
    \begin{align}
        \Delta y &= \left(Y-Y_{ref}\right)\cos\left(\psi_{ref}\right)-\left(X-X_{ref}\right)\sin\left(\psi_{ref}\right), \label{eq:delta_y}\\
        \Delta \psi & = \psi - \psi_{ref}, \label{eq:delta_psi}\\
         \Delta r &= \frac{d \Delta \psi}{dt}, \label{eq:delta_r}
    \end{align}
\end{subequations}
where $X_{ref}$ and $Y_{ref}$ are the reference coordinates of the centre of mass of the vehicle in the inertia frame and $\psi_{ref}$ is the reference heading angle. 

The references $X_{ref}$, $Y_{ref}$ and $\psi_{ref}$ are functions of the travelled distance along the path distance $s$ computed by integrating: 
\begin{equation}\label{eq:dot_s}
\dot s =\! \frac{{{v_{x}}\cos (\Delta {\psi _{ref}}) - {v_{y}}\sin (\Delta {\psi _{ref}})}}{{1 \!-\! {\kappa _{ref}}\left[ {\left( {Y \!-\! {Y_{ref}}} \right)\cos ({\psi _{ref}}) \!-\! \left( {X \!-\! {X_{ref}}} \right)\sin ({\psi _{ref}})} \right]}},
\end{equation}
where $v_{x}$, $v_{y}$ are the components of the vehicle speed along the $x$-axis and $y$-axis of the vehicle body frame, respectively, and $\kappa_{ref}$ is the reference curvature.

\subsection{Reward Design} \label{sec:reward}
At each discrete time step $k$ the agent gets the reward $\widetilde{r}_{k}$ computed as:
\begin{equation}\label{eq:reward}
\widetilde{r}_k = \widetilde{r}_{y,k} + \widetilde{r}_{\psi,k} + \widetilde{r}_{\dot{\delta},k} + \widetilde{r}_{\text{ed},k},
\end{equation}
with: 
\begin{subequations}\label{eq:r_decomposition}
    \begin{align}
       {\widetilde{r}_{y,k}} &= \left\{ {\begin{array}{*{20}{l}}
{ - {m_1}\ln (\Delta y_{th}^l ),}&{{\rm{if }}\;\;|\Delta {y_k}| \le \Delta y_{th}^l,}\\
{ - {m_2}\ln (|\Delta {y_k}| ),}&{{\rm{if }}\;\;\Delta y_{th}^l < |\Delta {y_k}| < \Delta y_{th}^u},\\
{ - M,}&{{\rm{if }}\;\;|\Delta {y_k}| \ge \Delta y_{th}^u,}
\end{array}} \right.\label{eq:r_y}\\
{\widetilde{r}_{\psi ,k}} &= \left\{ {\begin{array}{*{20}{l}}
{ - {m_3}\ln \left( {\Delta {\psi _{th}}  } \right),}&{{\rm{if }}\;\;|\Delta {\psi _k}| \le \Delta {\psi _{th}},}\\
{ - {m_4}\ln (|\Delta {\psi _k}|  ),}&{{\rm{if }}\;\;|\Delta {\psi _k}| > \Delta {\psi _{th}},}
\end{array}} \right. \label{eq:r_psi}\\
\widetilde{r}_{\dot{\delta},k} &= -m_5 |\dot{\delta}|, \label{eq:r_delta}\\
\widetilde{r}_{\text{ed},k} &= -m_6|\delta_{\text{ed},k}-\delta_k|, \label{eq:r_ed}
    \end{align}
\end{subequations}
where the thresholds (i.e., ${\Delta y_{th}^l}$, ${\Delta y_{th}^u}$, and ${\Delta \psi_{th}}$) and the weights (i.e., $M$, $m_j$, with $j=1,\ldots 6$) are positive constants, and $\delta_{\text{ed},k}$ is the action provided by the expert demonstrator at the time instant $k$. 

\noindent\textbf{Remarks}
\begin{itemize}
\item The component $\widetilde{r}_{y}$ in \eqref{eq:r_y} weighs the mismatch between the actual vehicle position and the reference one.  The agent receives positive rewards 
for actions keeping the lateral error within $\left[{\Delta y_{th}^l}\;\; {\Delta y_{th}^u}\right]$ and these rewards increase when $\Delta y$ decreases. In doing so, the reduction of the lateral error is promoted. However, when the lateral error is below the threshold ${\Delta y_{th}^l}$ the reward is still positive but constant. This is used for avoiding solutions which should provide a very precise tracking of the reference path but is not experimentally feasible because of measurement errors. Limiting the reward for low values of the lateral error also reduces the training time as there is no convenience in exploring policies that shrink the error below ${\Delta y_{th}^l}$. Finally, the agent gets a large negative reward (i.e., $-M$) when the lateral error exceeds  ${\Delta y_{th}^u}$ which also implies the early termination of the episode. This condition is used to reduce the learning time by avoiding exploration of actions that give unacceptable lateral errors. Furthermore, it promotes the seeking of solutions where the agent first learns to keep the vehicle sufficiently close to the path and subsequently learns how to finely follow the reference.
\item The sub-reward $\widetilde{r}_\psi$ in \eqref{eq:r_psi} aims to promote actions which lead to heading errors towards ${\Delta \psi_{th}}$. However,  similar to the reward $\widetilde{r}_{y}$, for heading errors below ${\Delta \psi_{th}}$,  the reward $\widetilde{r}_\psi$ is kept constant for speeding up the training by avoiding solutions that cannot experimentally provide the same level of accuracy  because of the residual noise in the heading angle estimates. 
\item The contribution $\widetilde{r}_{\dot{\delta}}$ in \eqref{eq:r_delta} is a penalty on the control input and is used to guide the training towards solutions with smooth steering angles, thus avoiding chattering phenomenon which could damage  and/or reduce the lifespan of  the steering actuation system. Furthermore, smooth steering angle dynamics results in a smoother motion of the vehicle. 
\item The term $\widetilde{r}_\text{ed}$ in \eqref{eq:r_ed} penalises the mismatch between the agent steering angle and that provided by the expert's demonstrator with the aim to guide the learning process quickly towards an initial acceptable control solution obtained, for instance, via well established control theories for vehicle path tracking. The contribution of the terms \eqref{eq:r_y}-\eqref{eq:r_delta} in the reward function \eqref{eq:reward} will then  lead to an optimal control policy $\pi^\star$ which will differ from that of an expert. The deviation between the optimal policy and that provided by the expert demonstrator can be adjusted through the tuning of the weights $m_j$, with $j=1,\ldots,6$. In other words, the use of \eqref{eq:r_y}-\eqref{eq:r_delta} within the proposed DRL-based path tracking control design opens the way of improving the closed-loop performance of pre-existing controllers not only in terms of tracking of the reference trajectory but also with respect to the aggressiveness of the control strategy and control effort. In this work the expert demonstrator's policy is the optimal LQ control action proposed for automated vehicles in [43]. 
\end{itemize}

The thresholds and the weights of the reward functions~\eqref{eq:r_decomposition} have been selected heuristically to obtain the satisfactory closed-loop performance in simulation.

\subsection{DRL Implementation Details} \label{sec:DRL_implementation}
\begin{figure}[t]
     \centering
     \subfloat[]
     {\includegraphics[width=0.25\textwidth]{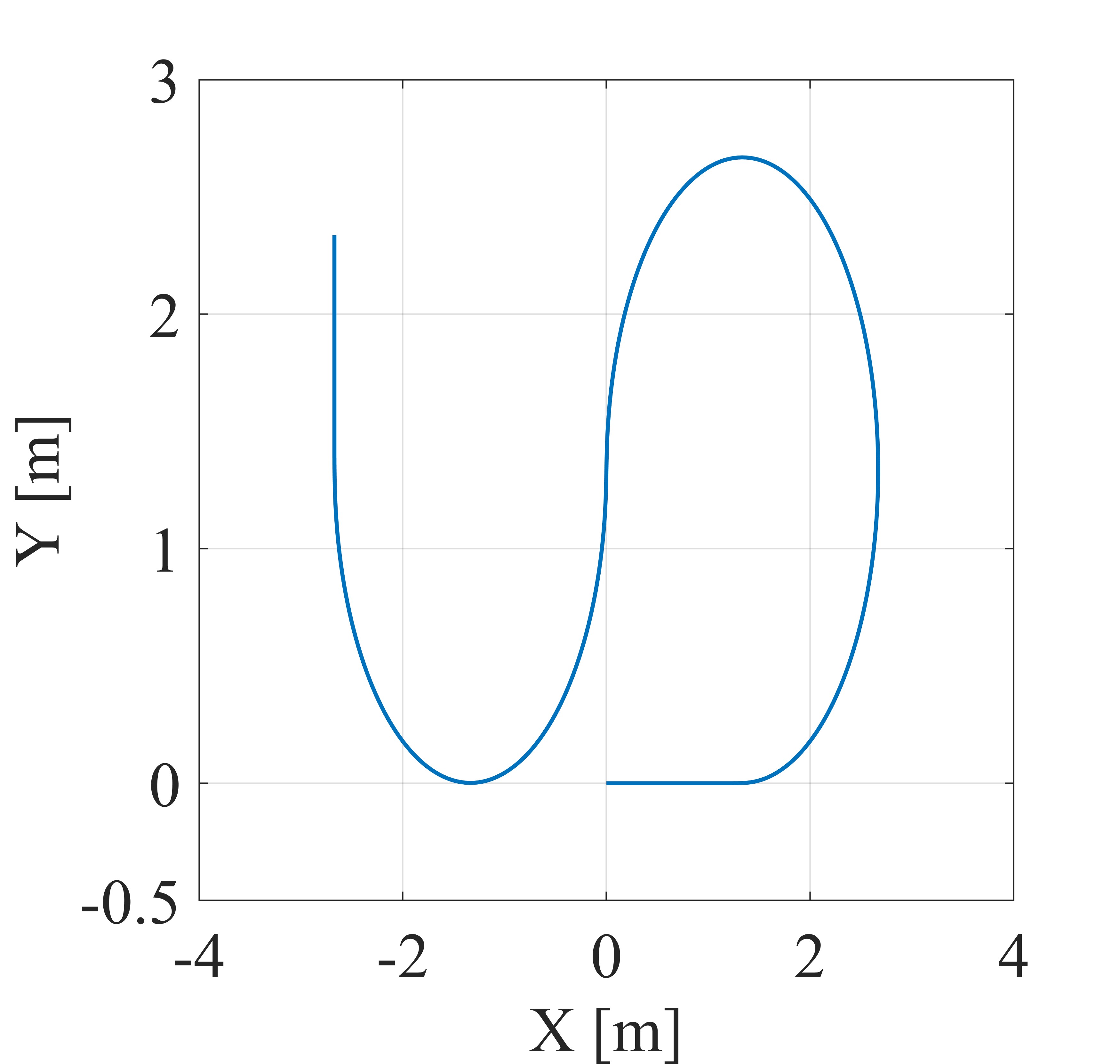}\label{fig:S_training}}
     \subfloat[]
          {\includegraphics[width=0.25\textwidth]{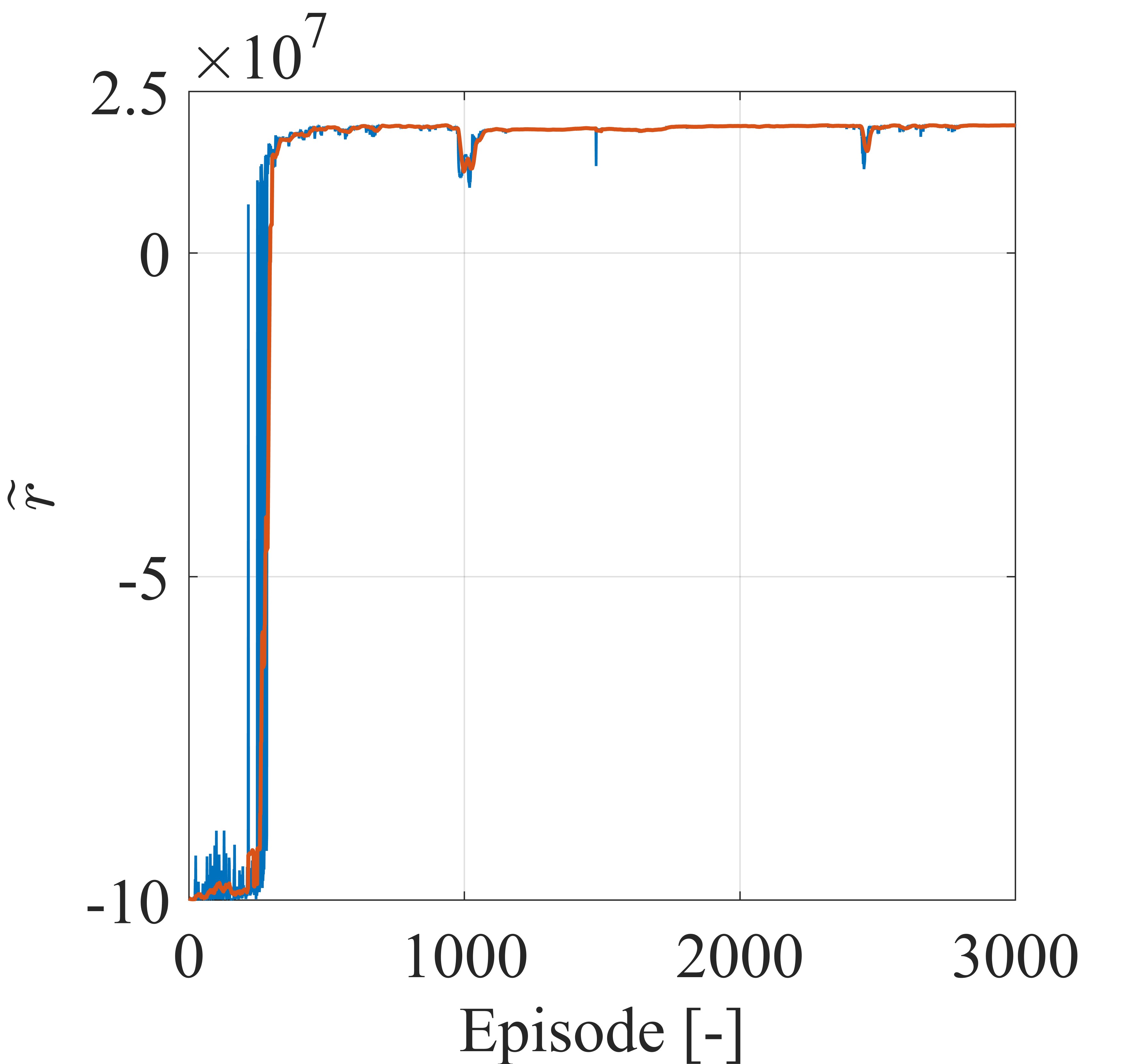}\label{fig:reward}}
     \caption{DRL training: (a) reference path; (b) evolution of the cumulative reward: value for each episode (blue line) and mean value (red line).}
\label{fig:DRL_sim}
\end{figure}

The proposed DRL-based path tracking framework has been implemented with the Reinforcement Learning Toolbox available in Matlab 2022b [39]. The actor-critic (DDPG) algorithm was used for the training phase because of its good balance between sample efficiency and computational effort during training [41],[42]. The actor network, i.e., the network that computes the control input from the states, has two ReLu hidden layers with 200 neurons each, and one tanh output layer with one neuron. The critic network, i.e., the one that evaluates the performance of the actor during training, has two ReLu hidden layers with 200 neurons each for the state path, and two ReLu hidden layers with 100 and 200 neurons each for the action path. The controller sampling time is $10$ ms. 
The model of the robotic vehicle parametrised as in Section ~\ref{sec:model_identification_validation} has been used \emph{(i)} for the training as a digital twin for emulating the environment; and \emph{(ii)} for the generation of the steering action $\delta_{\text{ed}}$ of the expert demonstrator in accordance with the LQ model-based control solution presented in [43]. Furthermore, the training has been carried out by using an $\mathcal{S}$-shape path (see also Fig. \ref{fig:S_training}) when the vehicle longitudinal speed is set as $0.5$ m/s which is larger than those usually adopted for the same robotic ground vehicle [44].


Fig.~\ref{fig:reward} shows the convergent of the cumulative reward which occurs in about 300 episodes.

\begin{figure}[t]
    \centering
    \includegraphics[width=0.4\textwidth]{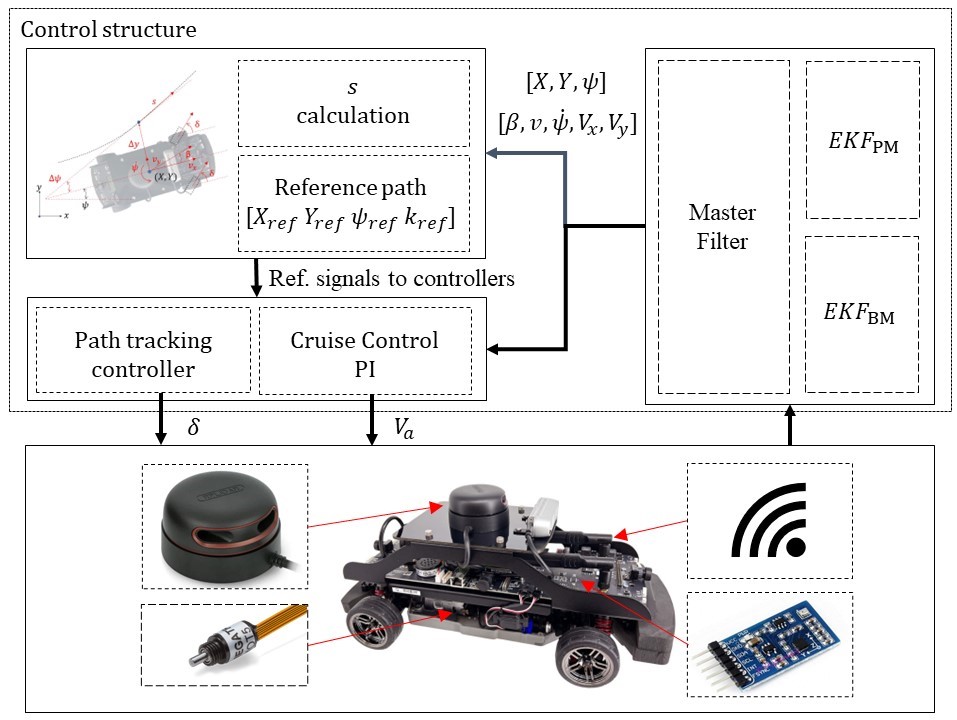}
    \caption{Schematic of the control architecture.}
    \label{fig:control_scheme}
    \vspace{-3mm}
\end{figure}

The control architecture hosting the trained agent as a path tracking controller is shown in Fig. \ref{fig:control_scheme} and it also includes: \emph{(i)} the proposed FEKF in Section \ref{sec:EKF_desing}, \emph{(ii)} the reference path generator whose outputs depend on the travelled distance $s$, and \emph{(iii)} a PI controller used to impose the desired speed over the manoeuvres.

\section{DRL Path Tracking Experimental Results}\label{sec:DRL_validation}
\begin{figure*}[t]
     \centering
     \subfloat[]
     {\includegraphics[width=0.33\textwidth]{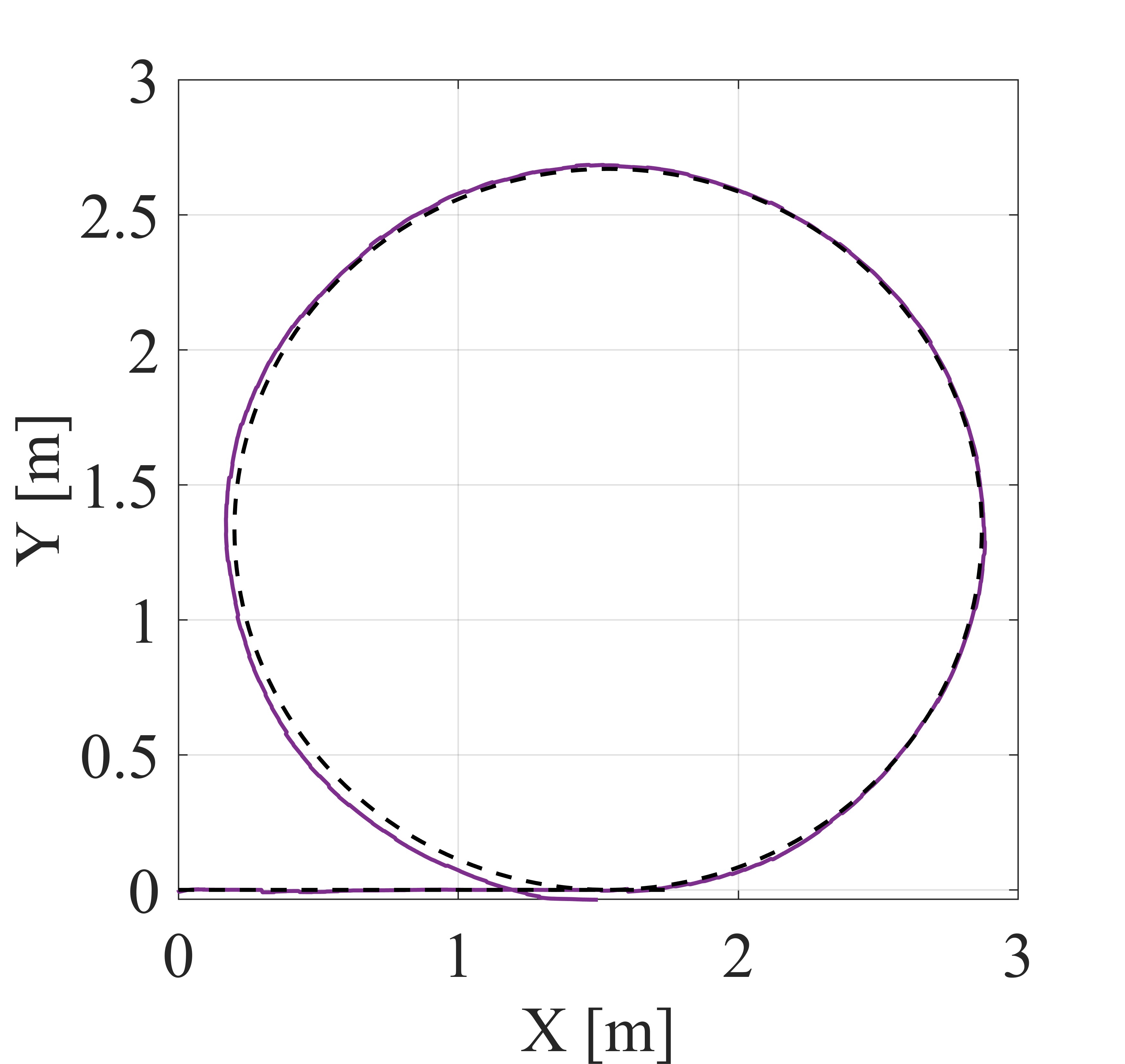}\label{fig:O_shape}}
     \subfloat[]
          {\includegraphics[width=0.33\textwidth]{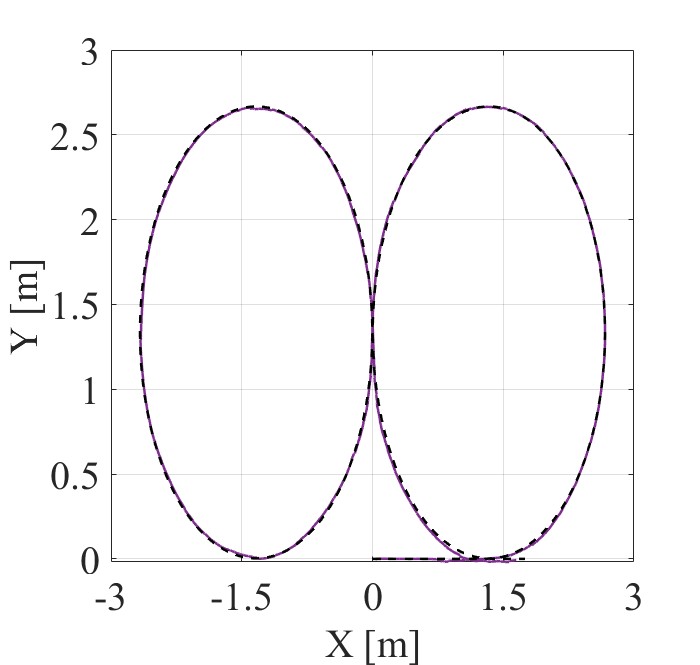}\label{fig:E_shape}}
      \subfloat[]
          {\includegraphics[width=0.33\textwidth]{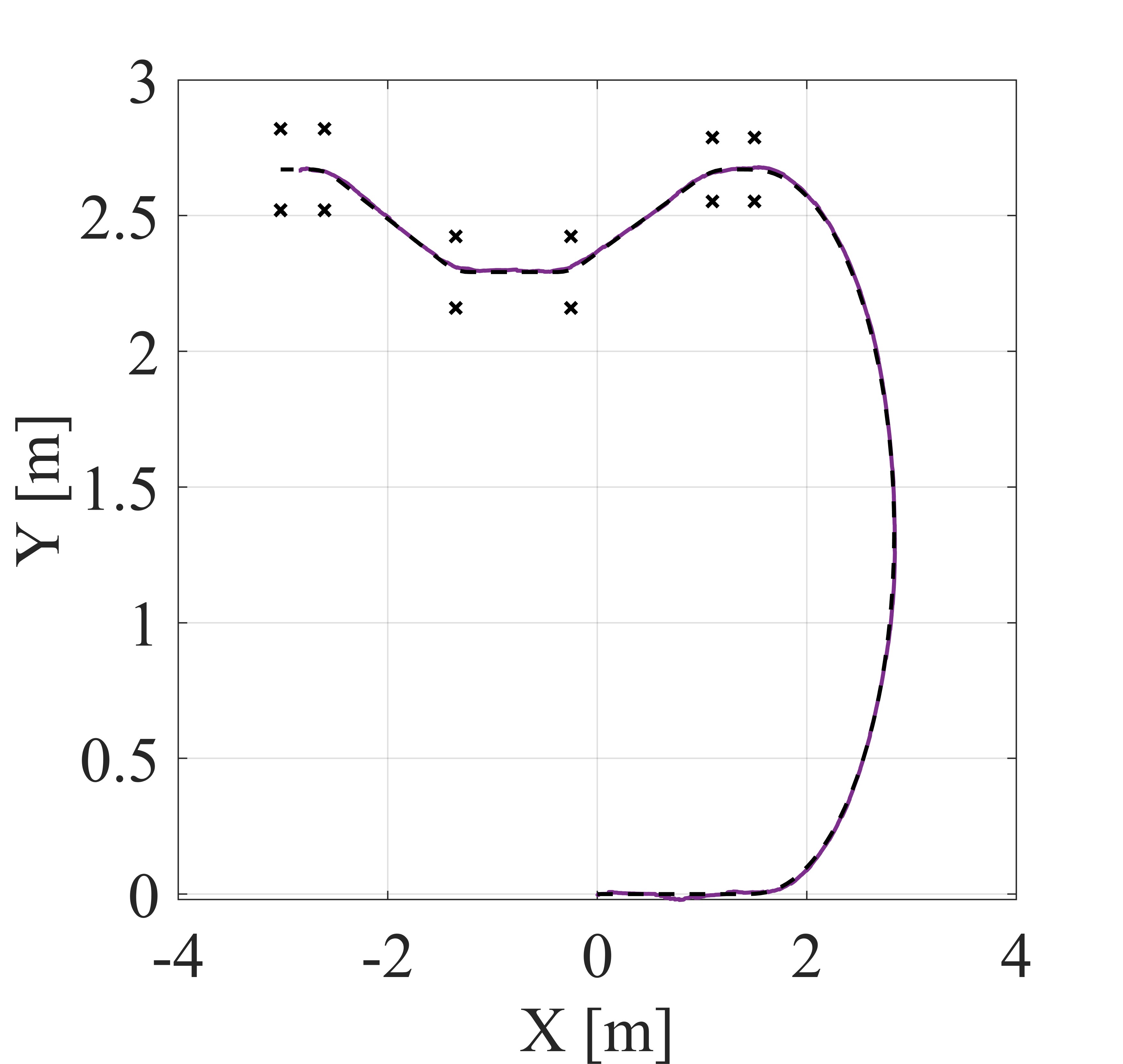}\label{fig:C_shape}}
     \caption{Experimental results of the DRL-based path tracking - reference path (dashed black line) and actual path obtained via the DRL agent (solid purple line):  (a) $\mathcal{O}$-shape, (b) $\infty$-shape, and (c) $\mathcal{C}$-shape.}
\label{fig:DRL_exp_trajectories}
\end{figure*}

The DRL-based path tracking strategy is tested on three scenarios not considered during the training  of the agent (i.e., three validation tests are considered). The selected paths are depicted in Fig.~\ref{fig:DRL_exp_trajectories} and are denoted in the rest of the paper as:  \emph{(i)} $\mathcal{O}$-shape, \emph{(ii)} $\infty$-shape and \emph{(iii)} $\mathcal{C}$-shape. Specifically, the $\mathcal{C}$-shape is a manoeuvre composed first by \emph{(i)} a U-turn, followed by \emph{(ii)} a collision avoidance manoeuvre for passengers cars regulated by the standard ISO 3888-2 but scaled by a factor $10$ to consider the size of the robotic vehicle. Fig. \ref{fig:DRL_exp_trajectories} also shows that the agent provides experimentally a precise tracking of the reference paths used for validation despite it has been  trained in simulation.

The closed-loop tracking performance of the DRL solution has been compared to those provided by \emph{(i)} a feed-forward feedback (FF-FB) strategy designed as in [43]; \emph{(ii)} the LQ approach (denoted as LQ$_\text{ed}$) used for generating the actions of the expert demonstrator during training  (see also Section \ref{sec:DRL_implementation}) and proposed in [43]; and \emph{(iii)} the discrete time LQ strategy equipped with a compensating feed-froward action (denoted  as LQ$_\textbf{cm}$) for having zero steady-state lateral error during constant cornering proposed in [37].  Notice that  the control architecture in Fig. \ref{fig:control_scheme} has been also used for the implementation of these benchmark path tracking controllers which have been implemented with a sampling time of $10$ ms. 

\begin{figure}[t]
     \centering
     \subfloat[]
     {\includegraphics[width=0.25\textwidth]{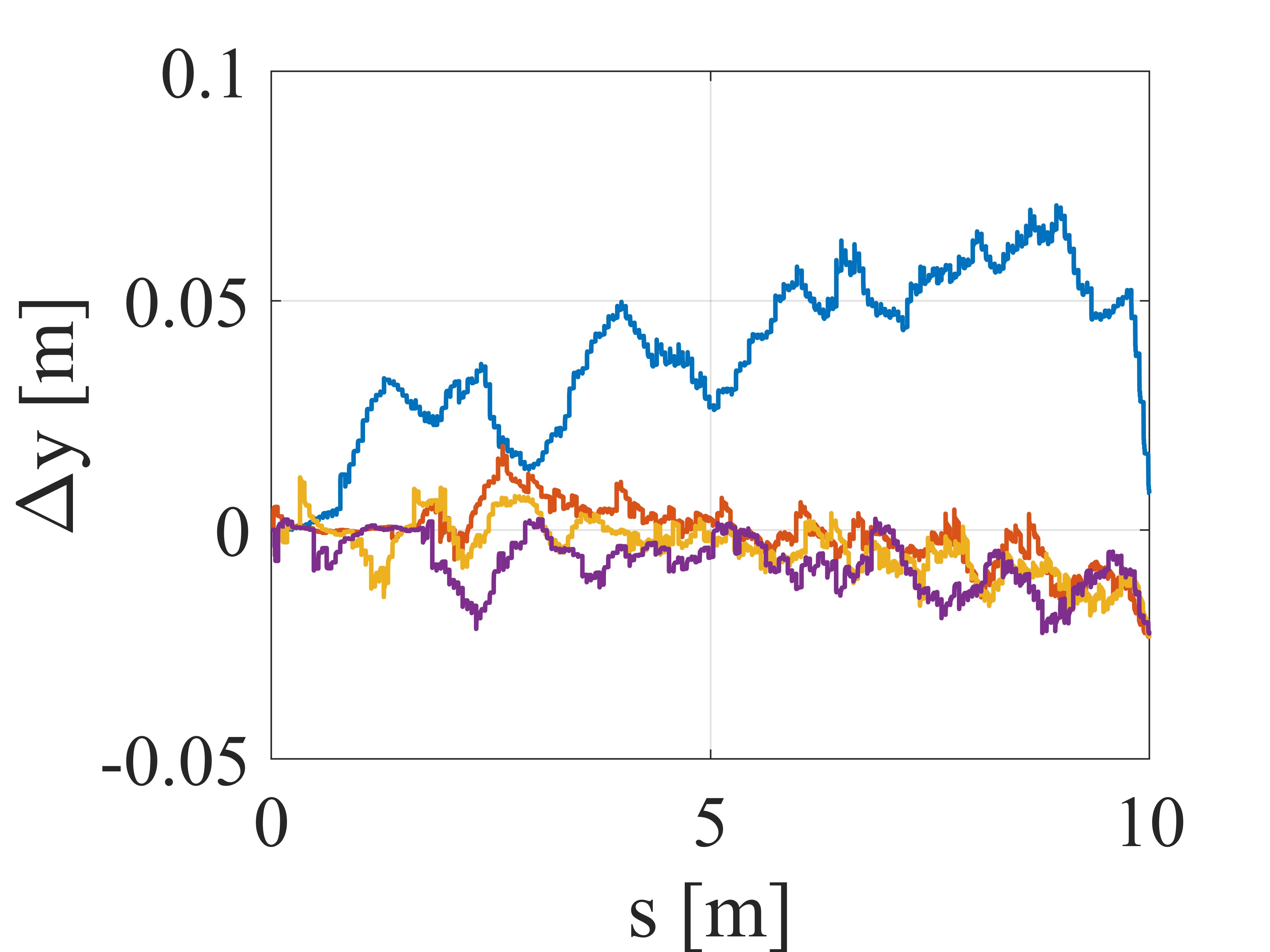}\label{fig:O_shape_Dy}}
     \subfloat[]
          {\includegraphics[width=0.25\textwidth]{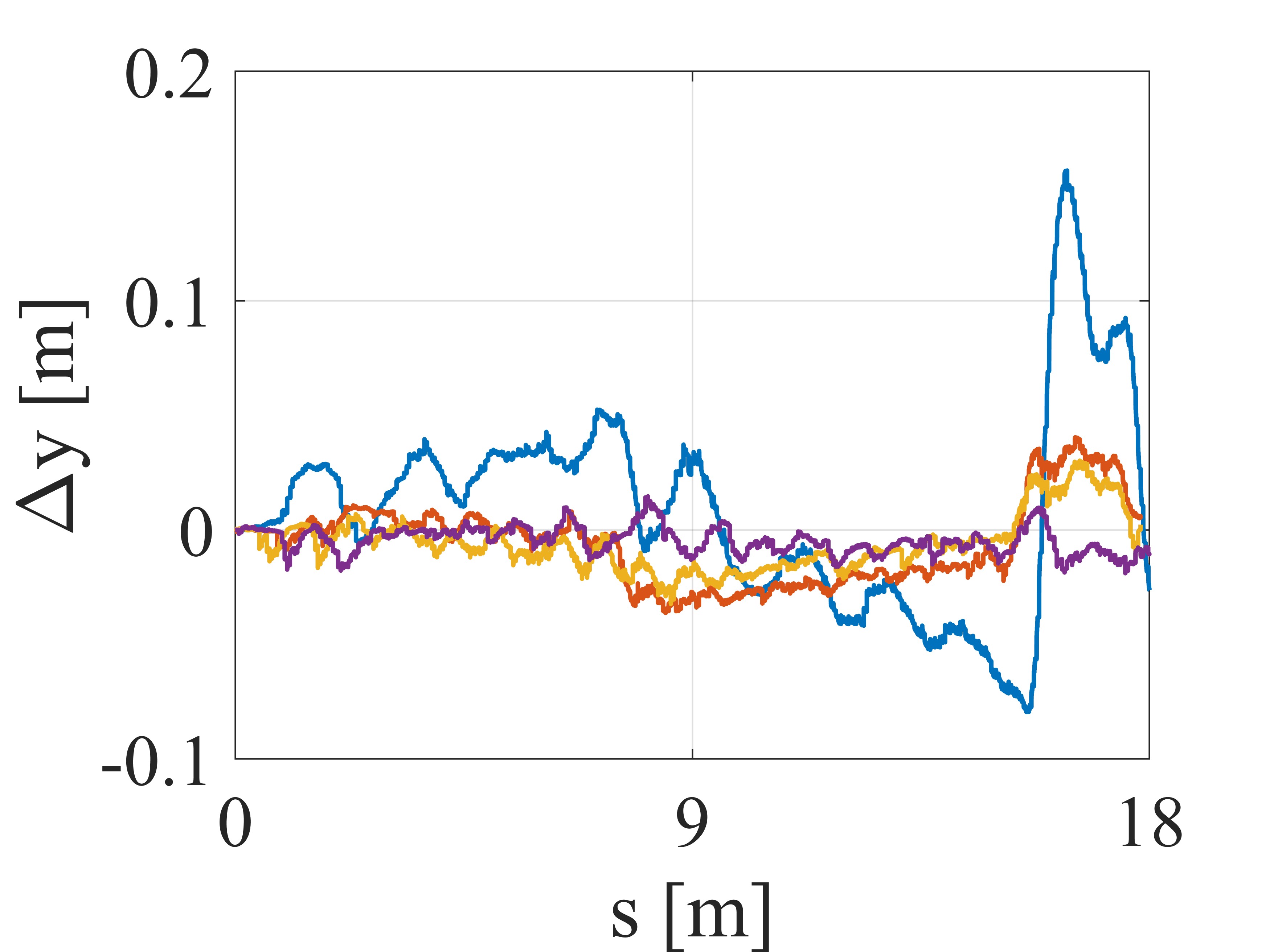}\label{fig:E_shape_Dy}}\\
      \subfloat[]
          {\includegraphics[width=0.25\textwidth]{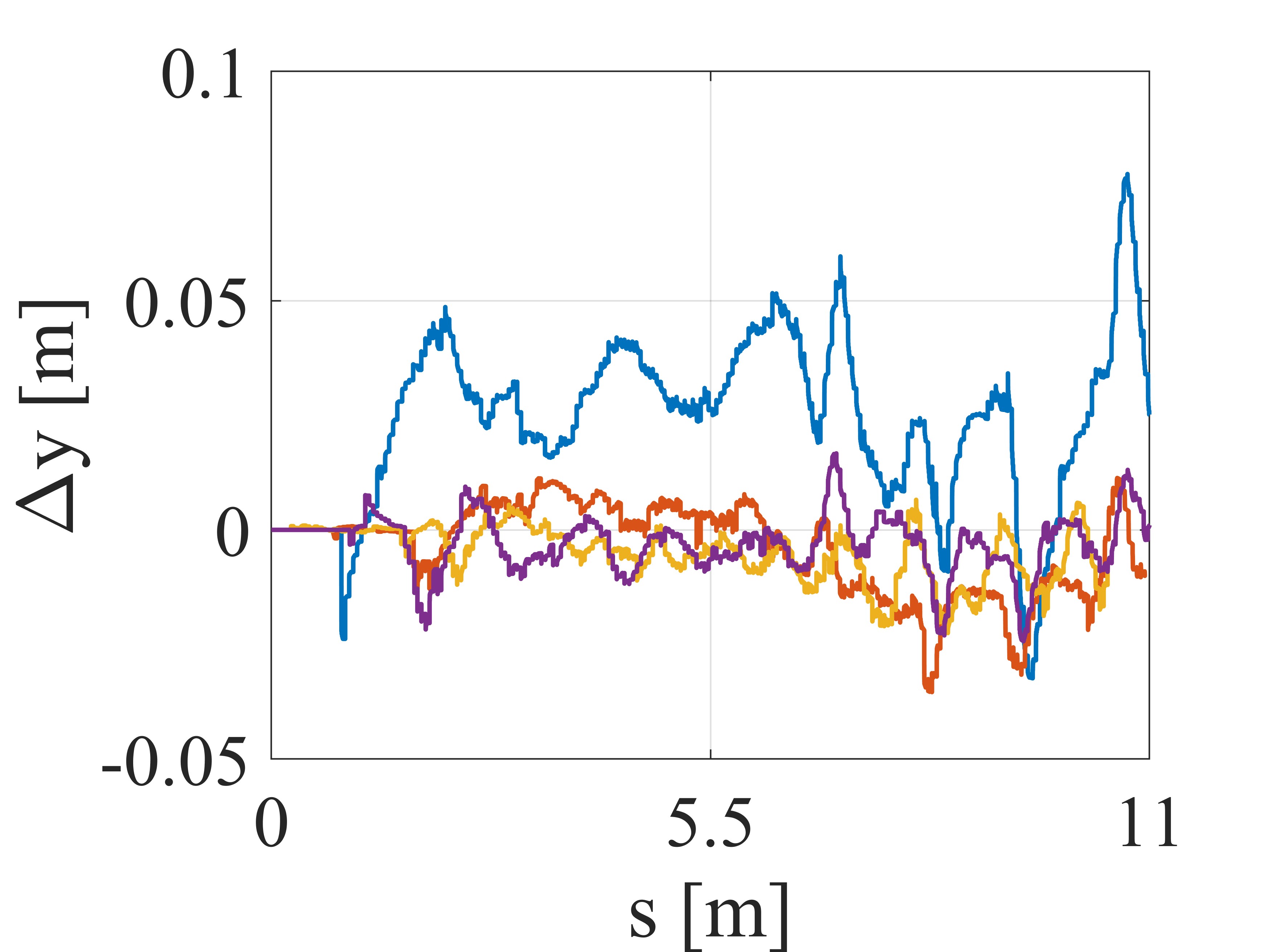}\label{fig:C_shape_Dy}}
     \caption{Lateral error experimental dynamics provided by the FF-FB  (blue line), the LQ$_\text{ed}$ (red line), the LQ$_\text{cm}$ (yellow line) and the DRL-based solution (purple line) for the (a) $\mathcal{O}$-path, (b) $\infty$-path, and (c) $\mathcal{C}$-path.}
\label{fig:DRL_exp_Dy}
\end{figure}

\begin{figure}[t]
    \centering
    \includegraphics[width=0.3\textwidth]{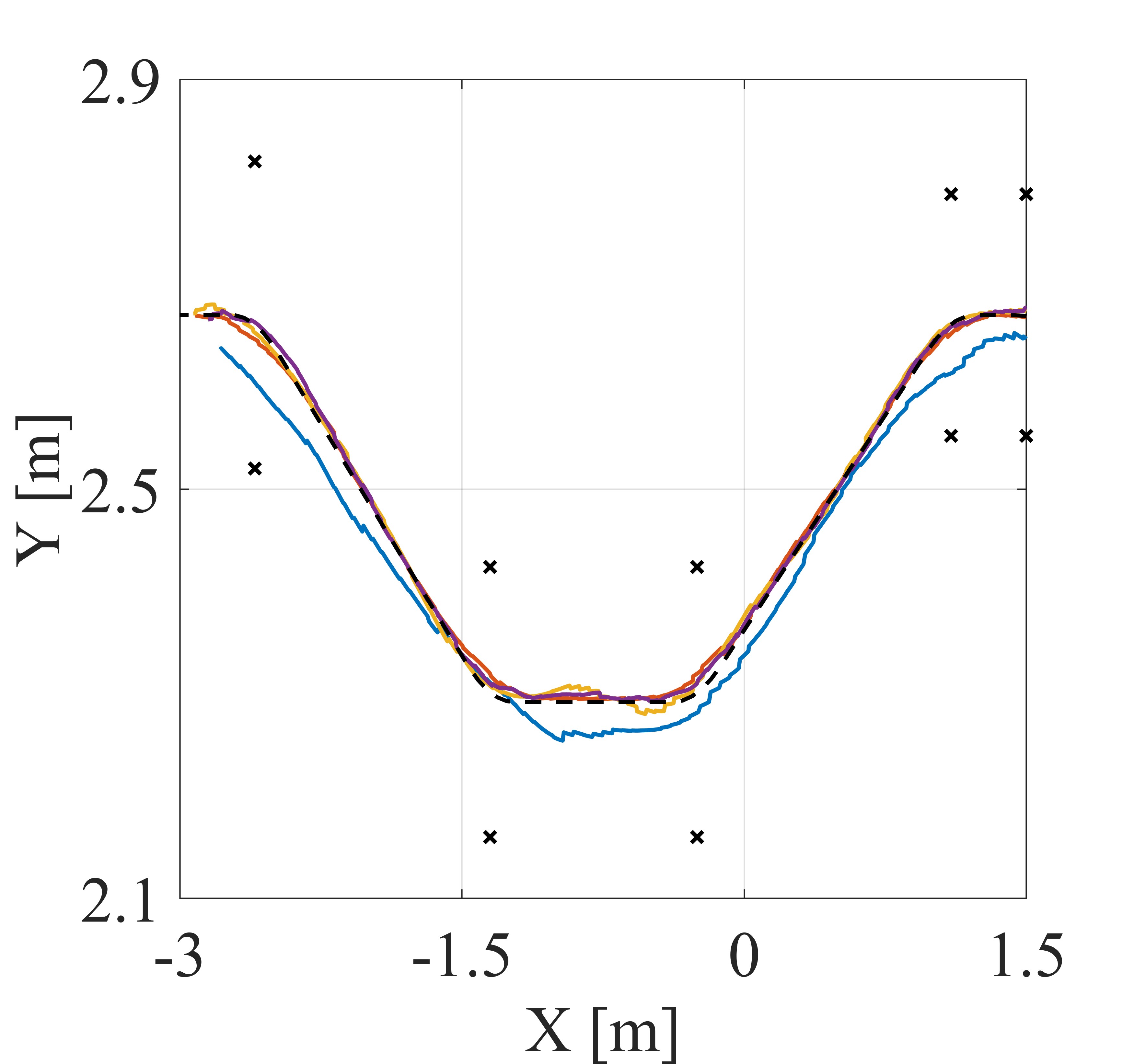}
    \caption{Experimental results for the collision avoidance test within the $\mathcal{C}$-shape path: reference path (dashed black line), FF-FB strategy (blue line), LQ$_\text{ed}$ algorithm (red line),  LQ$_\text{cm}$ algorithm (yellow line), and DRL-based solution LQ$_\text{cm}$ (purple line).}
    \label{fig:all_Ca_path}
\end{figure}

Fig. \ref{fig:DRL_exp_Dy} depicts the lateral error dynamics obtained for each manoeuvre and shows that:
\begin{itemize}
\item  the DRL-based solution outperforms the FF-FB strategy in terms of residual lateral error; 
\item the LQ$_\text{ed}$ and the LQ$_\text{cm}$ strategies provide similar closed-loop  behaviour;
\item the DRL-based control solution provides  comparable performance to the LQ algorithms for the $\mathcal{O}$-shape path and the U-turn within the $\mathcal{C}$-shape manoeuvre;
\item the DRL-based algorithm provides smaller residual lateral errors for the $\infty$-shape path and the collision avoidance manoeuvre included in the $\mathcal{C}$-shape manoeuvre (i.e., for $s>6$ m);
\item only the LQ-based and the DRL-based strategies pass the collision avoidance manoeuvre without hitting any of the edges of the two borders delimiting above and below the reference path (see also Fig.~\ref{fig:all_Ca_path}). The borders in Fig.~\ref{fig:all_Ca_path} are those described by the standard ISO 3888-2 but scaled of a factor $10$. 
\end{itemize}

\begin{figure}[t]
    \centering
    \includegraphics[width=0.3\textwidth]{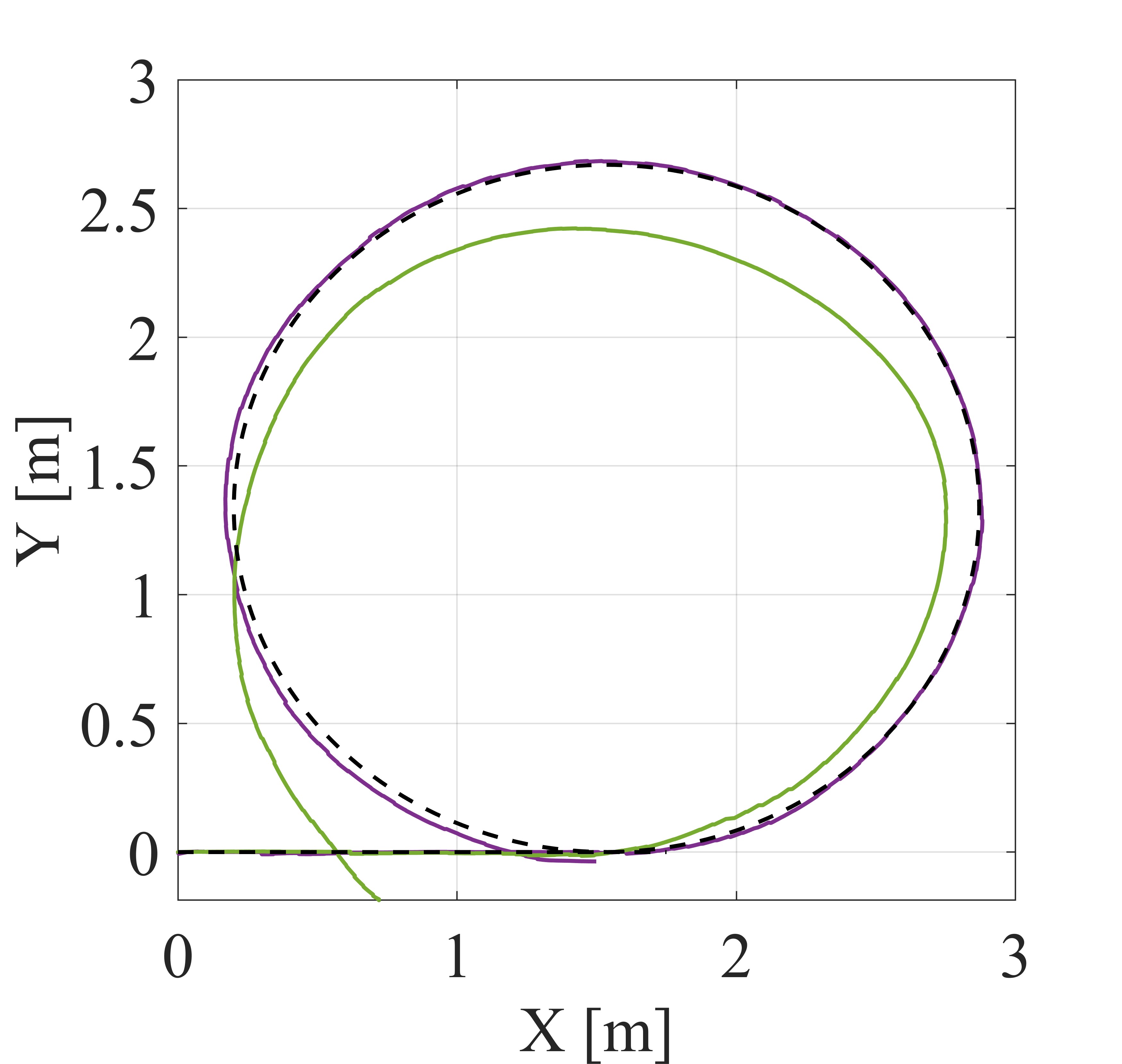}
    \caption{Experimental results for the $\mathcal{O}$-path: reference path (dashed black line) and actual path obtained via the DRL agent trained with the information of the expert demonstrator (purple line) and without the expert demonstrator (green line). }
    \label{fig:DRL_comparison}
\end{figure}

To show the impact of the information provided by the expert demonstrator on the closed-loop tracking performance, a second agent has been trained in accordance with the approach presented in Section \ref{sec:DRL_desing} but without the reward term $\widetilde{r}_{\text{ed}}$ in the reward function \eqref{eq:reward} (i.e., without considering the information of the expert demonstrator). Fig. \ref{fig:DRL_comparison} shows that the agent trained without the $\widetilde{r}_{\text{ed}}$-term is not able to follow the $\mathcal{O}$-path with the same precision of the agent trained by embedding the expert demonstrator's policy. Hence, the use of the expert demonstrator helps to find a more robust agent that can be used on paths different from those used during training, thus also mitigating the simulation-to-reality gap. 

%

\subsection{Analysis of KPIs}\label{sec:KPIs}
To quantitatively compare the path tracking control solutions the maximum error (ME), the root mean square error (RMSE) of the lateral error and the integral of absolute value of control action (IACA) are used. These key performance indicators (KPIs) are defined as: 
\begin{subequations}\label{eq:KPI_def}
    \begin{align}
       \text{ME}   &= \mathop {\max }\limits_{t \in \left[ {0,\;{T_f}} \right]} \left| \Delta y (t) \right|,\label{eq:KPI_ME}\\
\text{RMSE} &= \sqrt {\frac{1}{{{T_f}}}\int_0^{{T_f}} {\left| {\Delta y {{(t)}}} \right|^2dt } } ,\label{eq:KPI_RMSE}\\
\text{IACA} &= \frac{1}{{{T_f}}}\int_0^{{T_f}} {\left| {\delta (t)} \right|dt}, \label{eq:KPI_IACA}
    \end{align}
\end{subequations}
where $T_f$ is the duration of the manoeuvre. 

\begin{figure}[t]
     \centering
     \subfloat[]
     {\includegraphics[width=0.25\textwidth]{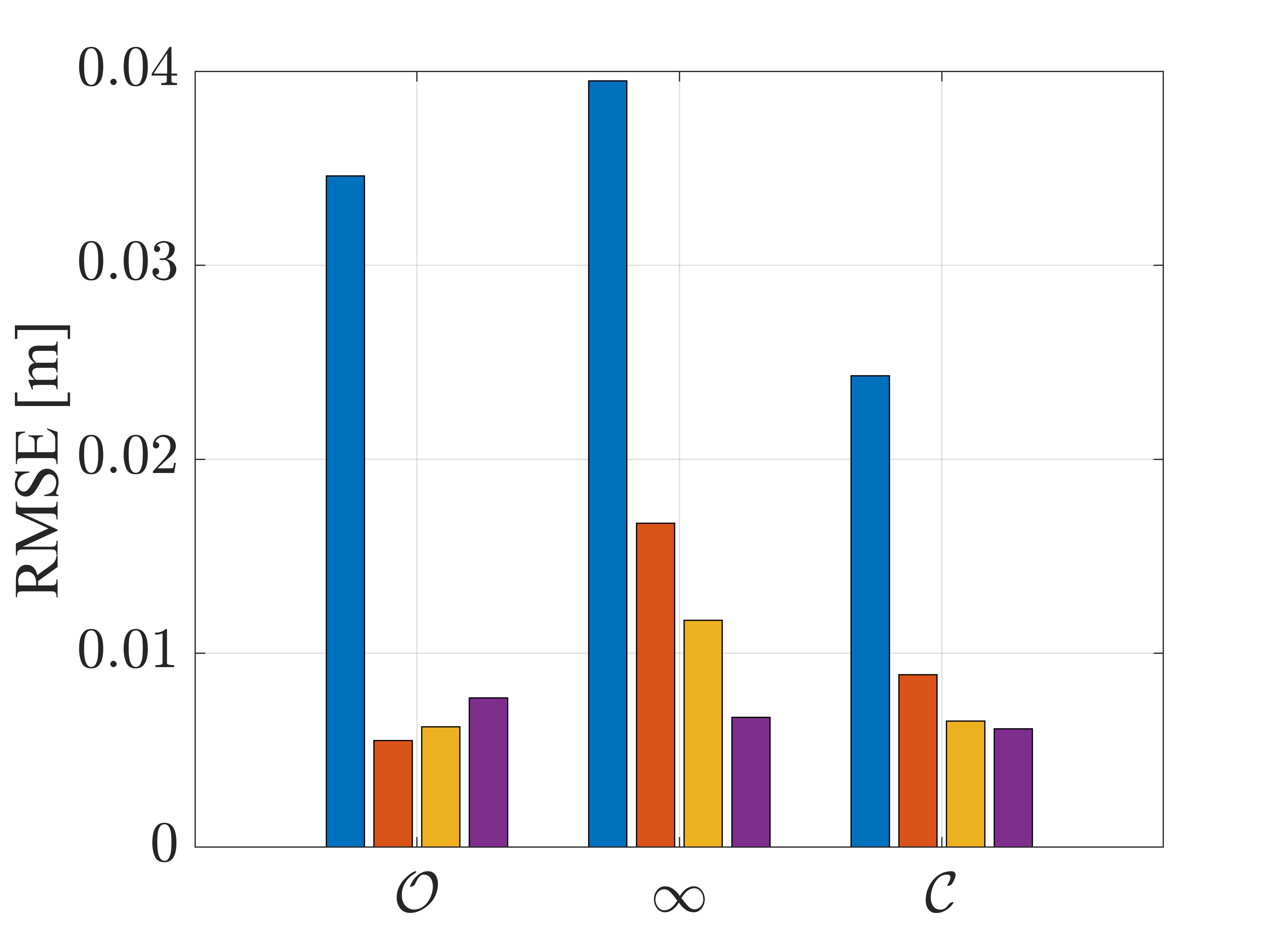}\label{fig:EMSE_Dy}}
     \subfloat[]
          {\includegraphics[width=0.25\textwidth]{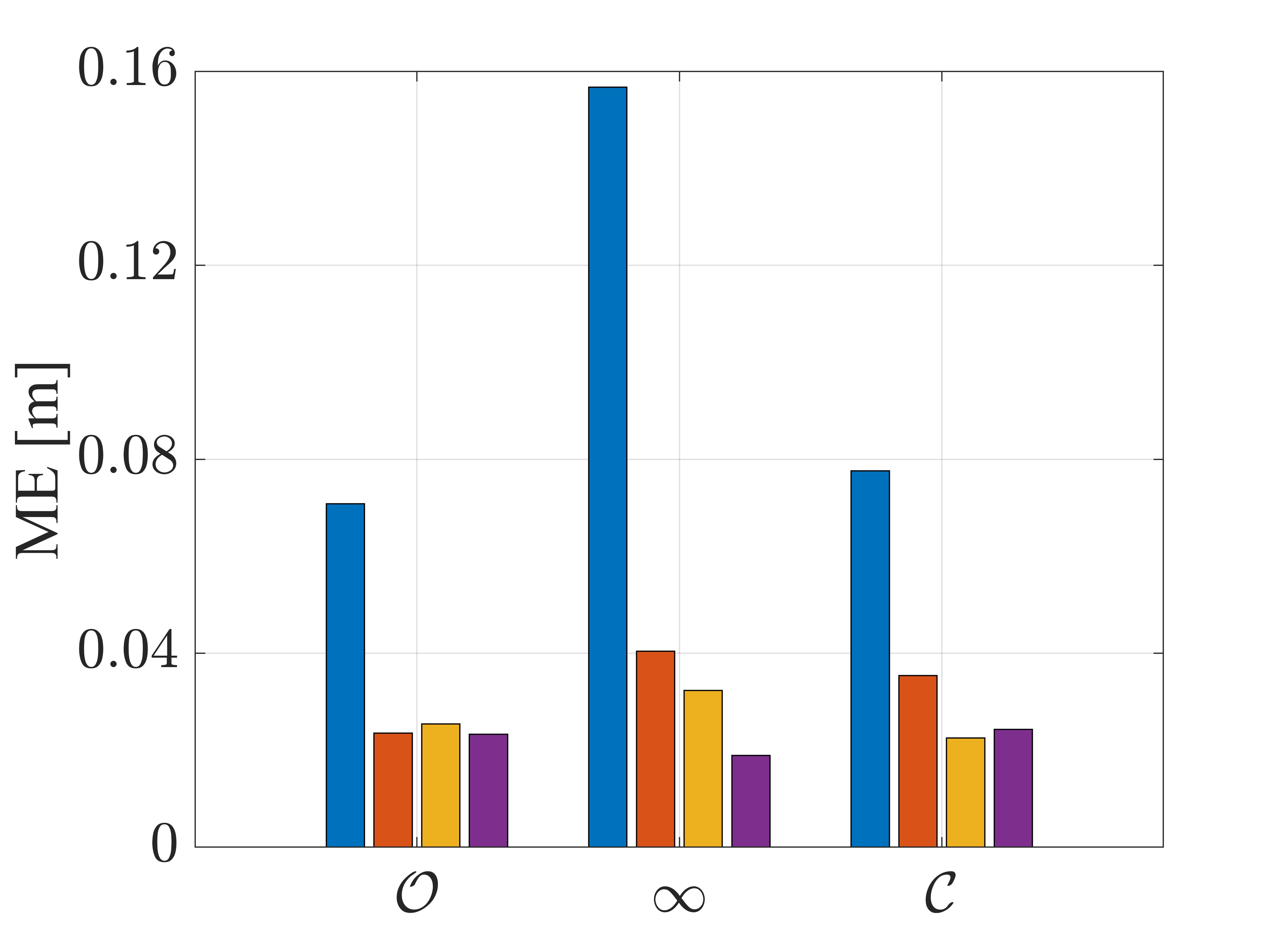}\label{fig:fig:ME_Dy}}\\
     \subfloat[]
     {\includegraphics[width=0.25\textwidth]{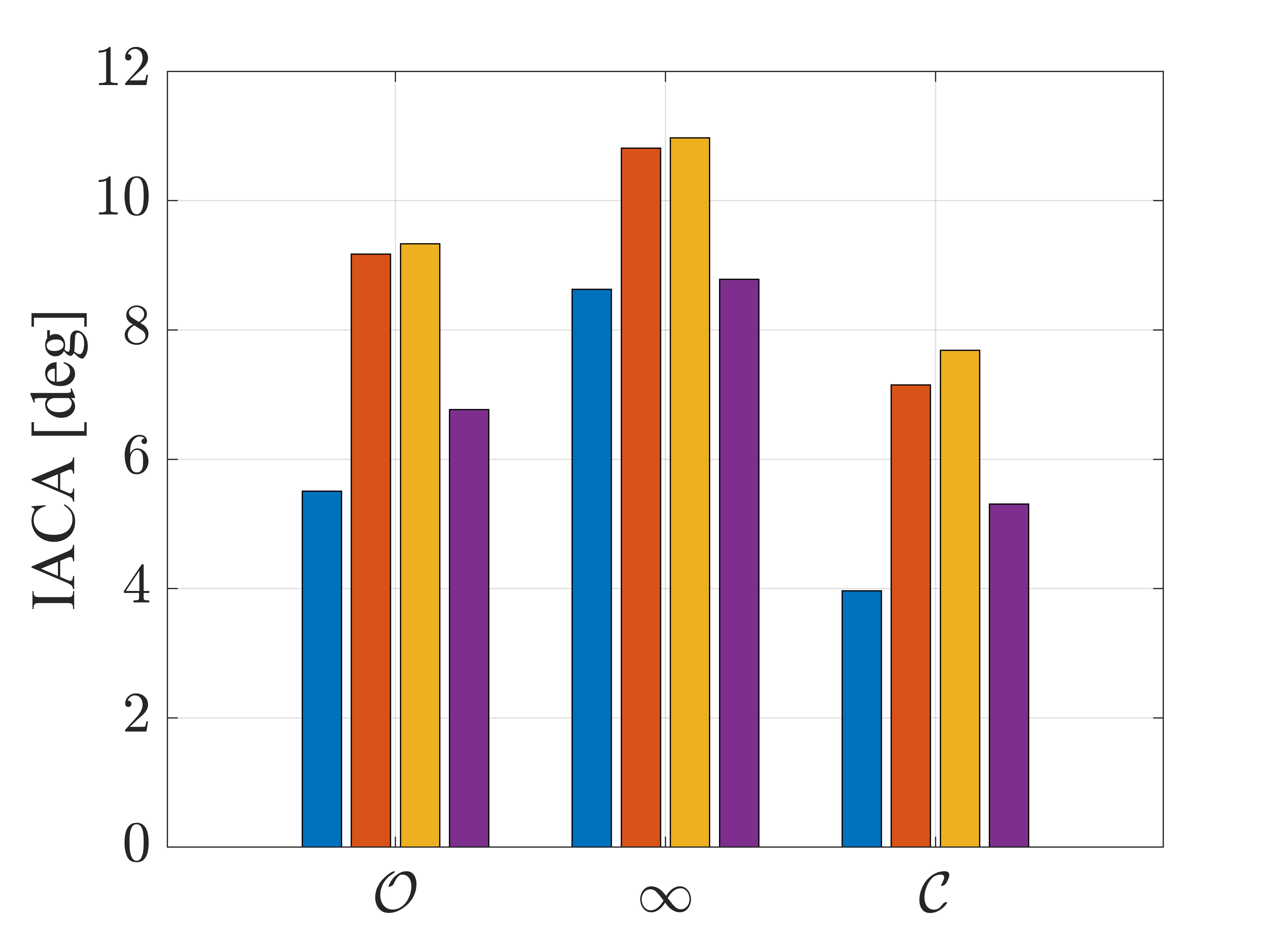}\label{fig:IACA}}         
     \caption{Experimental results for the key performance indicators: (a) RMSE, (b) ME, and (c) IACA: FF-FB  (blue bar), the LQ$_\text{ed}$ (red bar), the LQ$_\text{cm}$ (yellow bar) and the DRL-based solution (purple bar).}
\label{fig:KPIs}
\end{figure}

Fig. \ref{fig:KPIs} shows the KPIs for the benchmarking solutions and the DRL-based approach and confirms:
\begin{itemize}
\item the DRL-based control solution substantially reduces the ME and RMSE compared to the FF-FB strategy with an acceptable increase of the control effort. The DRL solution increases the IACA by a maximum of $26.26$\% compared to the FF-FB algorithm (i.e., for the $\mathcal{C}$-shape path). However, for the $\infty$-shape manoeuvre the DRL shrinks the RMSE and ME of about $5.9$ and $8.3$ times, respectively, with an increment of the control effort of $1.8$\%;
\item  compared to the LQ-based strategies, the DRL-based path tracking solution requires a smaller control effort (i.e., a smaller IACA). The reduction of the IACA ranges from $19$\% (w.r.t. the LQR$_\text{ed}$ over the $\infty$-path) to $31$\%  (w.r.t. the LQR$_\text{cm}$ over the $\mathcal{C}$-path);
\item in terms of  RMSE and ME, the DRL-based solution provides comparable performance w.r.t the LQ$_\text{cm}$ strategy for the $\mathcal{O}$-shape and $\mathcal{C}$-shape manoeuvres, with an acceptable increase in the RMSE for the $\mathcal{O}$-shape manoeuvre; 
\item when compared to the LQ$_\text{ed}$ approach, the DRL solution provides a comparable ME and an increase of about $2$ mm in the RMSE over the $\mathcal{O}$-shape manoeuvre. However, for the $\mathcal{C}$-manoeuvre, the DRL control method reduces the RMSE and ME indices of about $31.5$\% and  $31$\%, respectively;
\item the DRL-based solution  improves the RMSE and ME in the case of the  $\infty$-shape manoeuvre. Specifically, by using the DRL strategy \emph{(i)}  the  RMSE and ME reduce of about $43$\% and $41$\%, respectively w.r.t. those provided by the LQ$_\text{cm}$ strategy;  and \emph{(ii)}  the  RMSE and ME shrink of about $60$\% and $53$\%, respectively w.r.t. those provided by the LQ$_\text{ed}$ solution;
\item the difference in the closed-loop response between the DRL-based solution and the LQ$_\text{ed}$ algorithm is induced by the the reward terms \eqref{eq:r_y}-\eqref{eq:r_delta}. These terms have led to an agent that is able to provide comparable or better tracking performance with respect to the expert demonstrator with a smaller control effort. 
\end{itemize}

\section{Conclusions}  \label{Sec:conclusion}
The paper has presented an effective yet simple-to-replicate method for tuning the parameters of a model for a scaled robotic car. The model has been used: (i) for the design of a filtering system aimed at estimating vehicle states and improving the detection of the vehicle position through a federated extended Kalman filter, and (ii) as a digital twin for the training of a novel deep reinforcement learning path tracking control solution. Experimental results have shown the ability of the FEKF to remove spikes in the lidar measurements thus also opening the way to the implementation of path tracking control algorithms. Furthermore, the experimental study has confirmed that the DRL agent, trained with a reward function embedding the actions of an expert demonstrator, can accurately control the robotic car on paths that were not considered during the training phase. This confirms the ability of the proposed DRL control framework to adapt to various driving scenarios while also mitigating the simulation-to-reality gap issue. Finally, the DRL-based strategy has shown to provide closed-loop tracking performance comparable to or even better than LQ based and classical path tracking control solutions.

\section{References}

\begin{enumerate}[ {[}1{]} ]
\item J. Huang, S. Junginger, H. Liu, and K. Thurow, “Indoor positioning systems of mobile robots: A review,” Robotics, vol. 12, no. 2, 2023.
\item L. Lynch, T. Newe, J. Clifford, J. Coleman, J. Walsh, and D. Toal, “Automated ground vehicle (agv) and sensor technologies- a review,” in 2018 12th International Conference on Sensing Technology (ICST), 2018, pp. 347–352.
\item A. Rupp, M. Tranninger, R. Wallner, J. Zubaca, M. Steinberger, and M. Horn, “Fast and low-cost testing of advanced driver assistance systems using small-scale vehicles,” IFAC-PapersOnLine, vol. 52, no. 5, pp. 34–39, 2019, 9th IFAC Symposium on Advances in Automotive Control AAC 2019.
\item  Y. Zein, M. Darwiche, and O. Mokhiamar, “Gps tracking system for
autonomous vehicles,” Alexandria Engineering Journal, vol. 57, no. 4,
pp. 3127–3137, 2018.
\item A. Ribeiro, M. Koyama, A. Moutinho, E. de Paiva, and A. Fioravanti, “A comprehensive experimental validation of a scaled car like vehicle: Lateral dynamics identification, stability analysis, and control application,” Control Engineering Practice, vol. 116, p. 104924, 2021.
\item L. Liu, X. Wang, X. Yang, H. Liu, J. Li, and P. Wang, “Path planning techniques for mobile robots: Review and prospect,” Expert Systems with Applications, vol. 227, p. 120254, 2023.
\item J. Hu, Y. Zhang, and S. Rakheja, “Adaptive lane change trajectory
planning scheme for autonomous vehicles under various road frictions
and vehicle speeds,” IEEE Transactions on Intelligent Vehicles, vol. 8,
no. 2, pp. 1252–1265, 2023.
\item S. G. Tzafestas, “Mobile robot control and navigation: A global
overview,” Jorunal of Intelligent and Robotis Systems, vol. 91, pp.
35–58, 2018.
\item Q. Li, Z. Chen, and X. Li, “A review of connected and automated
vehicle platoon merging and splitting operations,” IEEE Transactions
on Intelligent Transportation Systems, vol. 23, no. 12, pp. 22 790–
22 806, 2022.
\item E. Yurtsever, J. Lambert, A. Carballo, and K. Takeda, “A survey of autonomous driving: Common practices and emerging technologies,”
IEEE Access, vol. 8, pp. 58 443–58 469, 2019.
\item A. Chalvatzaras, I. Pratikakis, and A. A. Amanatiadis, “A survey on map-based localization techniques for autonomous vehicles,” IEEE
Transactions on Intelligent Vehicles, vol. 8, no. 2, pp. 1574–1596,
2023.
\item S. Kuutti, S. Fallah, K. Katsaros, M. Dianati, F. Mccullough, and
A. Mouzakitis, “A survey of the state-of-the-art localization techniques and their potentials for autonomous vehicle applications,” IEEE Internet of Things Journal, vol. 5, no. 2, pp. 829–846, 2018.
\item K. Tian, M. Radovnikovich, and K. Cheok, “Comparing ekf, ukf,
and pf performance for autonomous vehicle multi-sensor fusion and
tracking in highway scenario,” in 2022 IEEE International Systems
Conference (SysCon), 2022, pp. 1–6.
\item Q. Liu, Y. Mo, X. Mo, C. Lv, E. Mihankhah, and D. Wang, “Secure
pose estimation for autonomous vehicles under cyber attacks,” in 2019
IEEE Intelligent Vehicles Symposium (IV), 2019, pp. 1583–1588.
\item Y. Yin, J. Zhang, M. Guo, X. Ning, Y. Wang, and J. Lu, “Sensor fusion of gnss and imu data for robust localization via smoothed error state kalman filter,” Sensors, vol. 23, no. 7, 2023.
\item K. Zindler, N. Geiß, K. Doll, and S. Heinlein, “Real-time egomotion estimation using lidar and a vehicle model based extended
kalman filter,” in 17th International IEEE Conference on Intelligent
Transportation Systems (ITSC), 2014, pp. 431–438.
\item C. He, C. Tang, and C. Yu, “A federated derivative cubature kalman filter for imu-uwb indoor positioning,” Sensors, vol. 20, no. 12, 2020.
\item M. Kazerooni and A. Khayatian, “Data fusion for autonomous vehicle navigation based on federated filtering,” in 2011 Australian Control Conference, 2011, pp. 368–373.
\item H. Wang, B. Lu, J. Li, T. Liu, Y. Xing, C. Lv, D. Cao, J. Li,
J. Zhang, and E. Hashemi, “Risk assessment and mitigation in local
path planning for autonomous vehicles with lstm based predictive
model,” IEEE Transactions on Automation Science and Engineering,
vol. 19, no. 4, pp. 2738–2749, 2021.
\item  X. Meng and C. G. Cassandras, “Eco-driving of autonomous vehicles for nonstop crossing of signalized intersections,” IEEE Transactions
on Automation Science and Engineering, vol. 19, no. 1, pp. 320–331,
2020.
\item S. Dixit, S. Fallah, U. Montanaro, M. Dianati, A. Stevens, F. Mccullough, and A. Mouzakitis, “Trajectory planning and tracking for
autonomous overtaking: State-of-the-art and future prospects,” Annual
Reviews in Control, vol. 45, pp. 76–86, 2018.
\item P. Stano, U. Montanaro, D. Tavernini, M. Tufo, G. Fiengo, L. Novella, and A. Sorniotti, “Model predictive path tracking control for automated road vehicles: A review,” Annual Reviews in Control, vol. 55, pp. 194– 236, 2023.
\item J. Hu, Y. Zhang, and S. Rakheja, “Adaptive trajectory tracking for carlike vehicles with input constraints,” IEEE Transactions on Industrial Electronics, vol. 69, no. 3, pp. 2801–2810, 2022.
\item W. T. Bryan, M. E. Boler, and D. M. Bevly, “A vehicle-independent
autonomous lane keeping and path tracking system**this work is
based on the lead author’s master’s thesis (bryan, 2020),” IFACPapersOnLine, vol. 54, no. 2, pp. 37–44, 2021, 16th IFAC Symposium
on Control in Transportation Systems CTS 2021.
\item Z. Chu, Y. Sun, C. Wu, and N. Sepehri, “Active disturbance rejection control applied to automated steering for lane keeping in autonomous vehicles,” Control Engineering Practice, vol. 74, pp. 13–21, 2018.
\item F. Karimi Pour, D. Theilliol, V.Puig, and G. Cembrano, “Healthaware control design based on remaining useful life estimation for autonomous racing vehicle,” ISA Transactions, vol. 113, pp. 196–209, 2021.
\item A. Liniger and J. Lygeros, “Real-time control for autonomous racing based on viability theory,” IEEE Transactions on Control Systems
Technology, vol. 27, no. 2, pp. 464–478, mar 2019.
\item E. Alcala´, V. Puig, J. Quevedo, and U. Rosolia, “Autonomous racing using linear parameter varying-model predictive control (lpv-mpc),”Control Engineering Practice, vol. 95, p. 104270, 2020.
\item Q. Yao, Y. Tian, Q. Wang, and S. Wang, “Control strategies on
path tracking for autonomous vehicle: State of the art and future
challenges,” IEEE Access, vol. 8, pp. 161 211–161 222, 2020.
\item B. R. Kiran, I. Sobh, V. Talpaert, P. Mannion, A. A. Al Sallab, S. Yogamani, and P. Perez, “Deep reinforcement learning for autonomous 
driving: A survey,” IEEE Transactions on Intelligent Transportation
Systems, vol. 23, no. 6, pp. 4909–4926, 2021.
\item D. Kamran, J. Zhu, and M. Lauer, “Learning path tracking for real
car-like mobile robots from simulation,” in 2019 European Conference
on Mobile Robots (ECMR), 2019, pp. 1–6.
\item E. Salvato, G. Fenu, E. Medvet, and F. A. Pellegrino, “Crossing the reality gap: A survey on sim-to-real transferability of robot controllers in reinforcement learning,” IEEE Access, vol. 9, pp. 153 171–153 187, 2021.
\item P. Cai, X. Mei, L. Tai, Y. Sun, and M. Liu, “High-speed autonomous drifting with deep reinforcement learning,” IEEE Robotics and Automation Letters, vol. 5, no. 2, pp. 1247–1254, apr 2020.
\item L. Dong, N. Li, H. Yuan, and G. Gong, “Accelerating wargaming
reinforcement learning by dynamic multi-demonstrator ensemble,”
Information Sciences, vol. 648, p. 119534, 2023.
\item J. Ramirez and W. Yu, “Reinforcement learning from expert demonstrations with application to redundant robot control,” Engineering Applications of Artificial Intelligence, vol. 119, p. 105753, 2023.
\item A. Edelmayer and M. Miranda, “Federated filtering for fault tolerant estimation and sensor redundancy management in coupled dynamics distributed systems,” in 2007 Mediterranean Conference on Control and Automation, 2007, pp. 1–6.
\item J. M. Snider et al., “Automatic steering methods for autonomous
automobile path tracking,” Robotics Institute, Pittsburgh, PA, Tech.
Rep. CMU-RITR-09-08, 2009.
\item uanser, “QCar,” https://www.quanser.com/products/qcar/, 2023, [Retrieved September 14, 2023].
\item The MathWorks Inc., “Matlab version: 9.13.0 (r2022b),” https://www.mathworks.com, 2022, [Retrieved September 14, 2023].
\item S. Cong, W. Wang, J. Liang, L. Chen, and Y. Cai, “An automatic
vehicle avoidance control model for dangerous lane-changing behavior,” IEEE Transactions on Intelligent Transportation Systems, vol. 23,
no. 7, pp. 8477–8487, 2022.
\item R. Sutton and A. G. Barto, Reinforcement learning: an introuduction. MIT press, 2018.
\item . Morales, Grokking deep reinorcement learning. Manning publications, 2020.
\item C. Chatzikomis, A. Sorniotti, P. Gruber, M. Zanchetta, D. Willans, and B. Balcombe, “Comparison of path tracking and torque-vectoring controllers for autonomous electric vehicles,” IEEE Transactions on Intelligent Vehicles, vol. 3, no. 4, pp. 559–570, 2018.
\item  Zhang, and S. Rakheja, “Adaptive trajectory tracking for car-like vehicles with input constraints,” IEEE Transactions on Industrial
Electronics, vol. 69, no. 3, pp. 2801–2810, 2022.
\end{enumerate}



\begin{IEEEbiography}[{\includegraphics[width=1in,height=1.25in,clip,keepaspectratio]{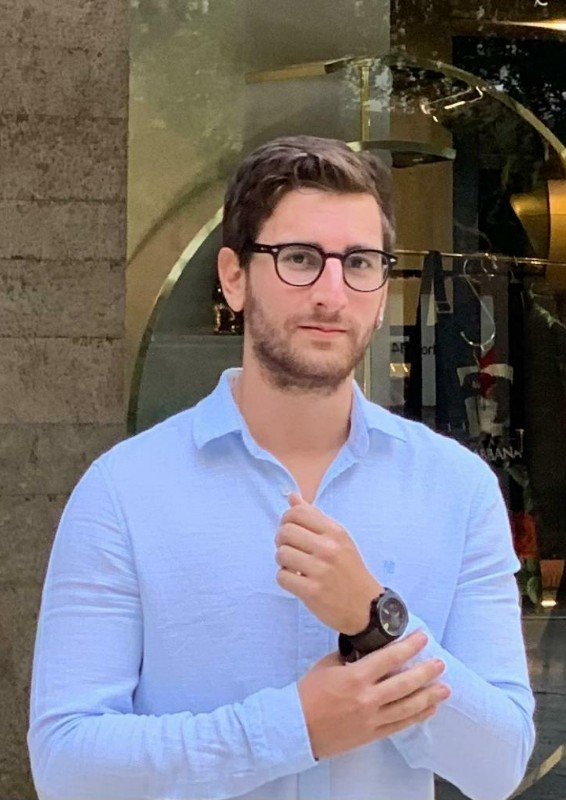}}]{Carmine Caponio} received the M. Sc. Degree in Mechanical Engineering cum laude in 2022 from the Politecnico di Bari, Bari, Italy. Currently he is working toward the Ph. D. degree in Automotive Engineering with the University of Surrey, Guildford, UK. His research interests focuses on model based and artificial intelligence based solutions for integrated chassis control of automated vehicles.

\end{IEEEbiography}

\begin{IEEEbiography}[{\includegraphics[width=1in,height=1.25in,clip,keepaspectratio]{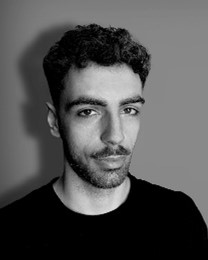}}]{Pietro Stano} received the M.Sc. degree in mechanical engineering from the Politecnico di Torino, Turin, Italy, in 2020. He is currently pursuing the Ph.D. degree in automotive engineering with the University of Surrey, Guildford, U.K. His main research interests include autonomous vehicle systems with a focus on trajectory planning and tracking in extreme handling conditions.

\end{IEEEbiography}

\begin{IEEEbiography}[{\includegraphics[width=1in,height=1.25in,clip,keepaspectratio]{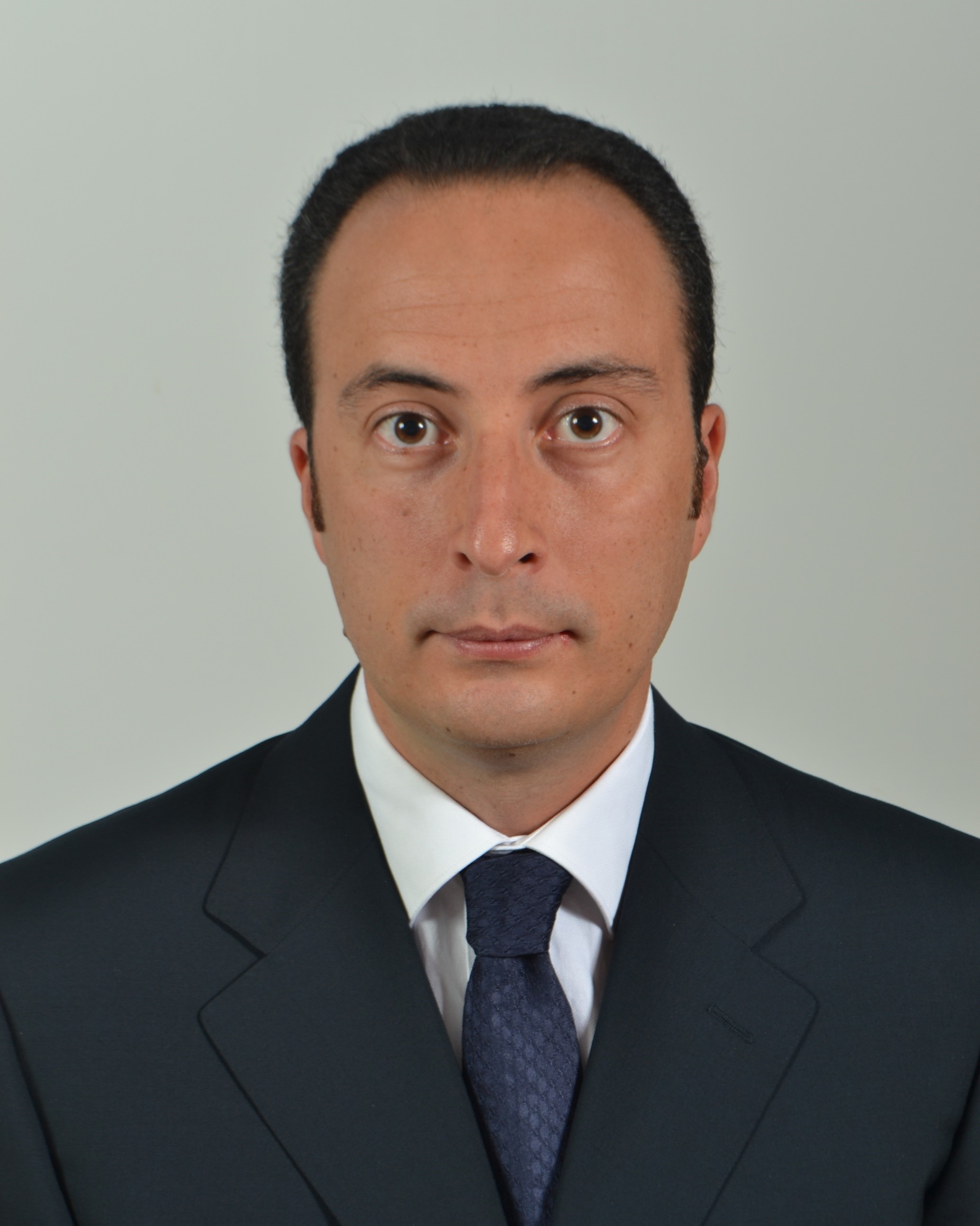}}]{Raffaele Carli} (M’17, SM’22) received the Laurea degree in Electronic Engineering with honours in 2002 and the Ph.D. in Electrical and Information Engineering in 2016, both from Politecnico di Bari, Italy.
From 2003 to 2004, he was a Reserve Officer with Italian Navy. From 2004 to 2012, he worked as System and Control Engineer and Technical Manager for a space and defense multinational company.

Dr. Carli is currently an Assistant Professor in Automatic Control at Politecnico di Bari, and his research interests include the formalization, simulation, and implementation of decision and control systems, as well as modeling and optimization of complex systems.

He is an Associate Editor of the IEEE TRANS. ON AUTOMATION SCIENCE AND ENGINEERING and the IEEE TRANS. ON SYSTEMS, MAN, AND CYBERNETICS. He was a member of the International Program Committee of 30+ international conferences and guest editor for special issues in international journals. He is the author of 90+ printed international publications.
\end{IEEEbiography}

\begin{IEEEbiography}[{\includegraphics[width=1in,height=1.25in,clip,keepaspectratio]{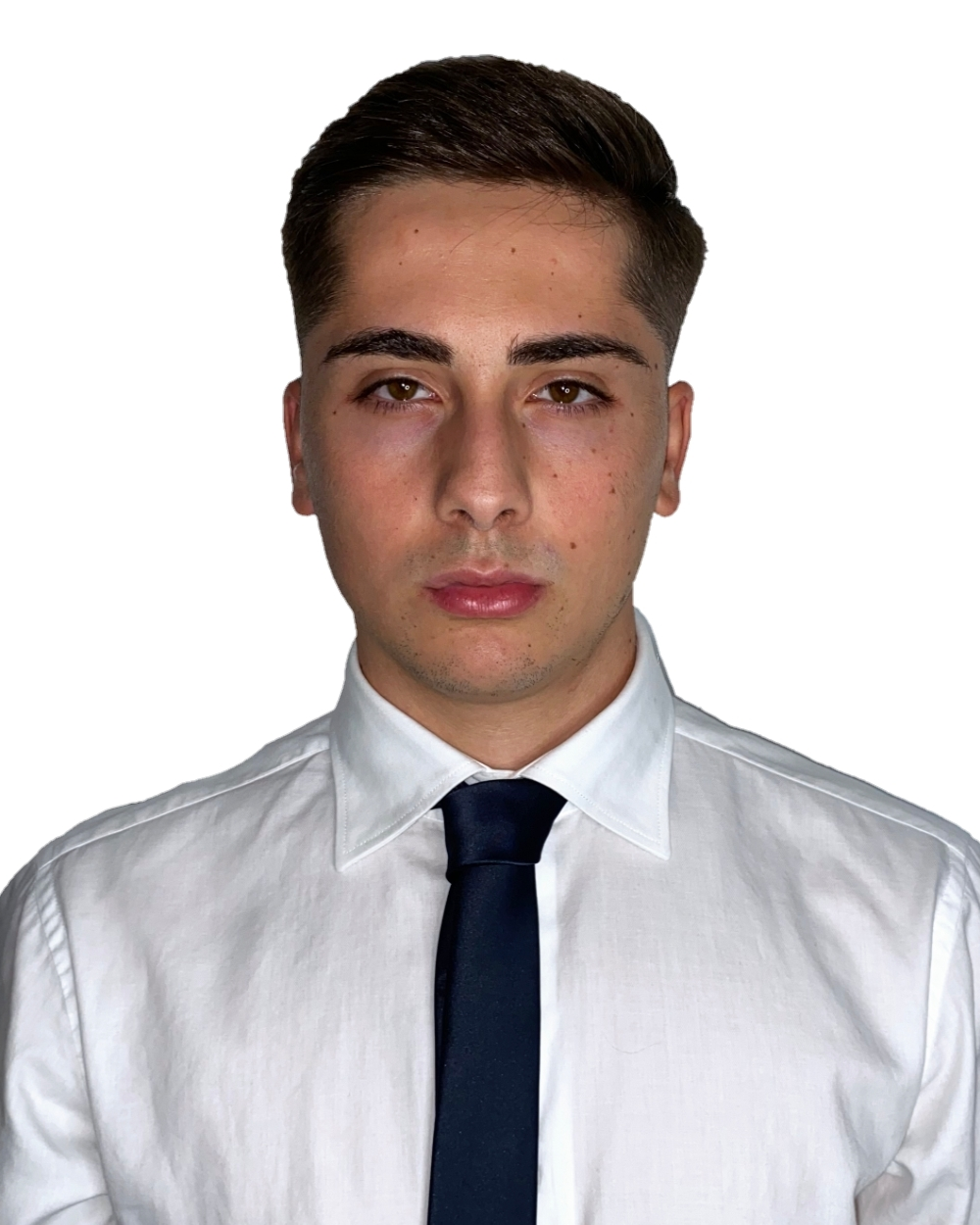}}]{Ignazio Olivieri} received the B.Sc. degree in computer science and automation engineering and the M.Sc. degree in automation engineering from the Politecnico di Bari, Bari, Italy, in 2021 and 2023, respectively. In 2023, he was a visiting student at the University of Surrey, Guildford, U.K. His research interests include autonomous vehicles modelling, model-based control, and deep reinforcement learning control.

\end{IEEEbiography}

\begin{IEEEbiography}[{\includegraphics[width=1in,height=1.25in,clip,keepaspectratio]{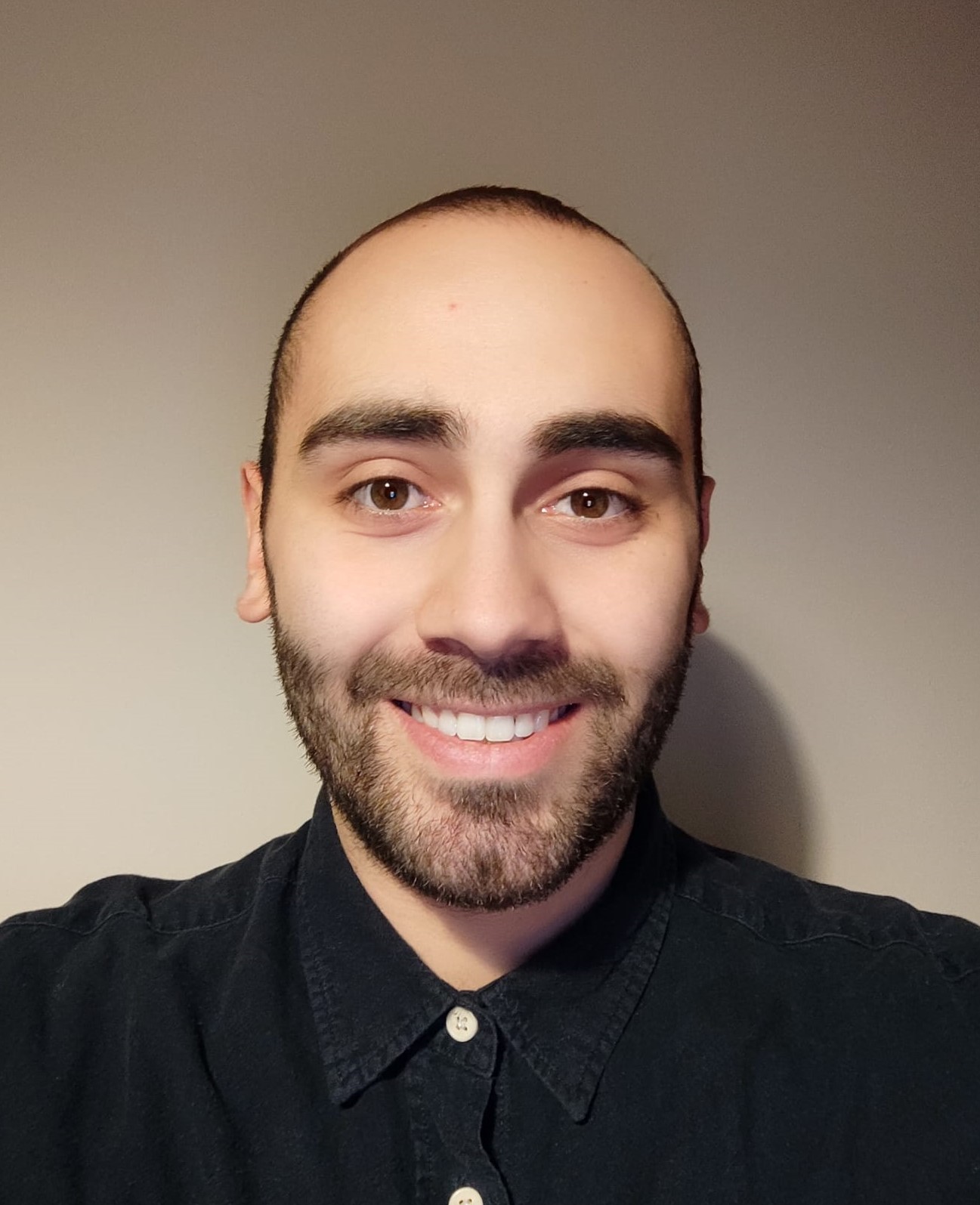}}]{Daniele Ragone} obtained his B.Sc. degree in computer science and automation engineering and the M.Sc. degree in automation engineering from the Polytechnic of Bari, Bari, Italy, in 2020 and 2023 respectively. In 2023 he was a visiting student at the University of Surrey, Guildford, UK. He focused on autonomous vehicle modeling, model parameter identification and state estimation with nonlinear Kalman filter.

\end{IEEEbiography}

\begin{IEEEbiography}[{\includegraphics[width=1in,height=1.25in,clip,keepaspectratio]{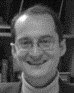}}]{Aldo Sorniotti} (M’12) received the M.Sc. degree in mechanical engineering and Ph.D. degree in applied mechanics from the Politecnico di Torino, Turin, Italy, in 2001 and 2005. He is a Full Professor in Applied Mechanics with the Politecnico di Torino, after being a Professor in advanced vehicle engineering with the University of Surrey, Guildford, U.K., where he led the Centre for Automotive Engineering. His research interest is on vehicle dynamics control for electric vehicles.

\end{IEEEbiography}

\begin{IEEEbiography}[{\includegraphics[width=1in,height=1.25in,clip,keepaspectratio]{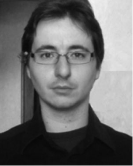}}]{Umberto Montanaro} received the M.Sc. degree in computer science engineering and the Ph.D. degrees in control engineering and mechanical engineering from the University of Naples Federico II, Naples, Italy, in 2005, 2009, and 2016, respectively. He is currently a Senior Lecturer in control engineering and autonomous systems with the University of Surrey, Guildford, U.K. Moreover, he is the director of Surrey team for the Control of Smart Multi-agent systems Operating autonomously and Synergistically (S-COSMOS). His research interests include adaptive control, and control of piecewise affine, mechatronic, automotive systems and coordination of networked autonomous systems.

\end{IEEEbiography}

\end{document}